\documentclass{article}

 \usepackage[preprint]{neurips_2026}

\usepackage[utf8]{inputenc} %
\usepackage[T1]{fontenc}    %
\usepackage[colorlinks=true]{hyperref}       %
\usepackage{url}            %
\usepackage{booktabs}       %
\usepackage{amsfonts}       %
\usepackage{nicefrac}       %
\usepackage{microtype}      %
\usepackage{xcolor}         %

\usepackage{amsmath}
\usepackage{amssymb}
\usepackage{amsthm}
\usepackage[textsize=tiny]{todonotes}
\usepackage{multirow}
\usepackage{wrapfig}
\usepackage{enumitem}
\usepackage[table]{xcolor}
\usepackage{subcaption}
\usepackage{algorithm}
\usepackage{algpseudocode}
\usepackage{listings}
\usepackage{xcolor}
\usepackage{etoc}
\usepackage{minitoc}
\usepackage{mathtools}
\usepackage{tabularx}
\usepackage{placeins}
\usepackage{wrapfig}
\usepackage{soul}
\usepackage{makecell}

\definecolor{KausableOrange}{HTML}{FF9B00}
\definecolor{KausableTeal}{HTML}{006471}
\definecolor{KausableRed}{HTML}{C35353}

\hypersetup{
  linkcolor=KausableTeal,
  citecolor=KausableRed,
  urlcolor=KausableTeal
}

\title{In-context learning to predict critical transitions in dynamical systems}

\author{%
Yunus Sevinchan\thanks{Equal contribution} \\
kausable \\
Heidelberg, Germany \\
\texttt{yunus@kausable.ai} \\
\And
Juan Nathaniel\footnotemark[1] \\
Columbia University\\
New York, USA \\
\texttt{jn2808@columbia.edu} \\
\And
Kai Ueltzhöffer\footnotemark[1] \\
kausable\\
Heidelberg, Germany \\
\texttt{kai@kausable.ai}\\
\And
Carla Roesch\footnotemark[1] \\
University of Edinburgh \\
Scotland, UK \\
\texttt{carla.roesch@ed.ac.uk} \\
\And
Vaios Laschos\footnotemark[1] \\
kausable\\
Heidelberg, Germany \\ 
\texttt{vaios@kausable.ai}
\And
Tobias Weber \\
kausable\\
Heidelberg, Germany \\ 
\texttt{tobias@kausable.ai}
\And
Hang Fan \\
Columbia University\\
New York, USA \\
\texttt{hf2526@columbia.edu} \\
\And 
Gregor Ramien\\
kausable\\
Heidelberg, Germany \\
\texttt{gregor@kausable.ai} \\
\And
Johannes Haux\\
kausable\\
Heidelberg, Germany \\
\texttt{johannes@kausable.ai} \\
\And
Pierre Gentine \\
Columbia University \\
New York, USA \\
\texttt{pg2328@columbia.edu}\\
\And
Benjamin Herdeanu\footnotemark[1]\\
kausable \\
Heidelberg, Germany \\
\texttt{benjamin@kausable.ai} \\
}

\begin{document}
\etocdepthtag.toc{mtoc}

\maketitle

\begin{abstract}
Critical transitions -- abrupt, often irreversible changes in system dynamics -- arise across human and natural systems, often with catastrophic consequences.
Real-world observations of such shifts remain scarce, preventing the development of reliable early warning systems. 
Conventional statistical and spectral indicators, such as increasing variance, tend to fail under realistic conditions of limited data and correlated noise, whereas existing deep learning classifiers do not extrapolate beyond their training data distribution.
In this work, we introduce TipPFN, an in-context learning (ICL) framework that uses a prior-data fitted network to infer a system's proximity to a critical transition.
Trained on our novel synthetic data generator, which is based on canonical bifurcation scenarios coupled to diverse, randomized stochastic dynamics, TipPFN flexibly capitalizes on contexts of various sizes, complexity and dimensionalities.
We demonstrate robust, state-of-the-art early detection of critical transitions in previously unseen tipping regimes, sim-to-real examples, and real-world observations in both ICL and zero-shot settings.
\end{abstract}

\newpage
\section{Introduction}

\begin{wrapfigure}{r}{2.35in}
    \centering
    \includegraphics[width=1.9in]{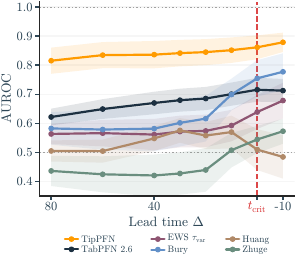}
    \caption{\textbf{TipPFN detects critical transitions} earlier and more accurately than classical early warning signals (EWS), state-of-the-art deep learning and ICL-based methods across the 14 systems of Table~\ref{tab:auroc-results-best-col} except DaphniaExt, spanning climate, engineering, and biology. Lines show across-system mean and standard error.
    }
    \label{fig:headline}
\end{wrapfigure}

Tipping points occur when variation in forcing or system parameters triggers an abrupt, potentially irreversible transition to a qualitatively different dynamical regime~\cite{fang2025tipping,ashwin2012tipping}.
Such transitions are central to climate~\cite{romanou2023stochastic}, ecological~\cite{veraart2012recovery}, infrastructural~\cite{council1996western}, and collective biological and social systems~\cite{gomeznava2023fish,ritchie2023rate,sevinchan2025collective}, where local shifts can cascade through interacting components with far-reaching consequences. Classical theory has largely focused on bifurcation (b-tipping) classes of tipping mechanism. Here, the loss of stability of an attracting state produces critical slowing down (CSD)~\cite{dakos2012robustness}, where recovery from perturbations becomes progressively weaker~\cite{wissel1984universal,van2007slow}.
This has motivated widely used early warning signals (EWS), including increasing lag-1 autocorrelation (AR1)~\cite{dakos2008slowing}, variance~\cite{carpenter2006rising}, and skewness~\cite{guttal2008changing}.
However, these indicators rely on restrictive assumptions, including near-equilibrium linearization, stationarity, sufficiently long records, and simple bifurcation structure~\cite{dakos2015resilience,kefi2013early}.
They may therefore fail for rate-induced (r-tipping) and noise-induced (n-tipping), where transitions can occur without pronounced CSD~\cite{ritchie2023rate,pavithran2021effect,huang2024deep}.

Characterizing critical transitions requires forecasting an unseen future regime from observations in a qualitatively distinct initial regime, so that the data-generating dynamics shift between what is observed and what must be predicted~\cite{liu2023outofdistributiongeneralizationsurvey,zhou2022domain,wu2025outofdistributiongeneralizationtimeseries}.
This makes tipping prediction fundamentally different from standard regression tasks, since models must infer the onset and type of critical transitions from sparse, noisy data and partially observed trajectories, often without examples of the event itself.
Throughout, we use \emph{generalization} to mean empirical transfer to held-out tipping mechanisms and domains.
Recent ML approaches have shown promise by learning features of canonical tipping systems directly from data~\cite{bury2021deep,li2026ultra,ritchie2023rate,zhuge2025deep,panahi2024machine}. However, many of these methods are trained on particular classes, limiting their ability to generalize across observations and different critical mechanisms.

A promising alternative has recently emerged in the form of prior-data fitted networks (PFNs), which are trained on large ensembles of synthetic datasets and learn to perform inference via in-context learning (ICL) \cite{muller2022transformers}. Rather than fitting a model to a single dataset, PFNs are trained across millions of simulated tasks, enabling them to infer structure and make predictions directly from limited observations. Thus, PFNs have already transformed tabular ML through models such as TabPFN \cite{hollmann2023tabpfntransformersolvessmall,hollmann2025accurate,qu2026tabiclv2betterfasterscalable}, which, despite being trained exclusively on synthetic data, achieve state-of-the-art performance on real-world benchmarks. More broadly, PFNs have demonstrated remarkable generalisation across diverse tasks, including time series forecasting \cite{hoo2026tablestimeextendingtabpfnv2}, causal inference \cite{robertson2025dopfnincontextlearningcausal,balazadeh2025causalpfnamortizedcausaleffect}, and robotic controls~\cite{schiff2025gradientfreedeepreinforcement}. 

\begin{figure}[t]
    \centering
    \includegraphics[width=\linewidth]{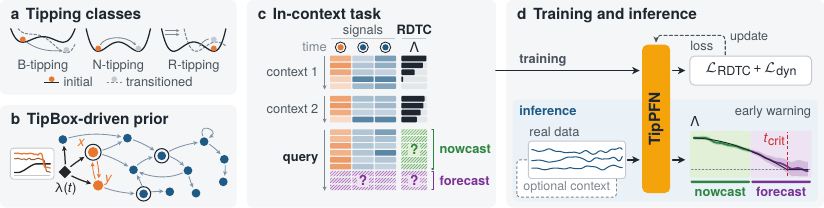}%
    \caption{
        \textbf{Overview.} \textbf{(a)} Tipping classes (Sec.~\ref{ch:background}).
        \textbf{(b)} The TipBox-driven prior embeds canonical b-tipping drivers $(x, y)$, forced by $\lambda(t)$, into randomized stochastic auxiliary dynamics, of which a random subset is observed.
        \textbf{(c)} A training task: context episodes with known RDTC $\Lambda$, and a query whose RDTC is masked and predicted for the observed rows (nowcast) and beyond them (forecast).
        \textbf{(d)} TipPFN is trained on such tasks and applied to real data, optionally with context; it transfers to other tipping classes and to several real-world systems.}
    \label{fig:overview}
\end{figure}

Our \textbf{key contributions} are (see Figure~\ref{fig:overview}):
\begin{itemize}
    \item \textbf{Model:} TipPFN, an in-context learning model that infers proximity to a critical transition (RDTC) from short, noisy, multivariate time series.
    \item \textbf{Training data and benchmark:} a training dataset and evaluation benchmark of stochastic dynamical systems across diverse tipping regimes and complexities, simulated with TipBox\footnote{\href{https://github.com/juannat7/tipbox}{\nolinkurl{github.com/juannat7/tipbox}}}, released together with the trained model and code\footnote{Code: \href{https://github.com/kausable/tippfn-neurips26}{\nolinkurl{github.com/kausable/tippfn-neurips26}}; model and data: \href{https://huggingface.co/kausable}{\nolinkurl{huggingface.co/kausable}}.}.
    \item \textbf{Generalization:} Robust performance across unseen tipping regimes, sim-to-real cases, and real-world systems in both ICL and zero-shot settings (see Fig.~\ref{fig:headline}).
\end{itemize}

\section{Background}\label{ch:background}
\paragraph{Critical transition in dynamical systems} Critical transition, specifically tipping points in dynamical systems, can arise through multiple distinct mechanisms, broadly classified as bifurcation-induced (b-tipping), noise-induced (n-tipping), and rate-induced tipping (r-tipping)~\cite{ashwin2012tipping} (Fig.~\ref{fig:overview}a). Consider a general nonautonomous stochastic differential equation (SDE) of the form:
\begin{equation}
    \frac{d \mathbf{x}}{dt} = f(\mathbf{x}, \lambda(t)) + \sigma \, \xi(t),
\end{equation}
where $\mathbf{x} \in \mathbb{R}^d$ denotes the system state, $\lambda(t)$ is a time-dependent forcing parameter, $\sigma > 0$ controls the noise amplitude, and $\xi(t)$ represents Gaussian white noise. 

\paragraph{Bifurcation-induced tipping} In b-tipping, a transition occurs when a quasi-static equilibrium $\mathbf{x}^\ast(\lambda)$ loses stability at a critical parameter value $\lambda_\mathrm{crit}$. Linearizing the dynamics around $\mathbf{x}^\ast(\lambda)$ yields the Jacobian $J(\lambda) = \partial f / \partial \mathbf{x} \big|_{\mathbf{x}^\ast(\lambda)}$, whose eigenvalues $\{\mu_i(\lambda)\}$ govern local stability. A bifurcation occurs when the real part of the leading eigenvalue crosses zero, i.e.,
\begin{equation}
\max_i \, \mathrm{Re}(\mu_i(\lambda_\mathrm{crit})) = 0.
\end{equation}

\paragraph{Rate-induced tipping} In r-tipping, a transition occurs when the rate of change of the forcing, given by $\dot{\lambda}$, is sufficiently large such that the system fails to track the moving equilibrium $\mathbf{x}^\ast(\lambda(t))$, leading to a transition without any local loss of stability, i.e., the linearized system satisfy $\max_i \mathrm{Re}(\mu_i(\lambda)) < 0$. Thus, these tipping mechanisms cannot, in general, be inferred solely from equilibrium stability analysis commonly used to check b-tipping (see Appendix~\ref{si-sec:add-background} for more details).

\paragraph{Noise-induced tipping} In n-tipping, a transition occurs primarily through stochastic perturbations $\sigma \, \xi(t)$ that induce escape from the basin of attraction of a stable equilibrium. This process can occur in conjunction with b-tipping and r-tipping in SDEs, a setup which form the basis for this work. 

\paragraph{Prior-data fitted networks} PFNs are transformer-based neural networks trained on synthetically generated data drawn from a predefined prior, enabling approximate Bayesian inference via ICL~\cite{muller2022transformers}. In this framework, inference is amortized at the dataset level. The model learns a mapping from a dataset $D$ to a posterior predictive distribution $p(y|x,D)$ by training on large collections of simulated tasks $(D_i, x_i, y_i)$. Rather than fitting a model to each new dataset, PFNs directly infer predictions conditioned on the observed data, performing Bayesian inference in a single forward pass. The Bayesian interpretation of PFN predictions only holds on the support of the training prior.

\paragraph{Related work} In addition to the classical EWS based on statistics (AR1, variance, skewness), there is an emergence of dynamics-based approaches extracting additional spectral structure of the systems, including tracking of dominant eigenvalues~\cite{miyauchi2026generalized,grziwotz2023anticipating,nathaniel2026koopman}. Recently, fully data-driven ML approaches, including reservoir computing~\cite{panahi2024machine,li2026ultra}, convolutional, and recurrent-based neural networks~\cite{bury2021deep,zhuge2025deep}, have been proposed to detect tipping points directly from time series data. These models can outperform both classical and spectral EWS when trained on simulated data, but are tailored to specific tipping regimes (e.g., b-tipping~\cite{bury2021deep,zhuge2025deep} or r-tipping~\cite{huang2024deep}). Closest to our approach, \cite{bury2021deep} also pre-trains on synthetic dynamical systems, but uses univariate inputs, a single categorical label per sequence (bifurcation type or no transition) as supervision, and no in-context learning. They also show limited generalization capabilities, which we further showcase throughout.

\section{TipPFN: Predicting tipping behaviors with PFNs}\label{sec:tippfn}

\begin{figure}[b]
    \centering
    \includegraphics[width=1.08\textwidth]{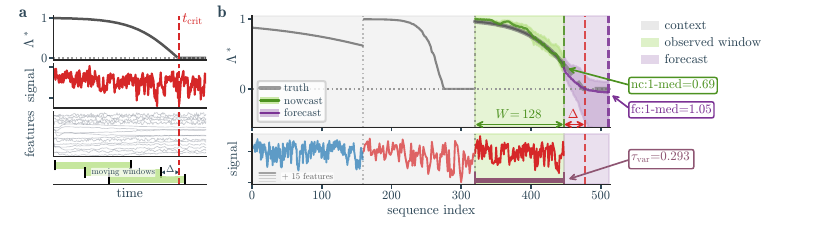}
    \caption{\textbf{Tipping Prediction with TipPFN.}
        \textbf{(a)} Example multi-variate time series from a critical episode and underlying RDTC~$\Lambda$.
        Green bars mark candidate observation windows ending $\Delta$ time steps before the critical time $t_{\mathrm{crit}}$.
        \textbf{(b)} Example query episode at $\Delta=30$ and context composed from a critical (red) and non-critical (blue) episode.
        TipPFN and TabPFN are conditioned on the shaded context and observed window, including the signal and all available features, and then predict relative distance to criticality, $\Lambda^*$, for the nowcast and forecast region.
        These predictions form the basis of the tipping-risk scores used for early-warning assessment.
        Baseline scores (e.g. EWS $\tau_\mathrm{var}$) are computed only from the observed signal window.
    }
    \label{fig:tippfn-prediction}
\end{figure}

TipPFN is designed to infer approaching critical transitions directly from observed trajectories.
To this end, we train a model on synthetic dynamical systems in which a control parameter~$\lambda$ is gradually varied to produce trajectories that progress from stable regimes to towards a critical transition.
By embedding controlled, stereotypical tipping systems within larger, randomized systems of nonlinear SDEs, the synthetic data provide a scalable source of diverse transition patterns and allow the model to learn transferable structure rather than system-specific rules.
This is essential for our setting, where the goal is generalization to previously unseen systems.

Next, we define the prediction target, the relative distance to criticality.
We then describe synthetic data generation, and finally the TipPFN training and inference setup; also see Fig.~\ref{fig:tippfn-prediction} and Appendix~\ref{si-sec:pfns}.

\paragraph{Relative distance to criticality (RDTC)} Direct supervision on critical events is often ill-posed, since the observed transition time can depend on noise, finite-time effects, and threshold choices. For the bifurcation-induced tipping systems used in training, however, proximity to criticality is well defined through the forcing parameter~$\lambda$.

We therefore introduce the \emph{relative distance to criticality} (RDTC) $\Lambda$ as a continuous supervision target.
Let $\tilde{\lambda}\in[0,1]$ denote the normalized control parameter, where $\tilde{\lambda}=0$ corresponds to a stable regime and $\tilde\lambda_{\mathrm{crit}}=1$ to the forcing value at which the underlying deterministic system undergoes a bifurcation.
We define
\begin{equation}
    \Lambda = 1 - \tilde\lambda.
\end{equation}
For $\Lambda=1$, the system is maximally far from the bifurcation within the normalized forcing range; for $0 < \Lambda < 1$, it remains in the sub-critical regime, reaching criticality at $\Lambda_{\mathrm{crit}}=0$.
RDTC serves as the primary training target for synthetic bifurcation systems. Furthermore, it is intended to act as a transferable measure of proximity to criticality across otherwise distinct systems.

\paragraph{Prior structure} TipPFN's prior $p(\psi)$ over generative processes $\psi$ factorises over a canonical tipping system $M$, a random potentially cyclic interaction graph $\mathcal{G}$ and an auxiliary nonlinear SDE system $\mathcal{Z}$. $\mathcal{G}$ specifies the interaction structure of $\mathcal{Z}$, and how it is driven by the state variables ${z_M(t) = (x_M(t), y_M(t))}$, the forcing parameter $\lambda(t)$ and the RDTC $\Lambda(t)$. This driving timeseries is generated by sampling from a flexible family of low-dimensional nonlinear dynamical systems designed to exhibit canonical tipping behavior~\cite{bury2021deep} (see Appendix~\ref{si-sec:forcing_datasets} for details).
We define the prior over the generative process $\psi = (M, \mathcal{G}, \mathcal{Z})$ as:
\begin{equation}
    p(\psi) = p(M)\, p(\mathcal{G})\, p(\mathcal{Z} \mid M, \mathcal{G}),
\end{equation}
where $M \sim \mathrm{Uniform}(\{\text{fold},\, \text{Hopf},\, \text{transcritical}\})$ determines the bifurcation class inducing qualitatively different signatures in the pre-tipping dynamics. $\mathcal{G}$ is generated by drawing from a distribution over directed, potentially cyclic graphs with power-law distributions over incoming and outgoing node degrees \cite{birmele2009scaleFreeBipartite} to yield the interaction structure within the auxiliary dynamic nodes, followed by random attachment of four driver nodes representing $M$'s state variables, its forcing parameter and the RDTC. Based on this dynamical causal structure, nonlinear SDE interaction terms are drawn from a distribution over multi-layer perceptrons. Additional linear stabilization terms and a global timescale parameter are added to the auxiliary nodes and also randomized. The hyperparameter distribution of the auxiliary system is chosen to keep it in a stable regime, preventing additional, spontaneous bifurcations due to intrinsic auxiliary system dynamics, which would confound the signal provided by the driving system.

\paragraph{Prior-data generation} As illustrated in Fig.~\ref{fig:metrics}a, to fit the prior, we specify a sampling scheme over context-query pairs $\mathcal{D}$ of the form
\begin{equation}
    p(\mathcal{D}) = \mathbb{E}_{\psi \sim p(\psi), \mathcal{S} \sim p(\mathcal{S} | \psi)}\bigl[p(\mathcal{D} \mid \mathcal{S})\bigr],
\end{equation}
which first samples a generative process $\psi = (M, \mathcal{G}, \mathcal{Z}) \sim p(\psi)$ and then a synthetic episode ensemble $\mathcal{S} = \{\mathcal{E}_k\}_{k=1}^{K} \sim p(\mathcal{S} \mid \psi)$ which fixes the canonical tipping system $M$, the interaction graph $\mathcal{G}$ and SDE system $\mathcal{Z}$.
Each episode $\mathcal{E}_k = (z^k_{1:T},u^k_{1:T},\tilde\lambda^k_{1:T})$ is based on an independent noise realization and contains the trajectories of the driving system $z^k_{1:T}$ and the auxiliary nodes $u^k_{1:T}$ created by the independent forcing schedule $\tilde\lambda^k_{1:T}$ (Bezier, ramp, or constant; Table~\ref{tab:v2-forcing}).
Forcing schedules are constrained to produce trajectories spanning multiple dynamical regimes: \textit{critical} (tipping), \textit{approaching} towards or \textit{receding} from criticality, \textit{constant}, and \textit{equilibrium} (non-critical); see Appendix~\ref{si-sec:forcing}.
From each sample $\mathcal{S} = \{\mathcal{E}_k\}_{k=1}^{K}$, one episode $\mathcal{E}^{\mathrm{query}}$ is held out as the query and a context set $\mathcal{C} \subset \mathcal{E} \setminus \mathcal{E}^{\mathrm{query}}$ of $\left| \mathcal{C} \right| \in \left\{0,..,3\right\}$ context trajectories, with $\left| \mathcal{C} \right|$ drawn with probabilities 0.2, 0.3, 0.3 and 0.2, is sampled uniformly from the remaining $K-1$ episodes.
A subset of feature dimensions of the driving system $z$ and the auxiliary system $u$ is randomly selected to yield the observed features $\mathbf{o}$ with $\mathrm{dim}(\mathbf{o}) \in \left\{1:16\right\}$, rendering some the resulting prediction problems only partially observed.
By resampling the time series with different resolutions, we realize variable-context training tasks in which query and context episodes share the same underlying system class and coupling structure, but differ in forcing and stochastic realization. This yields a query-context pair $\mathcal{D} = ((\Lambda_{1:T}, \mathbf{o}_{1:T}) \cup \mathcal{C})$. As a result, the model learns to use context when informative without depending on its presence.
In total, we trained on $\sim$3.7M context-query pairs~$\mathcal{D}$ (22k optimizer steps at batch size 168) from $\sim$270k $\psi$ samples with $K=4$ episodes and $M\in\mathbb{R}^2,~\mathcal{Z}\in\mathbb{R}^{14}$.

\paragraph{Input representation} Each training task is encoded as a tabular multi-episode sequence with episode and time identifiers, observed variables, target variables, and prediction masks. To avoid target leakage, variables are normalized using only observed values.
Observed and target variables are centered and transformed with $\mathrm{asinh}$ scaling, which reduces the effect of large amplitude differences across heterogeneous signals.
RDTC is transformed separately as $\Lambda^* = \tanh{(5\Lambda)}$, the nonlinearity providing finer resolution near criticality.
In the query episode, RDTC is always fully masked, whereas the remaining target variables are only partially masked.

\paragraph{Architecture} TipPFN is a transformer-based architecture with a similar structure to TabPFN \cite{hollmann2023tabpfntransformersolvessmall,hollmann2025accurate}; the main adaptations for tipping prediction lie in the task construction, masking scheme, and training objective described above.
Transformer attention is structured, such that all tokens from the context set $\mathcal{C}$ can attend to all other tokens of $\mathcal{C}$.
Tokens from the query trajectory can attend to all tokens from $\mathcal{C}$, themselves, and all tokens from the query trajectory with smaller time features, i.e., tokens representing preceding time steps (\emph{causal attention}), as illustrated in Fig.~ \ref{fig:metrics}b.
This ensures that no future information leaks into the estimate at the end of the observed window, as required for early warning signals. More details on the architecture of PFNs is provided in Appendix~\ref{si-sec:pfns}.
The released model has $\sim$210M parameters; much smaller variants perform almost as well (Sec.~\ref{sec:ablations}).

\paragraph{Training} TipPFN is trained as a TabPFN-style transformer on tabularized multi-episode trajectories.
The model predicts quantile distributions for masked targets and is optimized with a weighted pinball loss.
Along the temporal dimension, attention is causal, so each prediction can depend only on past and present observations. Context episodes are fully visible.

The parameters $\theta$ of a transformer model $q_\theta$ are optimised using the following training loss:
\begin{equation}
    \mathcal{L} =
    \mathbb{E}_{\mathcal{D} = ((\Lambda_{1:T}, \mathbf{o}_{1:T}) \cup \mathcal{C}) \sim p(\mathcal{D})}
    \left[
        \sum_{t=1}^T \frac{1}{99} \sum_{j=1}^{99} \ell_{\alpha_j}\!\left(\hat q_{\theta,j}\!\left(\mathbf{o}_{1:\min(t,t_\mathrm{nc})},\, \mathcal{C}\right),\, \Lambda^*_t\right)
    \right].
    \label{eq:loss}
\end{equation}
Here, $\hat q_{\theta,j}$ is the predicted quantile at level $\alpha_j \in \{0.01, \dots, 0.99\}$, $\ell_\alpha$ the pinball loss (Appendix~\ref{si-sec:tippfn-training}), $\Lambda^*_t$ the transformed RDTC at time step $t$ in the query episode, and $\mathbf{o}_{1:\min(t,t_\mathrm{nc})}$ denotes query observations up to step $t$ or a cut-off $t_\mathrm{nc} < t$, up to which observational data is available, to train nowcasting and forecasting capabilities. The RDTC prediction loss is augmented with a similar loss, scaled by a factor of 0.2 relative to the RDTC loss, which requires to predict a randomly selected subset of the observational features from the remaining set of at least 1 feature. This loss was added to encourage the model to learn a more general, task-agnostic representation of the nonlinear stochastic dynamics of the synthetic training data.

\paragraph{Posterior Predictive Distribution for RDTC} By minimising the training loss given in Equation~\ref{eq:loss}, the quantiles predicted by the transformer model $q_{\theta^\mathrm{opt}}$ approximate those of the true Bayesian posterior predictive distribution (PPD) \cite{muller2022transformers}:

\begin{align*}
    q_{\theta^\mathrm{opt}}\!\left(\Lambda_t \mid \mathbf{o}_{1:\min(t,t_\mathrm{nc})},\, \mathcal{C}\right) & \approx \, 
    p(\Lambda_t \mid \mathbf{o}_{1:\min(t,t_\mathrm{nc})},\, \mathcal{C}) \\
    & =
    \int_{\Psi}
        p(\Lambda_t \mid \mathbf{o}_{1:\min(t,t_\mathrm{nc})},\, \mathcal{C}, \psi)\,
        p(\psi \mid \mathbf{o}_{1:\min(t,t_\mathrm{nc})},\, \mathcal{C})\,
    \, d\psi.
    \label{eq:ppd}
\end{align*}

Thus, TipPFN approximates this intractable integral in a single transformer forward pass by amortising inference over a large ensemble of synthetic prediction problems sampled from $p(\psi)$ during training, yielding a predictive distribution approximating Bayes-optimal uncertainty estimates.
When no context episodes are available, the prediction relies entirely on the query trajectory and the prior $p(\psi)$. Increasing the available context sharpens the posterior $p(\psi \mid \mathbf{o}_{1:t},\, \mathcal{C})$ over generative processes compatible with the observed data, depending on the information content of the added context samples. 

\paragraph{Inference} TipPFN supports two inference modes depending on the operational setting:
For \textit{nowcasting}, it estimates the current RDTC at time $t$, conditioned on all observations up to that point:
\begin{equation}
    q_{\theta^\mathrm{opt}}\!\left(\Lambda_t \mid \mathbf{o}_{1:t},\,
    \mathcal{C}\right),
\end{equation}
where $\Lambda_{t}$ denotes the RDTC at the current time step $t$. This mode is appropriate when the primary goal is to monitor the system's instantaneous distance to a critical point in real time, issuing an updated estimate as each new observation arrives.

In the \textit{forecasting} setting, TipPFN predicts the future trajectory of RDTC, conditioned on observations up to the current time step:
\begin{equation}
    q_{\theta^\mathrm{opt}}\!\left(\Lambda_{t:T} \mid \mathbf{o}_{1:t},\,
    \mathcal{C}\right),
\end{equation}
where  $\Lambda_{t:T}$ denotes the sequence of RDTC values from the current time step $t$ to a future horizon $T$. This mode is appropriate when the goal is to anticipate if the system will approach a tipping point, enabling proactive intervention before the critical transition occurs.

\section{Experiments}

In this section, we evaluate the performance and predictive skill of TipPFN for tipping behavior across different mechanisms.
We benchmark against a comprehensive set of competitive baselines on datasets ranging from synthetic systems to real-world observations. Evaluation is based on classification performance (critical vs.\ non critical) using receiver operating characteristic (ROC) curves \cite{hanley1982meaning,fawcett2006introduction} and Area Under the ROC curve (AUROC) scores \cite{peterson1954theory,bradley1997use}, as well as lead-time analysis; more details in Appendix~\ref{si-sec:experiments}.

\paragraph{Validation datasets} We generated and collected cross-domain validation datasets covering 12 model families (canonical and semi-real) and 9 observational systems (sim-to-real \& real-world); see Appendix~\ref{si-sec:datasets} for the full list and descriptions.

\textit{Canonical} datasets are held-out realizations from the synthetic prior.~~
\textit{Semi-real} datasets are reduced-order models of real-world dynamics that are more complex than the canonical systems and include unseen tipping behavior, e.g., r-ipping.~~
\textit{Sim-to-real} datasets contain both simulated and observed trajectories, allowing us to test whether matched simulations can provide context for empirical transition prediction.~~ %
\textit{Real-world observation} datasets consist of empirical time series from systems without fully specified generative dynamics, including neurological seizure recordings, ocean circulation (AMOC), power-grid blackout, and cyanobacteria population dynamics; depending on the available data, they support either context-based transfer or strict zero-shot RDTC prediction.

\paragraph{Baselines} We compare TipPFN against classical EWS, including AR1, variance, and skewness in rolling windows, summarize their temporal trends with Kendall-$\tau$~\cite{kendall1938new}, and obtain ROC curves by thresholding these trend scores, following standard practice~\cite{dakos2012robustness}.
ML baselines include Bury~\cite{bury2021deep}, Huang~\cite{huang2024deep}, and Zhuge~\cite{zhuge2025deep}, operating on univariate time series.
TabPFN2.6~\cite{hollmann2025accurate} provides a state-of-the-art ICL baseline; more details are provided in Appendix~\ref{si-sec:baselines}.

\paragraph{Querying procedure \& scoring}\label{sec:querying-procedure}
All methods are evaluated on matched moving query windows ending~$\Delta$ time steps before the critical event.
Uni-variate baselines were evaluated on the driving time series, $z_M(t)$.
PFN-based models additionally receive context episodes, whereas classical and ML baselines use only the query window; see Fig.~\ref{fig:tippfn-prediction}.
The original window's time is re-indexed to start at 0 such that no positional information is leaked.
At test time, no method has direct access to future observations, hidden system parameters, or ground-truth RDTC, except for Zhuge, which also observes the forcing parameter and therefore receives privileged query-episode information.
On the synthetic systems, the auxiliary variables are in addition driven by the forcing and the RDTC (Sec.~\ref{sec:tippfn}), so methods that observe them, TipPFN and TabPFN, also receive indirect information about the forcing level that the univariate baselines do not.
Appendix~\ref{si-sec:aux-drive-sensitivity} reports results with auxiliary variables driven by the state variables only.

As the ML baselines require a window size $W\geq500$ steps while TipPFN uses $W\leq128$, we construct length-adapted baseline inputs by resampling, back-filling, or forward-filling the query window.
For each model, we pick the best-performing variant when reporting results (Appendix~\ref{si-sec:score-head-analysis}).

TipPFN and TabPFN receive the same context episodes and query window.
For each query trajectory, they predict the (transformed) RDTC $\Lambda^*$ over the observed window and future time points; see Fig.~\ref{fig:tippfn-prediction}.
TabPFN is cast as tabular regression by treating trajectory time points as rows and context episodes as in-context training examples.
Predicted RDTC distributions are then converted into tipping-risk scores, e.g. $1-\mathrm{median}(\Lambda)$ or $\mathrm{CDF}(\Lambda < 0.05)$ at the final nowcast or forecast time point, which rank queries by predicted risk.
For TabPFN and TipPFN, we report the final forecast-time $1-\mathrm{median}(\Lambda)$ score for all methods.
As illustrated in Fig.~\ref{fig:tippfn-prediction}b, this score is the parameter-free estimate of forecast remaining RDTC at the forecast horizon, and therefore aligns directly with the early-warning task.
Note that the forecast is always $192 - W$ steps into the future, so shorter windows make a longer-horizon forecast, thus also increasing uncertainty.
We report score sensitivity analyses in Table~\ref{tab:auroc-score-head-summary} and Appendix~\ref{si-sec:score-head-analysis}.

\begin{table}[b]
\centering
\caption{
    \textbf{Average pre-tip AUROC by system and dataset.}
    Balanced AUROC for detecting critical episodes, averaged over positive lead times ($\Delta > 0$).
    Each column uses one fixed evaluation setting across all listed rows that yields highest AUROC across all these systems (see Table~\ref{tab:auroc-column-choices} for details).
    The last row leaves out the three canonical systems, which share their bifurcation types with the training data.
}
\label{tab:auroc-results-best-col}
\renewcommand{\arraystretch}{0.88}
\resizebox{\columnwidth}{!}{%
\begin{tabular}{lccccccccc}
\toprule
Dataset & \shortstack[c]{\strut TipPFN\\\strut 0c} & \shortstack[c]{\strut TipPFN\\\strut 1c} & \shortstack[c]{\strut TipPFN\\\strut 2c} & \shortstack[c]{\strut TabPFN\\\strut 1c} & \shortstack[c]{\strut TabPFN\\\strut 2c} & \shortstack[c]{\strut EWS\\\strut $\tau_{\mathrm{var}}$} & \shortstack[c]{\strut Bury~\cite{bury2021deep}\\\strut $P_{\mathrm{tip}}$} & \shortstack[c]{\strut Huang~\cite{huang2024deep}\\\strut $P_{\mathrm{tip}}$} & \shortstack[c]{\strut Zhuge~\cite{zhuge2025deep}\\\strut $\Delta_{\mathrm{tip}}^{\pm}$} \\
\midrule
\multicolumn{10}{l}{\emph{Canonical}} \\
B-Fold & .876 & .929 & \textbf{.955} & .574 & .786 & .633 & .697 & .462 & .522 \\
B-Hopf & .861 & .935 & \textbf{.952} & .640 & .820 & .635 & .842 & .496 & .665 \\
B-Trans. & .864 & .925 & \textbf{.950} & .581 & .807 & .543 & .722 & .453 & .458 \\
\midrule
\multicolumn{10}{l}{\emph{Semi-real}} \\
B-Harv. & .846 & .958 & \textbf{.974} & .685 & .810 & .914 & .967 & .873 & .734 \\
B-RM TC & .815 & .951 & \textbf{.963} & .494 & .741 & .604 & .784 & .604 & .570 \\
B-RM Hopf & .769 & .829 & \textbf{.904} & .499 & .689 & .557 & .638 & .484 & .422 \\
B-SEIRx & .582 & .822 & \textbf{.905} & .495 & .685 & .498 & .468 & .507 & .381 \\
B-AMOC & .853 & .898 & .940 & .616 & .825 & .925 & \textbf{.948} & .509 & .679 \\
R-Bautin & .308 & .543 & \textbf{.700} & .404 & .489 & .451 & .308 & .538 & .180 \\
R-SN & .644 & .802 & \textbf{.892} & .335 & .569 & .440 & .354 & .273 & .080 \\
R-Compost & \textbf{.767} & .623 & .665 & .472 & .493 & .481 & .727 & .678 & .506 \\
R-AMOC & .257 & .782 & \textbf{.913} & .521 & .654 & .270 & .102 & .712 & .112 \\
\midrule
\multicolumn{10}{l}{\emph{Real-world \& Sim-to-Real}} \\
SWEC-iEEG & .444 & .512 & .506 & .489 & .506 & .524 & \textbf{.541} & .488 & -- \\
TAC & .532 & .508 & .517 & .530 & .477 & \textbf{.565} & .476 & .522 & -- \\
DaphniaExt & .500 & .717 & \textbf{.832} & .511 & .651 & .540 & .483 & .404 & -- \\
\midrule
\emph{All datasets} & .661 & .782 & \textbf{.838} & .523 & .667 & .572 & .604 & .534 & .443 \\
\emph{All except canonical} & .610 & .745 & \textbf{.809} & .504 & .633 & .564 & .566 & .549 & .407 \\
\bottomrule
\end{tabular}
}
\end{table}

\subsection{Generalization across synthetic transition systems}

We start with the canonical bifurcation setting, evaluating on held-out realizations from the fold, Hopf, and transcritical families used to define the training prior. TipPFN already performs strongly without additional context trajectories, with robust AUROC across all three systems and clear gains over classical early-warning baselines (Table~\ref{tab:auroc-results-best-col}, Fig.~\ref{fig:auroc-all-colwise-best}).

Importantly, the binary task is stricter than a simple critical-versus-noncritical split: negative examples include not only equilibrium trajectories, but also trajectories that remain subcritical while approaching or receding from criticality.
Strong performance therefore indicates that TipPFN learns reusable dynamical signatures rather than memorizing individual trajectories.

\paragraph{Beyond canonical bifurcations}
We next test generalization.
On semi-real reduced-order systems, TipPFN typically matches or exceeds the evaluated baselines, showing transfer from simple synthetic bifurcations to more realistic unseen dynamics (Table~\ref{tab:auroc-results-best-col}; Fig.~\ref{fig:auroc-all-colwise-best}).
A stronger test is r-tipping, which lies outside the training mechanisms altogether. Despite this mismatch, TipPFN remains competitive on Bautin, compost-bomb, saddle-node, and AMOC r-tipping systems.

These results use one fixed configuration per method, selected for strong aggregate performance across all systems.
Under a more permissive best-per-system comparison, several baselines improve on individual tasks, but their gains are often system-specific (Table~\ref{tab:auroc-results-best-per-system}).
TipPFN remains best-performing on many datasets, and where it is outperformed, it typically is close to the best baseline.

\subsection{Leveraging context}\label{sec:context}
\begin{figure}[t]
    \centering
    \includegraphics[width=1\linewidth]{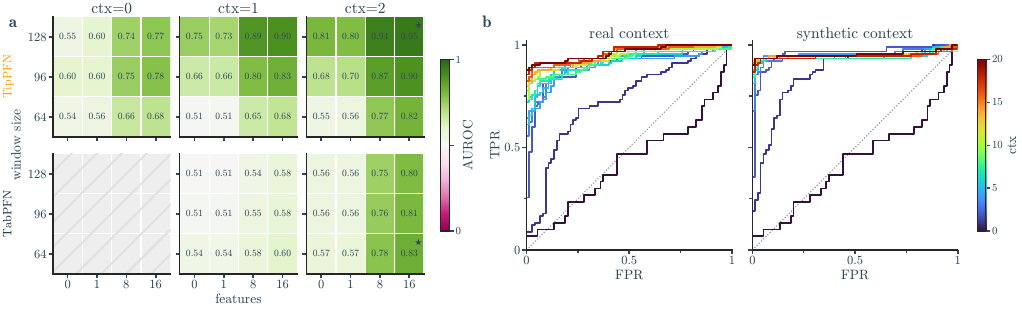}
    \caption{\textbf{Context matters.}
        \textbf{(a)} Multi-parameter sweep over number of context episodes, observed feature channels, and window size $W$ for TipPFN and TabPFN.
        Color denotes the AUROC of the RDTC nowcast on the semi-real b-tipping systems of Table~\ref{tab:auroc-results-best-col}, pooled over systems and averaged over all eight lead times $\Delta \in [-10, 80]$; stars mark the best configurations.
        Note that TabPFN cannot be used in the zero-context setting.
        \textbf{(b)} TipPFN ROC on \texttt{DaphniaExt} for $\Delta = 2,~W=16$ and varying number of real or simulated context episodes.
    }
    \label{fig:auroc-results-ctx-sweep}
\end{figure}

A central motivation for TipPFN is that context provides system-specific information at inference time.
This is particularly valuable for tipping prediction, where observations are often short, noisy, and only weakly informative on their own.
Rather than reducing trajectories to one-dimensional early-warning statistics, TipPFN conditions on full multivariate context episodes and can therefore infer which feature patterns distinguish tipping from non-tipping behavior.

Figure~\ref{fig:auroc-results-ctx-sweep}a shows that TipPFN benefits most from combining additional context, higher-dimensional observations, and sufficiently long windows.
For both TipPFN and TabPFN, performance improves when context episodes are added even under a fixed input budget, suggesting that the gain comes not only from observing the query longer, but from comparing it against a broader range of system behavior.
Increasing the number of observed features yields a further boost, highlighting the value of multivariate inputs.

\subsection{Real-world transfer}

We next evaluate TipPFN on sim-to-real and real-world datasets, where true RDTC labels are typically unavailable.
We therefore construct surrogate RDTC targets, $\Lambda = \Lambda(t)$, from experimentally observed or independently estimated transition markers, and test whether TipPFN can generalize using limited empirical context or, where available, matched simulations.

On \texttt{DaphniaExt}, we analyse TipPFN on extremely sparse ($W=16$) real observations and matched simulations~\cite{ricker1954logistic}.
As shown in Fig.~\ref{fig:auroc-results-ctx-sweep}b, simulated and observed context reach comparable performance but respond differently to context size.
Adding observed episodes keeps improving performance and simulated context provides even better results, compensating for sparse observations.
Furthermore, while TabPFN and TipPFN both perform well on predicting the extinction event, only TipPFN discerns the underlying bifurcation~\cite{drake2010early}, see Figs.~\ref{fig:daphnia-forcing-schedule}, \ref{fig:auroc-daphnia-rdtc-choice} and Appendix~\ref{si-sec:daphnia-results}.

We further evaluate TipPFN on a noisy experimental system with a stochastic subcritical Hopf transition (\texttt{TAC}~\cite{bonciolini2018experiments}, surrogate RDTC from the known critical point) and on seizure onset in multichannel iEEG bandpower trajectories (\texttt{SWEC-iEEG}).
In the standard protocol, all methods stay close to chance on both (Table~\ref{tab:auroc-results-best-col}).
While the Hopf-specific Bury baseline performs strongly on \texttt{TAC}, TipPFN improves substantially with real and simulated context, especially when the supplied episodes are themselves critical (see Fig.~\ref{fig:auroc-tac-cnc}, Appendix~\ref{si-sec:tac-results}). On \texttt{SWEC-iEEG}, no method separates pre-ictal from interictal segments (Appendix~\ref{si-sec:swec-results}).

\begin{figure}
    \centering
    \includegraphics[width=1\linewidth]{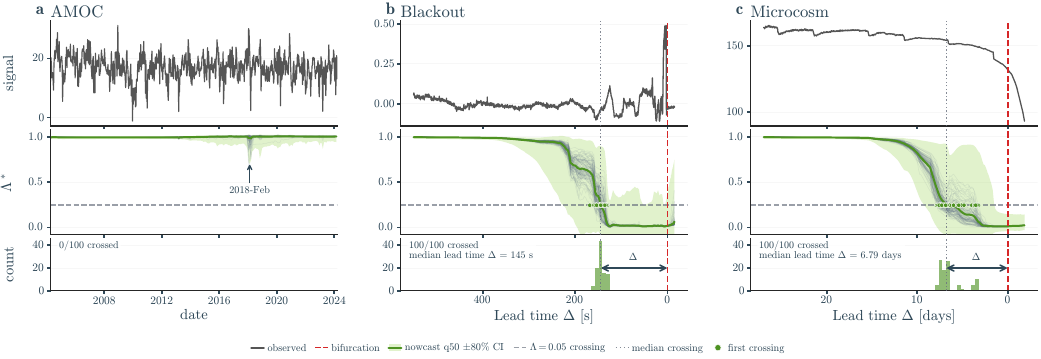}
    \caption{
        \textbf{Zero-shot TipPFN RDTC nowcasts} on three univariate real-world time series, without per-system fine-tuning.
        Rows: observed signal; TipPFN predictions of $\Lambda^*$ (means of the predicted median and 10--90\% quantiles over 100 ensemble members with randomly retained time points); distribution of the members' first crossings of $\Lambda^*_\mathrm{thrs} = \tanh(5\Lambda_\mathrm{thrs}) \approx 0.245$.
        Red dashed lines mark the bifurcation time from dominant-eigenvalue analysis~\cite{grziwotz2023anticipating}, where available.
    }
    \label{fig:zero-shot-selected}
\end{figure}

\paragraph{Zero-shot prediction on real-world systems} We also evaluate TipPFN in a strict zero-shot setting, without context trajectories.
The relevant output is the predicted RDTC trajectory, a time-resolved nowcast of distance to criticality from a single observed time series.
This allows critical-point timing to be estimated even when no matched empirical or simulated context is available.

Across six real-world observational time series (see Fig.~\ref{fig:zero-shot-selected}, Appendix~\ref{fig:zero-shot-si}), TipPFN's RDTC predictions are informative:
In the \texttt{AMOC} example, we detect an increase in RDTC variance coinciding with an abnormal reversal in AMOC slow-down~\cite{lee2024pause}.
In a \texttt{power-grid} time series preceding a blackout, TipPFN signals hidden instability 2~minutes before the failure becomes apparent.
In the \texttt{microcosm} series, it estimates a cyanobacteria population collapse approximately a week from the actual event.
Because visible regime shifts may lag behind hidden loss of stability, zero-shot RDTC prediction can provide advance warning and create a window for mitigation or response.

\subsection{Ablations and sensitivity analysis}\label{sec:ablations}
To isolate what drives TipPFN's performance, we run ablations on 12 simulated systems, varying the training target, model size, and architecture (Appendix~\ref{si-sec:ablations}). 

\paragraph{TipPFN setup} Performance gains come from RDTC as target combined with in-context learning, which improves AUROC on most systems (Sec.~\ref{sec:context}). Replacing RDTC with a single tipping label per episode (a classifier with the same architecture) or with a dense binary target reduces AUROC.

\paragraph{Baseline parity} Prior and model scale do not account for the performance gap as the CNN-LSTM of~\cite{bury2021deep}, retrained on our prior, remains below TipPFN, and a model with 140k rather than 210M parameters still reaches 0.857 AUROC, against 0.893 for the full model (Table~\ref{tab:si-abl-scaling}).

\paragraph{Data and context availability} Training is also data-efficient, as a smaller variant reaches 90\% of its final AUROC after 8\% of one epoch, although this does not disentangle data volume from the number of optimisation steps. TipPFN is sensitive to the source of its context but robust to its noise levels. Context taken from a different draw of the same system lowers AUROC, whereas adding noise of one standard deviation to the context has less of an impact (Tables~\ref{tab:si-abl-context} and~\ref{tab:si-abl-sample-eff}). TipPFN's advantage is also not a result of the shared 128-step window. Even at each baseline's native input length, TipPFN still leads on 6 of 8 canonical and b-tipping systems (Appendix~\ref{si-sec:native-window}).

\section{Conclusion}\label{sec:conclusion}

We introduce TipPFN, a transformer-based prior-data fitted network designed to predict the relative distance to criticality (RDTC) directly from observed trajectories.
TipPFN is trained exclusively on a novel synthetic data-generation framework combining canonical tipping systems with randomized high-dimensional nonlinear stochastic dynamics, enabling prediction across diverse transition regimes.
We evaluate TipPFN against state-of-the-art machine learning and dynamical systems baselines on more than 20 datasets spanning synthetic, sim-to-real, and real-world settings across multiple domains, noise levels, and tipping mechanisms.
TipPFN outperforms existing approaches on most simulated systems while demonstrating strong generalization across previously unseen systems.

Unlike existing tipping-predicting methods, TipPFN can leverage additional context trajectories, auxiliary features, and variable observation windows through ICL, while remaining effective even in the absence of context information. %
The framework naturally integrates real observations, simulated trajectories, or hybrid combinations thereof, enabling prediction in settings where observational data are sparse or expensive~\cite{nathaniel2023metaflux,kim2024spatiotemporal}.

By inferring hidden proximity to a transition rather than relying on explicit early warning indicators, TipPFN provides a universal, system-agnostic estimate of criticality without requiring system-specific retraining or manual selection of diagnostic metrics.
Finally, TipPFN consistently and accurately provides early prediction of tipping characteristics, enabling real-world deployment where early intervention is critical.

\paragraph{Limitations}

On our synthetic benchmarks, the auxiliary variables respond to both the state variables and the forcing, so part of TipPFN’s advantage comes from observed variables that carry information about the forcing level. Such knowledge of the forcing, direct or indirect, is common in practice especially when detecting R-tipping, and future work will examine how it helps in real systems where the forcing is known.

Beyond this, we plan to combine ICL with near-real-time adaptation for online settings, for example through low-rank updates~\cite{hu2022lora} or test-time training~\cite{sun2020test,tandon2025end}, so that learned priors evolve as new observations arrive, and to address the spatial dependence of critical transitions in climate~\cite{romanou2023stochastic,nathaniel2024chaosbench} and ecological systems~\cite{guttal2008changing}.

\section*{Acknowledgements and disclosures}
JN, HF, PG acknowledge funding, computing, and storage resources from the NSF Science and Technology Center (STC) Learning the Earth with Artificial Intelligence and Physics (LEAP) (Award \#2019625) and by NASA under award No 80NSSC25K0062. We thank Maximilian Hüttenrauch for his contributions to this work.
Data from the RAPID AMOC observing project is funded by the Natural Environment Research Council, U.S. National Science Foundation (NSF) with support from NOAA. They are freely available from \url{https://rapid.ac.uk/}.
YS, KU, VL, TW, GR, JH and BH are affiliated with kausable, a company developing machine-learning methods for complex dynamical systems. The authors declare no other competing interests.

\bibliography{neurips_2026}

\clearpage
\newpage
\appendix

\etocsettagdepth{mtoc}{none}        %
\etocsettagdepth{atoc}{subsection}  %
\etocsettocstyle{\section*{Appendix contents}}{\bigskip}
\tableofcontents
\etocdepthtag.toc{atoc}

\renewcommand{\theequation}{S\arabic{equation}}
\renewcommand{\thefigure}{S\arabic{figure}}
\renewcommand{\thetable}{S\arabic{table}}

\setcounter{equation}{0}
\setcounter{figure}{0}
\setcounter{table}{0}

\section{Additional background}\label{si-sec:add-background}

\textbf{Critical transitions. }
As discussed in the main text, tipping phenomena can be broadly classified into three distinct types depending on the mechanisms driving state transitions, namely bifurcation-induced tipping (b-tipping), noise-induced tipping (n-tipping), and rate-induced tipping (r-tipping)~\cite{ashwin2012tipping}. Refer to Fig.~\ref{fig:overview}a for illustration on the different mechanisms underlying each one of them.

\textbf{Bifurcation-induced tipping and critical slowing down. }
In b-tipping, critical transitions arise from local bifurcations, where an equilibrium $\mathbf{x}^\ast(\lambda)$ loses stability as parameters vary. Linearizing the dynamics around $\mathbf{x}^\ast(\lambda)$ yields
\begin{equation}
    \frac{d \boldsymbol{\epsilon}}{dt} = J(\lambda)\,\boldsymbol{\epsilon},
\end{equation}
where $\boldsymbol{\epsilon} = \mathbf{x} - \mathbf{x}^\ast(\lambda)$ and the Jacobian matrix is
\begin{equation}
    J_{ij}(\lambda) = \left.\frac{\partial f_i(\mathbf{x}, \lambda)}{\partial x_j}\right|_{\mathbf{x}=\mathbf{x}^\ast(\lambda)}.
\end{equation}

A bifurcation occurs when the linear stability of the equilibrium changes, typically because one or more eigenvalues $\mu_i(\lambda)$ of $J(\lambda)$ cross the stability boundary in the complex plane, i.e.,
\begin{equation}
    \max_i \mathrm{Re}(\mu_i(\lambda_c)) = 0.
\end{equation}
The manner in which eigenvalues approach this boundary constrains the bifurcation type: a real eigenvalue crossing zero is associated with steady-state bifurcations (e.g., fold, transcritical), whereas a complex-conjugate pair crossing the imaginary axis is associated with a Hopf bifurcation. In discrete-time systems, the corresponding stability boundary is the unit circle.

As the system approaches such a bifurcation, perturbations decay increasingly slowly, a phenomenon known CSD~\cite{fang2025tipping,dakos2008slowing}. In particular, as $\lambda \to \lambda_\mathrm{crit}$, the leading eigenvalue satisfies $\mathrm{Re}(\mu_i(\lambda)) \to 0$, implying a vanishing local contraction rate and increasingly slow recovery from perturbations.

\textbf{Rate-induced tipping. }
In contrast to b-tipping, r-tipping arises from the inability of the system to track a moving attracting state under time-dependent forcing~\cite{ashwin2012tipping}. Consider again the nonautonomous system with control parameter $\lambda(t)$. For each fixed $\lambda$, assume the associated frozen system admits a locally stable equilibrium $\mathbf{x}^\ast(\lambda)$ satisfying
\begin{equation}
    f(\mathbf{x}^\ast(\lambda), \lambda) = 0,
    \qquad
    \max_i \mathrm{Re}(\mu_i(\lambda)) < 0.
\end{equation}

Defining the tracking error
\begin{equation}
    \boldsymbol{\epsilon}(t) := \mathbf{x}(t) - \mathbf{x}^\ast(\lambda(t)),
\end{equation}
and linearizing around $\mathbf{x}^\ast(\lambda(t))$ yields
\begin{equation}
    \frac{d \boldsymbol{\epsilon}}{dt}
    =
    J(\lambda(t))\,\boldsymbol{\epsilon}
    +
    \underbrace{\frac{d}{dt}\mathbf{x}^\ast(\lambda(t))}_{\text{branch drift}}
    +
    \text{higher-order and stochastic terms}.
\end{equation}

The additional term $\frac{d}{dt}\mathbf{x}^\ast(\lambda(t))$ represents the motion of the equilibrium induced by the changing control parameter. R-tipping occurs when this control-induced drift is sufficiently large relative to the local contraction governed by $J(\lambda(t))$, so that the trajectory can no longer track the moving equilibrium, even though the instantaneous dynamics remain locally stable. Unlike in b-tipping, no eigenvalue needs to cross the stability boundary, and therefore classical indicators based on CSD may fail to provide reliable early warning signals.

\textbf{Noise-induced tipping. }
In n-tipping, transitions occur through stochastic perturbations that induce escape from the basin of attraction of a stable equilibrium, even when $\mathrm{Re}(\mu_i(\lambda)) < 0$. This mechanism can act independently or interact with both b-tipping and r-tipping in stochastic systems.

\textbf{Early warning signals. }
CSD manifests in observable time series statistics~\cite{dakos2012robustness,carpenter2006rising}. For a discrete trajectory $\{x_t\}_{t=1}^T$, the \textit{lag-1 autocorrelation} (AR1) is defined as
\begin{equation}
\mathrm{AR1} =
\frac{\sum_{t=2}^{T} (x_t - \bar{x})(x_{t-1} - \bar{x})}
{\sum_{t=1}^{T} (x_t - \bar{x})^2},
\end{equation}
where $\bar{x} := \frac{1}{T}\sum_{t=1}^T x_t$ denotes the empirical mean over the time window.

To connect AR1 and variance to local stability, consider the scalar linearized stochastic dynamics
\begin{equation}
x_{t+1}-\bar{x} = a(x_t-\bar{x}) + \sigma \xi_t,
\end{equation}
where $\{\xi_t\}$ is an i.i.d.\ zero-mean stochastic process and $|a|<1$ ensures stability. Under stationarity and independence of the innovations $\sigma\xi_t$, $\mathrm{AR1} \approx a$. If the system arises from a continuous-time linearization with leading eigenvalue $\mu<0$ and sampling interval $\Delta t$, then $a = e^{\mu \Delta t}$, so $\mathrm{AR1}\to 1$ as $\mu \to 0^-$. 

Similarly, the stationary variance satisfies
\begin{equation}
    \mathrm{VAR}(x) = \frac{\sigma^2}{1-a^2},
\end{equation}
which increases as $a\to 1$. 

A detailed derivation of the relationship between AR1, variance, and the underlying spectral properties of the linearized dynamics is provided in e.g., \cite{grziwotz2023anticipating,miyauchi2026generalized}.

These signatures, rising AR1 and variance, form the basis of classical EWS for b-tipping. However, these indicators rely on local linearization near equilibrium and are therefore intrinsically tied to eigenvalue dynamics. In n-tipping, transitions occur due to stochastic escape from a basin of attraction even when $\mathrm{Re}(\mu_i)<0$, while in r-tipping, rapid parameter changes prevent the system from tracking its quasi-static equilibrium without any eigenvalue crossing. In both cases, critical slowing down may be weak or absent.

While bifurcation theory provides a principled framework for characterizing tipping behaviour, its practical application is limited. Identifying tipping points requires access to the governing equations, parameters, and equilibrium states needed to compute the Jacobian and its eigenvalues. In real-world systems, these quantities are typically unknown or only partially observed, motivating data-driven approaches that infer tipping behaviour directly from observed trajectories without requiring explicit knowledge of the underlying dynamical system.

\clearpage
\newpage
\section{Prior-Data Fitted Networks}\label{si-sec:pfns}

\begin{figure}[h]
    \centering
    \includegraphics[width=\textwidth]{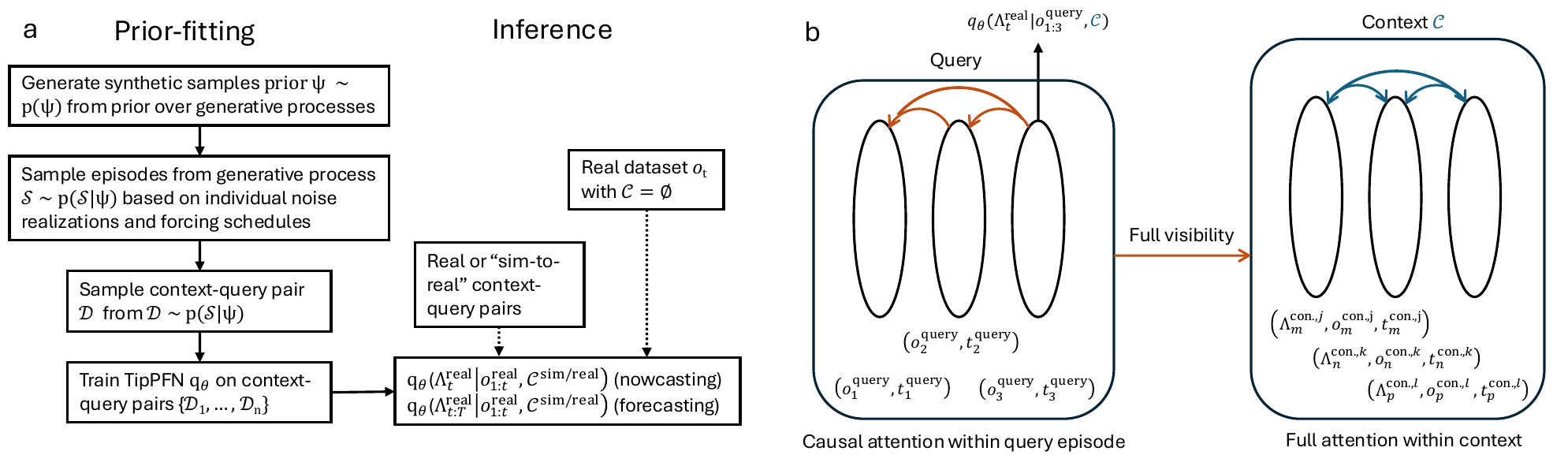}
    \caption{(a) Prior-data fitting and inference and (b) attention structure of TipPFN. Schematics based on~\cite{muller2022transformers}.}
    \label{fig:metrics}
\end{figure}

\subsection{Driver variables}\label{si-sec:forcing_datasets}
Following \cite{bury2021deep}, the driver $M$ is generated from randomly constructed two-dimensional dynamical systems of the form

\begin{equation}
    \dot{x} = \sum_{i=1}^{10} a_i p_i(x,y), \qquad
    \dot{y} = \sum_{i=1}^{10} b_i p_i(x,y),
\end{equation}

where $(x,y) \in \mathbb{R}^2$ and $\{p_i(x,y)\}_{i=1}^{10}$ denotes the set of all monomials up to third order:

\begin{equation}
    p(x,y) = (1, x, y, x^2, xy, y^2, x^3, x^2y, xy^2, y^3).
\end{equation}

The coefficients $a_i, b_i$ are independently sampled from $\mathcal{N}(0,1)$, after which a random subset (50\%) is set to zero to induce sparsity. To encourage bounded trajectories, coefficients associated with cubic terms are constrained to be negative. We then perform integration for $10^4$ timesteps and $10^{-2}$ discretization to select trajectories that converge to an equilibrium, defined as the final 10 points with difference of less than $10^{-8}$. We then use AUTO-07P program~\cite{doedel2007auto} to identify bifurcation types and their corresponding critical values along the identified equilibrium branch as each nonzero parameter is varied within the interval $[-5,5]$.
Table~\ref{tab:driver-hparams} lists the relevant driver-generation hyperparameters.

\begin{table}[h]
\centering
\small
\caption{\textbf{Hyperparameters of the synthetic polynomial driver.}}
\begin{tabularx}{\linewidth}{lX}
\toprule
Quantity & Value or distribution \\
\midrule
Polynomial degree & All monomials up to degree 3; 10 coefficients per equation, 20 total. \\
Coefficient prior & $a_j,b_j\sim\mathcal{N}(0,1)$ before sparsification. \\
Coefficient sparsity & Exactly 50\% of the 20 coefficients are set to zero uniformly at random. \\
Boundedness bias & Cubic coefficients $x^3,x^2y,xy^2,y^3$ in both equations are replaced by their negative absolute values. \\
Initial condition for equilibrium screening & $z_0\sim\mathcal{N}(0,4I_2)$. \\
Equilibrium screening integration & Forward Euler with $\Delta t=0.01$ for up to 100 time units. \\
Convergence criterion & Norm of the range of the final 10 states below $10^{-8}$ and maximum trajectory amplitude below $10^3$. \\
Stability criterion & All Jacobian eigenvalues at the equilibrium have negative real part. \\
Recovery rate & $r_M=|\max_i \operatorname{Re}\mu_i|$ at the stable equilibrium. \\
Model retry budget & Up to 100 random polynomial models per TipBox attempt. \\
Bifurcation retry budget & Up to 50 TipBox attempts per raw sample; AUTO output suppressed. \\
\bottomrule
\end{tabularx}
\label{tab:driver-hparams}
\end{table}

For a selected bifurcation point, let $p_0$ be the initial coefficient value and
$p_{\mathrm{crit}}$ the AUTO-07P bifurcation value. The schedule
$\tilde{\lambda}_k(t)$ is applied as
\begin{equation}
    \lambda_{M,k}(t)=p_0+\tilde{\lambda}_k(t)\bigl(p_{\mathrm{crit}}-p_0\bigr).
    \label{eq:driver-param-schedule}
\end{equation}
The RDTC target is
\begin{equation}
    \Lambda_k(t)=1-\tilde{\lambda}_k(t).
    \label{eq:synthetic-rdtc-target}
\end{equation}
Thus $\Lambda=0$ marks the bifurcation point, $\Lambda>0$ is subcritical, and
$\Lambda<0$ means the schedule has passed the detected critical value.

\paragraph{Driver SDE simulation}
Given $M$ and a forcing schedule, TipBox simulates the stochastic driver with
Euler--Maruyama at integration step $\Delta t=0.01$ and output sampling interval
$1.0$. A 100-time-unit burn-in is run at the initial parameter before the recorded
trajectory starts. The additive driver noise scale is
\begin{equation}
    \sigma_M = \sqrt{2r_M}\,\sigma_{\mathrm{tilde}}\,\xi,
    \qquad
    \sigma_{\mathrm{tilde}}=0.01,\quad
    \xi\sim\operatorname{Triangular}(0.75,1.0,1.25),
\end{equation}
and Brownian increments are scaled by $\sigma_M\sqrt{\Delta t}$.

\subsection{Training Dataset Forcing Schedule}
\label{si-sec:synthetic-forcing-schedules}

\begin{figure}[h]
    \centering
    \includegraphics[width=1\linewidth]{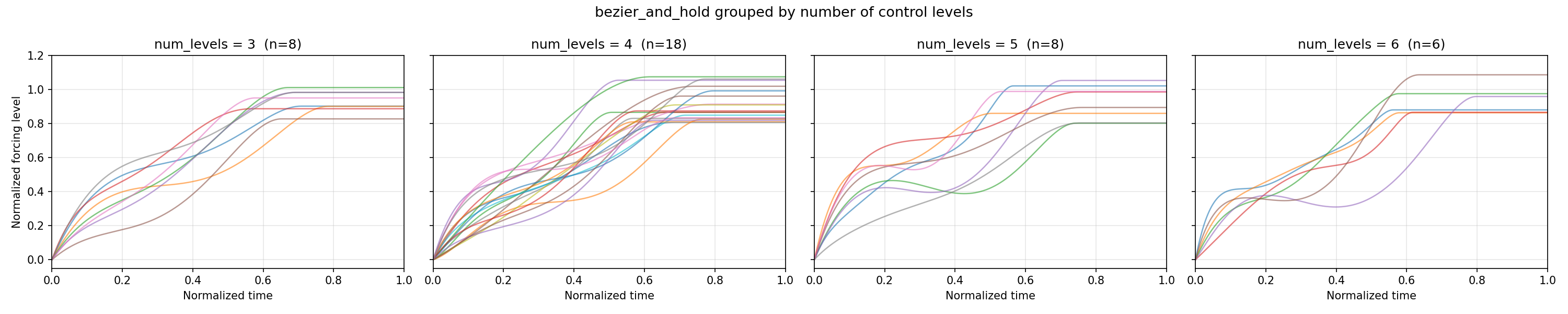}
    \caption{Bezier-type forcing schedules used in generation of the training dataset.}
    \label{fig:bezier-forcings}
\end{figure}

For the training dataset, forcing schedules are sampled independently per episode, while the bifurcation system $M$ is shared across the $K=4$ episodes of a raw sample.
Table~\ref{tab:v2-forcing} summarizes the full schedule prior.
All schedules start at $\tilde{\lambda}=0$.
Examples of bezier-type schedules are shown in Fig.~\ref{fig:bezier-forcings}.

Note that this partially non-linear and beyond-criticality forcing is unlike that used in validation, see Section~\ref{si-sec:forcing}.
During training, it is causally permissible to also show non-linear forcings, through which the model may learn to recognise a wider range of dynamical system responses to changes in forcing.

\begin{table}[h]
\centering
\caption{\textbf{Forcing schedule prior}. Final levels below 1 create subcritical approaches, level 1 reaches the
bifurcation, and levels above 1 pass the bifurcation.}
\label{tab:v2-forcing}
\small
\begin{tabularx}{\linewidth}{lclX}
\toprule
Schedule & Probability & Parameters & Definition and effect \\
\midrule
\texttt{bezier\_and\_hold} & 0.4 &
\begin{tabular}[t]{@{}l@{}}
\texttt{start}=0\\
\texttt{smooth\_hold}=true\\
$h\sim U(0.6,0.8)$\\
$L=1$ w.p. $1/2$; else $L\sim U(0.6,1.2)$\\
$n_c\in\{3,4,5,6\}$ with ratios $2:3:2:1$\\
control range $[0.1L,0.9L]$
\end{tabular}
&
Bezier ramp from 0 to final level $L$ by fraction $h$, followed by a hold at
$L$. Interior control-point values are clipped to $[0,L]$; duplicating the final
control point makes the join to the hold phase approximately smooth. \\
\midrule
\texttt{ramp\_and\_hold} & 0.4 &
\begin{tabular}[t]{@{}l@{}}
$h\sim U(0.6,0.8)$\\
$L=1$ w.p. $1/2$; else $L\sim U(0.6,1.2)$
\end{tabular}
&
Linear ramp from 0 to $L$ until fraction $h$, followed by a constant hold at
$L$. \\
\midrule
\texttt{constant} & 0.2 &
$\tilde{\lambda}(t)=0$ &
Stationary no-forcing baseline at the initial stable equilibrium, with
$\Lambda(t)=1$. \\
\bottomrule
\end{tabularx}
\end{table}

\subsection{Augmenting SDE System}
\label{si-sec:synthetic-augmenting-sde}

The two-dimensional driver is embedded into a higher-dimensional stochastic
system so that TipPFN does not only see canonical normal-form trajectories. 

The auxiliary variables $u_k(t)=(u_{1,k}(t),\ldots,u_{14,k}(t))$ solve an Ito SDE
with diagonal diffusion,
\begin{equation}
    du_{i,k}(t)
    =
    \left[
        (1-\eta_i) f_i\!\left(r_{i,k}(t)\right)
        - \gamma_i u_{i,k}(t)
    \right]dt
    + \sigma_i\,dW_{i,k}(t),
    \label{eq:auxiliary-sde}
\end{equation}
where $r_{i,k}(t)$ concatenates the parent auxiliary variables and external-input
channels selected by $\mathcal{G}$ for variable $i$. The functions $f_i$ are
sampled MLPs, one per dynamic variable. The same graph, MLPs, and SDE parameters
are shared across the episodes of a raw sample, while initial states,
Brownian noise, TipBox noise, and forcing schedules vary by episode.

\paragraph{Directed graph sampler}
The graph $\mathcal{G}$, which is structuring the interactions within the variables of the auxiliary system~\cite{herdeanu2026_causal_dynamics}, is generated by a scale-free graph model based on bi-partite graphs~\cite{birmele2009scaleFreeBipartite}: For each raw sample, the code first samples a directed bipartite graph between variable nodes $V$ and interaction nodes $H$, with $|V|=14$ and $|H|=8$.
For every variable $i$, the desired outgoing and incoming interaction degrees
$D_i^{+}$ and $D_i^{-}$ are sampled independently from the truncated power law
\[
    \Pr(D=d)=\frac{d^{-2}}{\sum_{\ell=1}^{8}\ell^{-2}},
    \qquad d\in\{1,\ldots,8\}.
\]
Conditional on these sampled degrees, each
candidate edge $i\to h$ is included with probability $D_i^{+}/8$, and each
candidate edge $h\to i$ with probability $D_i^{-}/8$, independently across
interaction nodes. The bipartite graph is then projected to a directed variable
graph: an edge $i\to j$ is added whenever there exists an interaction node
$h$ such that $i\to h$ and $h\to j$. Multiple such paths collapse to one
directed edge, and self-loops are retained.
After this projection, four external input channels are added in the training set and in all synthetic evaluation sets: the two state variables of the canonical driver $M$, its forcing parameter $\lambda$ and the RDTC $\Lambda=1-\tilde\lambda$.
The auxiliary variables therefore carry a noisy, filtered readout of the forcing level; Appendix~\ref{si-sec:aux-drive-sensitivity} reports results on evaluation sets in which they are driven by the state variables only. Each external-input node connects to
each auxiliary variable independently with probability $p_{\mathrm{inp}}=0.8$.
MLP hyperparameters are given in table~\ref{tab:sde-parameter-hparams}

\begin{table}[h]
\centering
\small
\caption{\textbf{Parameter hyperparameters for the auxiliary SDE.}
These are sampled once for the shared SDE system inside a raw sample.}
\begin{tabularx}{\linewidth}{lX}
\toprule
Quantity in Eq.~\ref{eq:auxiliary-sde} & Value or distribution \\
\midrule
MLP depth & 2 linear layers \\
Hidden width & 32 hidden units \\
Activation & \texttt{tanh}\\
Weight and bias initialization scale & For each MLP, $s_w=1.0+\operatorname{HalfCauchy}(1.0)$ resampled until $s_w\le 1.5$; weights and biases use $\mathcal{N}(0,s_w^2)$. \\
Hidden-layer sparsity & For each MLP, $s_p=0.5-\operatorname{HalfCauchy}(0.5)$ resampled until $s_p\ge 0.1$; hence $s_p\in[0.1,0.5]$. \\
Sparsity application & Hidden-layer weights only; retained weights are scaled by $(1-s_p)^{-1/2}$. With 2 layers, there is no interior hidden-to-hidden layer, so this setting is effectively inactive for variable MLPs. \\
Preactivation noise & 0.0. \\
Self-loop coefficient & $\gamma_i=20\,|\mathcal{N}(0,1)|$. \\
Friction coefficient & $\eta_i=0.5\,|\mathcal{N}(0,1)|$; the code multiplies the MLP flow by $1-\eta_i$. \\
Diffusion coefficient & $\sigma_i=0.05\,|\mathcal{N}(0,1)|$ with independent diagonal noise. \\
\bottomrule
\end{tabularx}
\label{tab:sde-parameter-hparams}
\end{table}

\subsection{Task Construction for TipPFN}
\label{si-sec:synthetic-task-construction}

Each raw sample is transformed into a masked tabular time-series task.

\paragraph{Row selection by resampling} 
A subset of time steps is selected based on stratified jitter, to yield query time-series with lengths between 92 and 256 time steps, constrained to a total context size of 320 time steps or less, to reserve 192 of the 512 step budget for the query trajectory and potential forecasting time steps, with indices $1:T^*$ referring to the resampled time steps.
Using $\Lambda_{1:T^*} = 1 - \tilde{\lambda}_{1:T^*}$, this yields the query-context pair $\mathcal{D} = ((\Lambda_{1:T^*}, \mathbf{o}_{1:T^*}) \cup \mathcal{C})$ with prior distribution $p(\mathcal{D})$ (see Fig.~\ref{fig:tippfn-prediction}), hyperparameters are given in Table~\ref{tab:partition-hparams}.

\begin{table}[h]
\centering
\small
\caption{\textbf{Episode partitioning and temporal subsampling.}}
\begin{tabularx}{\linewidth}{lX}
\toprule
Quantity & Value or distribution \\
\midrule
Maximum sequence length & 512 rows. \\
Query episodes & Always 1 episode. \\
Context episodes & $N_{\mathrm{ctx}}\in\{0,1,2,3\}$ with probabilities $(0.2,0.3,0.3,0.2)$. \\
Episode selection & Deterministic-random selection; query and context episodes do not overlap. \\
Context length budget & $N_{\mathrm{ctx}}=0:[0,0]$, $1:[92,256]$, $2:[184,320]$, $3:[276,320]$ total rows. \\
Context per-episode bounds & Minimum 92 rows, maximum 256 rows. \\
Query length budget & Target 192 rows, with per-episode bounds $[1,256]$. \\
Temporal subsampling & Stratified jitter with endpoints retained. \\
Overflow handling & Shrink context; use available episodes if insufficient; error on empty query. \\
Identifier normalization & Episode ID and time are min-max normalized to $[-1,1]$ over valid rows. \\
\bottomrule
\end{tabularx}
\label{tab:partition-hparams}
\end{table}

\paragraph{Column selection}
The dynamic variables are partitioned into target columns, which the model may be asked to predict, and observation columns. RDTC is
always the first target column. Additional targets are sampled from
the auxiliary time series $\mathbf{u}$, and the drivers $x$ and $y$. Feature columns are sampled from the remaining var.

The number of target columns is sampled from $\{1,\ldots,16\}$ with probabilities
proportional to

\begin{align*}
\rho_{\mathrm{act}} & = & 
(1.00,0.95,0.90,0.86,0.82,0.78,0.74,0.70, \\
 && 0.67,0.64,0.61,0.58,0.55,0.52,0.49,0.46).
\end{align*}

The number of feature columns is sampled from $\{1,\ldots,20\}$ with probabilities
proportional to

\begin{align*}
\rho_{\mathrm{feat}} & = &
(1.00,0.97,0.94,0.91,0.88,0.85,0.82,0.79,0.76,0.73,\\
 && 0.70,0.67,0.64,0.61,0.58,0.55,0.52,0.49,0.46,0.43).
\end{align*}

\paragraph{Masking}
RDTC is always masked for every query row, making $\Lambda(t)$ the primary hidden target. The additional query task is sampled with a 50/50 mixture. In the \texttt{none} task, no additional query mask is applied; these views train nowcasting of current distance to criticality from currently observed signals. In the \texttt{forecast} task, a cutoff fraction $c\sim U(0.2,0.4)$ is sampled and further selected action columns are masked after that cutoff.

\paragraph{Normalization and missing values}
Normalization is applied after partitioning and masking, using only information available under inference-time constraints. Raw RDTC is transformed
to the model target
\begin{equation}
    \Lambda^* = \tanh(5\Lambda).
    \label{eq:rdtc-tanh-transform}
\end{equation}
All other columns are median-centered
and transformed with an adaptive inverse-hyperbolic-sine map,
\begin{equation}
    x \mapsto \operatorname{asinh}\!\left(\frac{x-\operatorname{median}(x)}{s}\right),
    \qquad
    s=\max(0.01,\,\operatorname{IQR}(x)).
    \label{eq:asinh-payload-transform}
\end{equation}
The median and IQR are computed from the observed rows only, excluding masked prediction targets and excluding RDTC from the statistics. They are computed once per task over the whole observed window, so only the prediction at the last observed row, at which all risk scores are read, and the forecast rows are free of later observations; earlier points of a nowcast trajectory are normalized using later observations.

\paragraph{Training targets}
The model predicts 99 quantiles for every masked action value and is optimized
with pinball loss. For distributional decoding and validation metrics, quantile
outputs are sorted to enforce non-crossing quantiles; CDF scores are 0 or 1 outside the
predicted quantile range. RDTC targets
receive unit loss weight, while auxiliary signal reconstruction targets receive
weight 0.2. This makes distance-to-criticality prediction the primary objective
while using reconstruction of observed dynamics as an auxiliary regularizer.

\subsection{TipPFN Model Architecture}
\label{si-sec:tippfn-architecture}

The trained model is a transformer architecture. A task is represented as a padded table with row index
$r=1,\ldots,R$ and column index split into action columns and feature columns.
Each row has two identifier values, episode id and time, denoted
$\iota_r=(e_r,t_r)$. Let $y_{r,a}$ denote the value in action slot
$a \in \{1,\ldots,A\}$ and $x_{r,f}$ the value in feature slot
$f \in \{1,\ldots,F\}$. The first action slot is reserved for RDTC,
$y_{r,1}=\Lambda^*_r$ after the tanh transform in
Equation~\ref{eq:rdtc-tanh-transform}. A Boolean mask $m_{r,a}$ indicates which
action values are prediction targets.

The model uses separate scalar encoders for action and feature values,
\[
    E_y:\mathbb{R}\to\mathbb{R}^{d_{\mathrm{model}}},
    \qquad
    E_x:\mathbb{R}\to\mathbb{R}^{d_{\mathrm{model}}},
\]
and fixed random positional buffers for rows, action columns, and
identifier/feature columns. For active, unmasked slots the initial tokens are
therefore
\[
    h^{(0)}_{r,a}=E_y(y_{r,a})+p^{\mathrm{row}}_r+p^{\mathrm{act}}_a,
    \qquad
    h^{(0)}_{r,A+g}=E_x(v_{r,g})+p^{\mathrm{row}}_r+p^{\mathrm{feat}}_g,
\]
where $a=1,\ldots,A$, $g=1,\ldots,F+2$, and
$v_r=(e_r,t_r,x_{r,1},\ldots,x_{r,F})$. Masked action tokens are replaced by a
zero vector before entering the transformer.

\begin{table}[h]
\centering
\small
\caption{\textbf{Resolved TipPFN architecture hyperparameters.}}
\begin{tabularx}{\linewidth}{lX}
\toprule
Quantity & Value \\
\midrule
Maximum rows $R$ & 512. \\
Identifier columns & 2: episode id and time. \\
Maximum feature columns $F$ & 20 payload features, plus the 2 identifier columns in the feature stream. \\
Maximum action columns $A$ & 16. \\
Total token columns per row & $16 + 2 + 20 = 38$. \\
Hidden width $d_{\mathrm{model}}$ & 1280. \\
Transformer blocks & 8 row-column blocks. \\
Attention heads & 8 heads; head dimension 160. \\
MLP width & Ratio 4; hidden width 5120. \\
Dropout & 0.1 in attention and MLP residual branches during training. \\
Output dimension & 99 per action token. \\
Quantile levels & $\alpha_j=j/100$, $j=1,\ldots,99$. \\
Row attention mode & Causal. \\
Causal time tolerance & $10^{-6}$. \\
Learned parameters & 209,959,779, excluding fixed positional buffers. \\
Fixed positional buffers & 704,000 scalar entries: $512\times1280$ row, $22\times1280$ id/feature-column, and $16\times1280$ action-column buffers. \\
\bottomrule
\end{tabularx}
\label{tab:tippfn-architecture-hparams}
\end{table}

Each row-column block applies column attention, row attention, and an MLP with
pre-normalization and residual connections:
\[
\begin{aligned}
    H &\leftarrow H + \operatorname{Dropout}
        \bigl(\operatorname{Attn}_{\mathrm{col}}(\operatorname{LN}(H))\bigr), \\
    H &\leftarrow H + \operatorname{Dropout}
        \bigl(\operatorname{Attn}_{\mathrm{row}}(\operatorname{LN}(H))\bigr), \\
    H &\leftarrow H + \operatorname{Dropout}
        \bigl(\operatorname{MLP}(\operatorname{LN}(H))\bigr).
\end{aligned}
\]
Column attention is applied independently within each row across the active
action, identifier, and feature columns. Row attention is applied independently
for each column across the temporal/episode rows. Attention uses fused scaled
dot-product attention with bias-free query, key, value, and output projections.
The MLP is a two-layer feed-forward network with GELU activation using the
\texttt{tanh} approximation.

\paragraph{Causal row mask}
Let $c_i$ indicate that row $i$ belongs to a context episode. In causal mode,
the set of row keys visible to query row $i$ is
\begin{equation}
\mathcal{A}(i)=
\begin{cases}
    \{j: c_j=1\}\cup\{i\}, & c_i=1,\\
    \{j: c_j=1\}\cup\{j: e_j=e_i,\ t_j<t_i-10^{-6}\}\cup\{i\}, & c_i=0.
\end{cases}
\label{eq:tippfn-causal-mask}
\end{equation}
Padded keys are removed from $\mathcal{A}(i)$. Thus context rows can exchange
information bidirectionally within the context set, while query rows can attend
to all context rows and only to their own past and present query rows.

\subsection{Training Configuration}
\label{si-sec:tippfn-training}

Training uses the synthetic task distribution described above.

For an output vector $\hat q_{r,a,1:99}$ and target $y_{r,a}$, the pinball loss
at quantile level $\alpha$ is
\[
    \ell_{\alpha}(\hat q,y)
    =
    \begin{cases}
        \alpha (y-\hat q), & y \geq \hat q,\\
        (\alpha-1)(y-\hat q), & y < \hat q.
    \end{cases}
\]
The per-entry loss is the mean of $\ell_{\alpha_j}$ over the 99 quantile
levels. Let $\mathcal{M}_{\mathrm{rdtc}}$ be the masked RDTC entries, and let
$\mathcal{M}_{\mathrm{sig}}$ be all other masked target entries. The training
loss is
\[
    \mathcal{L}
    =
    1.0\,
    \operatorname{mean}_{(r,a)\in\mathcal{M}_{\mathrm{rdtc}}}
    \ell(\hat q_{r,a},y_{r,a})
    +
    0.2\,
    \operatorname{mean}_{(r,a)\in\mathcal{M}_{\mathrm{sig}}}
    \ell(\hat q_{r,a},y_{r,a}),
\]
with empty groups skipped.

Training hyperparameters are given in table~\ref{tab:tippfn-training-hparams}.

\begin{table}[h]
\centering
\small
\caption{\textbf{Resolved optimization and trainer hyperparameters.}}
\begin{tabularx}{\linewidth}{lX}
\toprule
Quantity & Value \\
\midrule
Per-device batch size & 42 transformed views. \\
Global batch size & 168 transformed views across 4 DDP workers (4 NVIDIA H200 GPUs, one node). \\
Data-loader workers & 1 worker, \texttt{persistent\_workers=false}, \texttt{pin\_memory=true}. \\
Precision & \texttt{bf16-mixed}. \\
Optimizer & Muon~\cite{jordan2024muon} for 2D trainable tensors, AdamW for remaining tensors. \\
Learning rate & 0.01. \\
Weight decay & 0.01 for decay groups; no decay for biases, 1D norms, and parameters matching no-decay keywords. \\
Muon parameters & Momentum 0.95; Nesterov enabled; Newton-Schulz steps 5; $\epsilon=10^{-7}$. \\
AdamW parameters & $\beta_1=0.9$, $\beta_2=0.999$, $\epsilon=10^{-7}$. \\
Scheduler & Linear warmup followed by linear decay. \\
Warmup & 1\% of 200,000 steps, i.e. 2,000 optimizer steps. \\
Minimum LR scale & 0.1, giving final LR 0.001 from the initial LR 0.01. \\
Maximum optimizer steps & 200,000. \\
Released checkpoint & Step 22,000 (learning rate $0.91\times$ peak). \\
Gradient clipping & Global norm clipping at 1.0. \\
Random seeding & Lightning \texttt{seed\_everything=true}. \\
\bottomrule
\end{tabularx}
\label{tab:tippfn-training-hparams}
\end{table}

\subsection{Linear Forcing schedules}\label{si-sec:forcing}
In the following we provide more details on the forcing schedule to generate the various tipping scenarios used for validation datasets, see~Appendix\ref{si-sec:datasets}.
Note that these differ from the non-linear forcing schedules used in the training dataset.

For each episode in the synthetic and semi-real b-type systems, the a forcing schedule was randomly selected.
To gain a balanced distribution of positive (critical) and negative (non-critical) episodes, an episode was grouped as \emph{critical} with 50\% probability.
If not critical, one of the four non-critical groups described in Table~\ref{tab:trajectory-classes-lambda} was uniform randomly selected.

The goal of these forcing schedules was to provide a realistic evaluation task: instead of just distinguishing between critical and equilibrium trajectories, the evaluation task includes distinguishing non-critically forced b-type systems (\emph{approaching} criticality, but not reaching it) from actual critical forcings.
We employ exclusively linear forcing schedules for all experiments; changing the forcing at an unobserved future time would be causally intractable for a prediction model~\cite{nathaniel2025deep}.

\begin{figure}[h]
    \centering
    \includegraphics[width=1\linewidth]{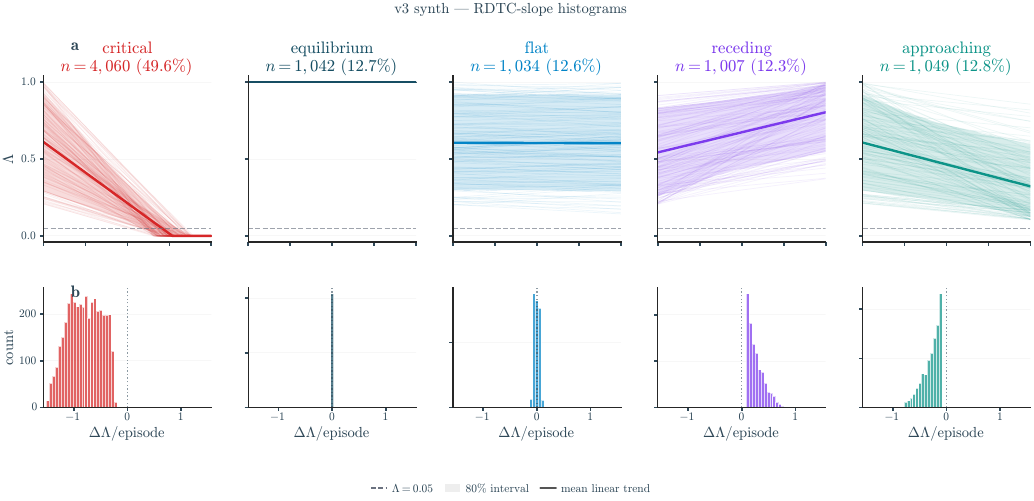}
    \caption{
        Forcing schedules $\Lambda(t)$ and overall group sizes (number of episodes) and composition for the canonical evaluation datasets (b\_fold, b\_hopf, b\_transcritical).
        Refer to Table~\ref{tab:si-system-sample-episode-counts} for the actually used episode counts out of this larger dataset.
    }
    \label{fig:forcing-synth}
\end{figure}

\begin{figure}[h]
    \centering
    \includegraphics[width=1\linewidth]{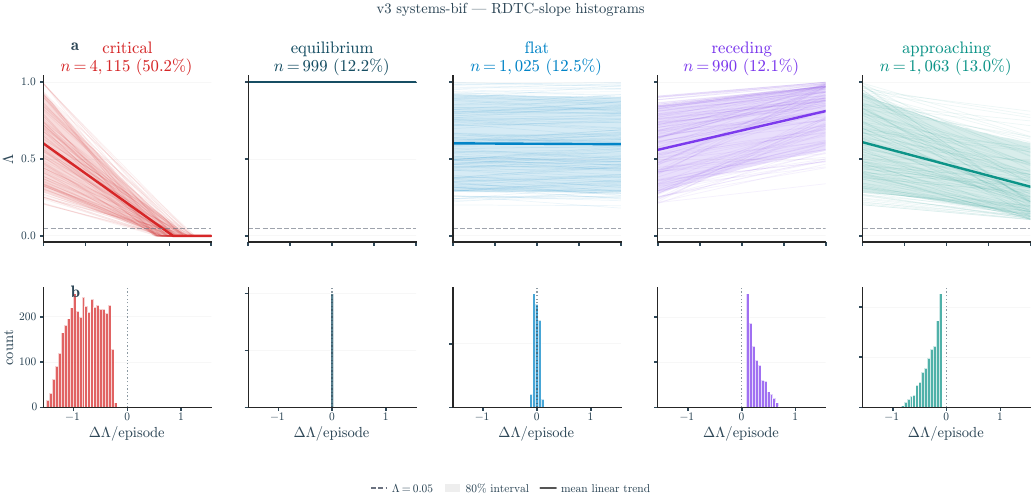}
    \caption{
        Forcing schedules $\Lambda(t)$ and overall group sizes (number of episodes) and composition for the semi-real bifurcation systems (see Table~\ref{tab:tipping}).
        Refer to Table~\ref{tab:si-system-sample-episode-counts} for the actually used episode counts out of this larger dataset.
    }
    \label{fig:forcing-systems-bif}
\end{figure}

As figures \ref{fig:forcing-synth} and \ref{fig:forcing-systems-bif} show, these normalized forcing schedules $\Lambda(t) = 1-\tilde\lambda(t)$ also varied in the initial starting point and the time of reaching criticality.
To avoid strong perturbation, the tipping system was warmed-up by slowly and smoothly varying the control parameter from equilibrium to the parameter that is the start of the forcing schedule (not shown).

\begin{table}[th]
\centering
\caption{
    Trajectory classes defined by the evolution of the normalized control parameter $\tilde{\lambda}(t)$ and their corresponding $\Lambda(t)$ behavior.
}
\label{tab:trajectory-classes-lambda}
\begin{tabular}{p{0.22\linewidth} p{0.44\linewidth} p{0.26\linewidth}}
\hline
\textbf{Trajectory type} & \textbf{Definition of $\tilde{\lambda}(t)$} & \textbf{$\Lambda(t)$ behavior} \\
\hline

\makecell[l]{\textit{Critical}\\\textit{(tipping)}}
&
$
\tilde{\lambda}(t) =
\begin{cases}
\tilde{\lambda}_0 + \alpha t, & t < t_c, \\
1, & t \ge t_c,
\end{cases}
\quad \alpha > 0
$
&
$\Lambda(t) \to 0$.
Critical value $\tilde{\lambda}=1$ is reached at $t_c$.
\\

\makecell[l]{\textit{Approaching}\\\textit{(non-critical)}}
&
$
\tilde{\lambda}(t) = \tilde{\lambda}_0 + \alpha t,
\quad
\tilde{\lambda}(t) < 1 - \epsilon \ \forall t,
\quad \epsilon > 0
$
&
$\Lambda(t) \to \epsilon$.
The trajectory remains bounded away from criticality.
\\

\makecell[l]{\textit{Receding}\\\textit{(non-critical)}}
&
$
\tilde{\lambda}(t) = \tilde{\lambda}_0 - \alpha t,
\quad
\tilde{\lambda}_0 < 1 \ \forall t
$
&
$\Lambda(t) \to 1$.
The trajectory moves away from the critical point.
\\

\makecell[l]{\textit{Flat}\\\textit{(non-critical)}}
&
$
\tilde{\lambda}(t) = \tilde{\lambda}_0 + \delta(t),
\quad
|\delta(t)| \ll 1
$
&
$\Lambda(t) \approx \mathrm{const.}$
\\

\makecell[l]{\textit{Equilibrium}}
&
$
\tilde{\lambda}(t) = 0
\quad \forall t
$
&
$\Lambda(t) \approx 1$.
System remains far from criticality.
\\

\hline
\end{tabular}
\end{table}

\paragraph{R-tipping}
The notion of a \emph{normalized} forcing schedule does not easily translate to r-type tipping systems, which require to surpass not a certain forcing value but a critical change in forcing~$\dot\lambda$.
In particular, there is no straight-forward equivalent of the \emph{approaching} or \emph{receding} groups.
To still create some variability in schedules, we split the non-critical group into \emph{equilbibrium} and \emph{flat} groups, where the flat group has a higher but still non-critical maximum forcing rate.

\FloatBarrier
\subsection{Compute Resources}\label{si-sec:compute-resources}

All experiments were conducted on rented GPU servers (IONOS cloud) and a single dedicated CPU server (Hetzner, Germany). 
We report hardware specifications, per-stage compute, and total project compute below.

\begin{itemize}
    \item \textbf{Dataset generation.}
      The synthetic training dataset was generated on a single CPU server (AMD EPYC 9454P, 48~cores / 96~threads,
      125\,GB RAM) over approximately 8~days at \(\sim\!80\%\) CPU utilisation (\(\sim\!7{,}400\) CPU core-hours).
    
    \item \textbf{Model training (reported).}
      The checkpoint used in all experiments was trained on 4\(\times\) NVIDIA H200 NVL GPUs (141\,GB HBM3e each, 1\,TB system RAM) for approximately 47~hours (\(\sim\!189\) H200 GPU-hours) at near-full utilisation (\(\sim\!100\%\) SM occupancy, \(\sim\!493\)\,W average draw per device).
    
    \item \textbf{Evaluation (reported).}
      All evaluation runs used NVIDIA H200 NVL GPUs (1--4 per run).
      The paper-bound evaluation runs (7~datasets, including ablations and
      step-size variants) consumed approximately \(\sim\!150\) GPU-hours in
      total.
      Individual runs ranged from \(\sim\!1\)\,h (lean / fast configurations) to \(\sim\!20\)\,h (full SWEC and TAC sweeps on 4\(\times\) H200), with TabPFN and Bury baselines requiring considerable CPU and GPU time.
    
    \item \textbf{Total project compute.}
      The full research project, including preliminary and failed experiments not reported in the paper, consumed substantially more compute than the final experiments.
      Over the course of development, the training project logged \(\sim\!1{,}400\) GPU-hours across 44~runs (spanning smaller NVIDIA RTX~6000~Ada, RTX~A6000, RTX~Pro class GPUs).
      For evaluation, we logged \(\sim\!750\) GPU-hours across 347~runs (all on H200).
      Including dataset generation, we estimate the total project compute at \(\sim\!2{,}150\) GPU-hours plus \(\sim\!7{,}400\) CPU core-hours.
\end{itemize}

\subsection{Released assets}\label{si-sec:released-assets}

We release the following assets.
\begin{itemize}
    \item \textbf{Code} for data loading, training, evaluation and all baselines: \url{https://github.com/kausable/tippfn-neurips26} (Apache-2.0).
    \item \textbf{TipPFN checkpoint} (210M parameters, step 22,000): \url{https://huggingface.co/kausable/tippfn-neurips26} (Apache-2.0); \textbf{training data}: \url{https://huggingface.co/datasets/kausable/tippfn-neurips26-train} (CC BY 4.0).
    \item \textbf{Evaluation sets} for the canonical, semi-real, TAC and DaphniaExt rows of Table~\ref{tab:auroc-results-best-col}: \url{https://huggingface.co/datasets/kausable/tippfn-neurips26-eval}. Simulated data are CC BY 4.0; the TAC and DaphniaExt recordings are derived from Dryad deposits under CC0 (\url{https://doi.org/10.5061/dryad.4cj4k}, \url{https://doi.org/10.5061/dryad.q3p64}).
    \item \textbf{Baseline checkpoints} trained by us: the Bury et al.~\cite{bury2021deep} classifiers retrained on our training data (\url{https://huggingface.co/kausable/tippfn-neurips26-bury-retrained}, Apache-2.0), and the Huang et al.~\cite{huang2024deep} classifier, trained by Juan Nathaniel with the authors' code and data~\cite{huang2024code} (\url{https://huggingface.co/kausable/tippfn-neurips26-huang}, CC BY 4.0).
\end{itemize}
The following assets are not redistributed.
The SWEC-iEEG recordings may be used for research and education only and may not be redistributed; the code repository provides a guide and a preprocessing script that rebuild our evaluation set from the recordings (\url{http://ieeg-swez.ethz.ch/}).
The TabPFN 2.6 weights are licensed by Prior Labs and are obtained through the \texttt{tabpfn} package.
The original classifiers of Bury et al.\ and Zhuge et al.\ are downloaded from the authors' repositories (\url{https://github.com/ThomasMBury/ewstools}, \url{https://github.com/zhugchzo/dl_occurrence_tipping}).

\clearpage
\newpage
\section{Experiment details}\label{si-sec:experiments}

\subsection{Datasets}\label{si-sec:datasets}

\begin{figure}[h!]
    \centering
    \begin{subfigure}[h]{0.24\linewidth}
        \centering
        \includegraphics[width=\linewidth]{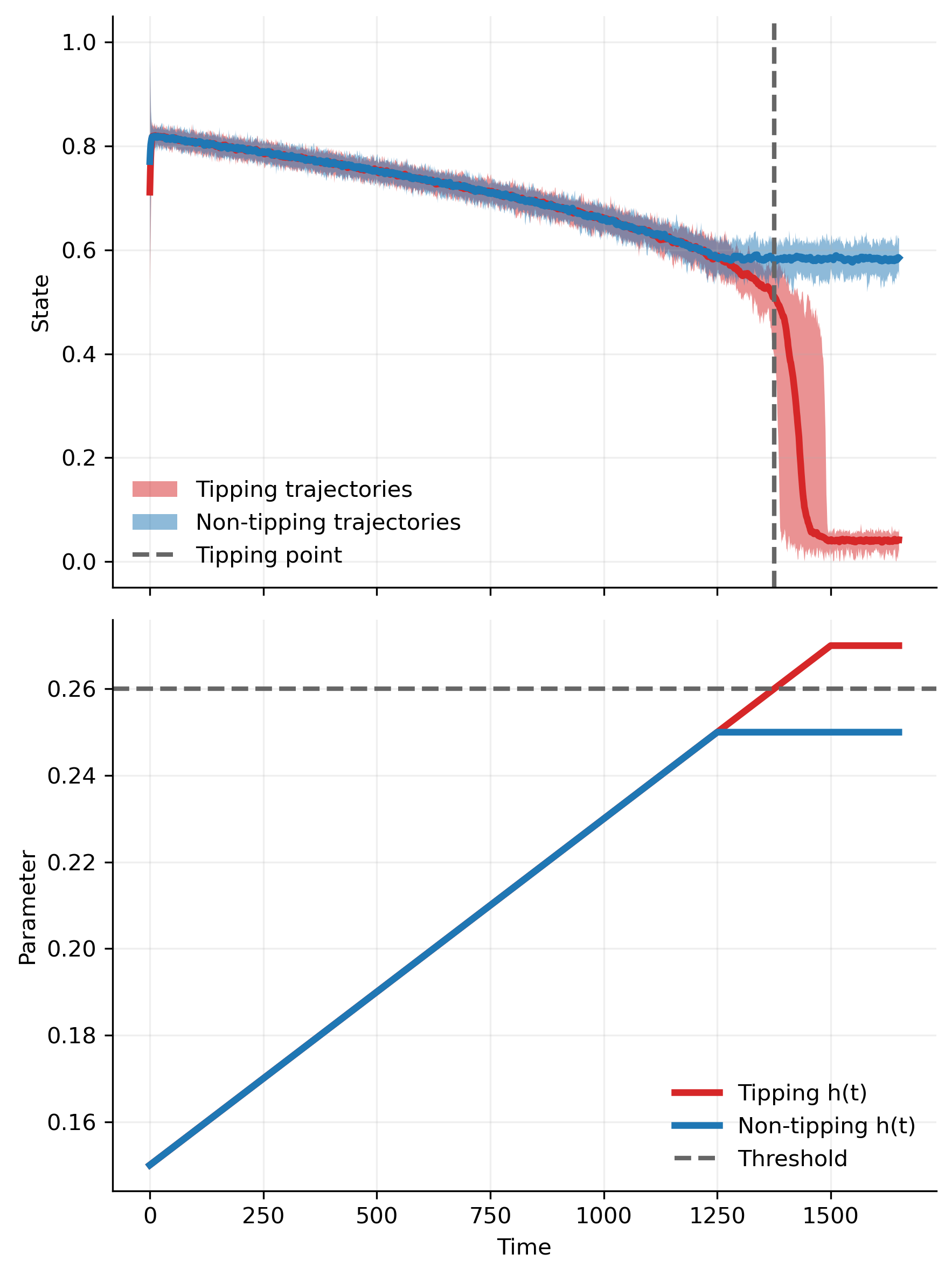}
        \caption{May's harvesting}
        \label{fig:bifurcation_examples_harvesting}
    \end{subfigure}
    \hfill
    \begin{subfigure}[h]{0.24\linewidth}
        \centering
        \includegraphics[width=\linewidth]{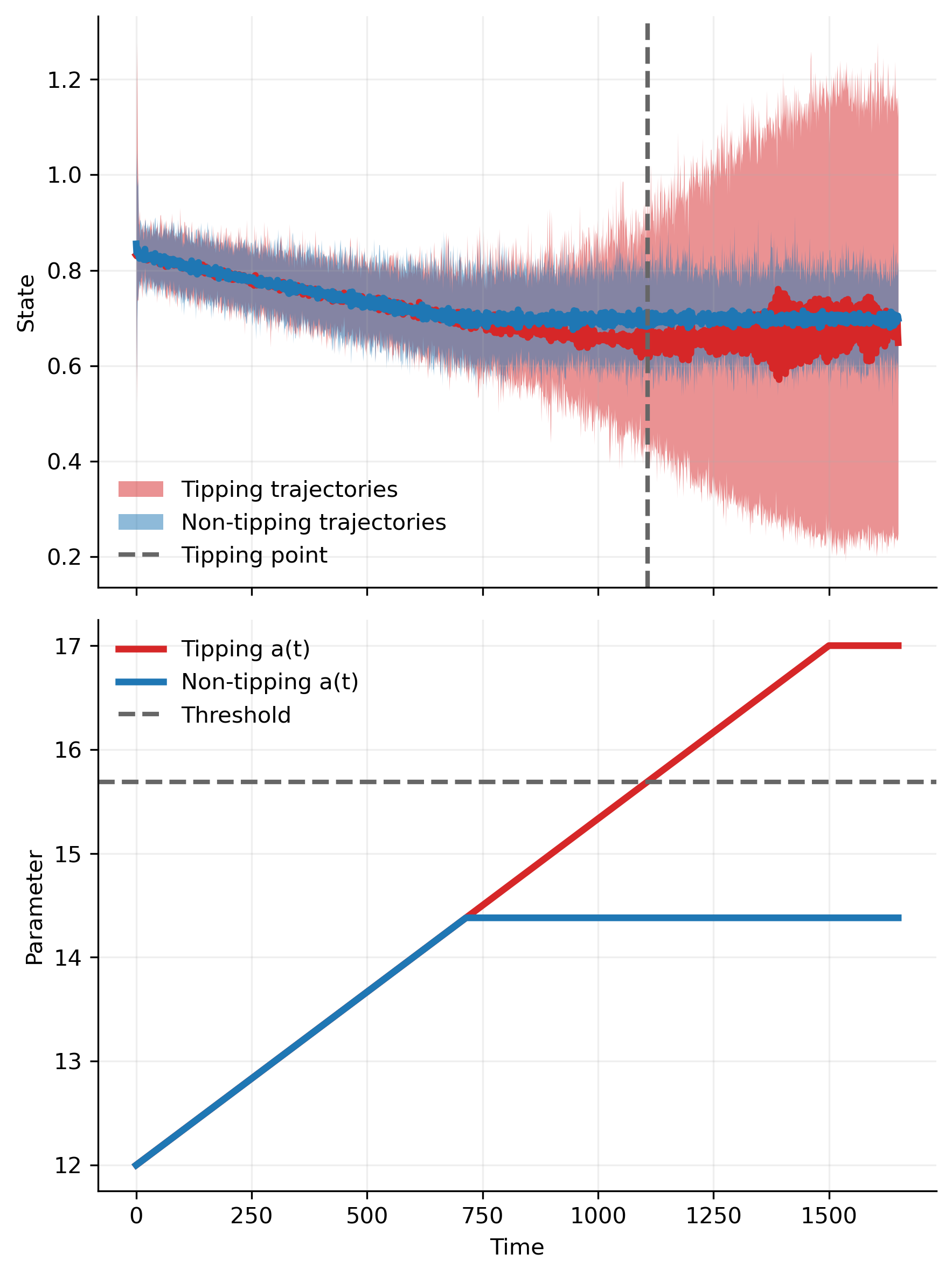}
        \caption{Rosenzweig-MacArthur}
        \label{fig:bifurcation_examples_rm}
    \end{subfigure}
    \hfill
    \begin{subfigure}[h]{0.24\linewidth}
        \centering
        \includegraphics[width=\linewidth]{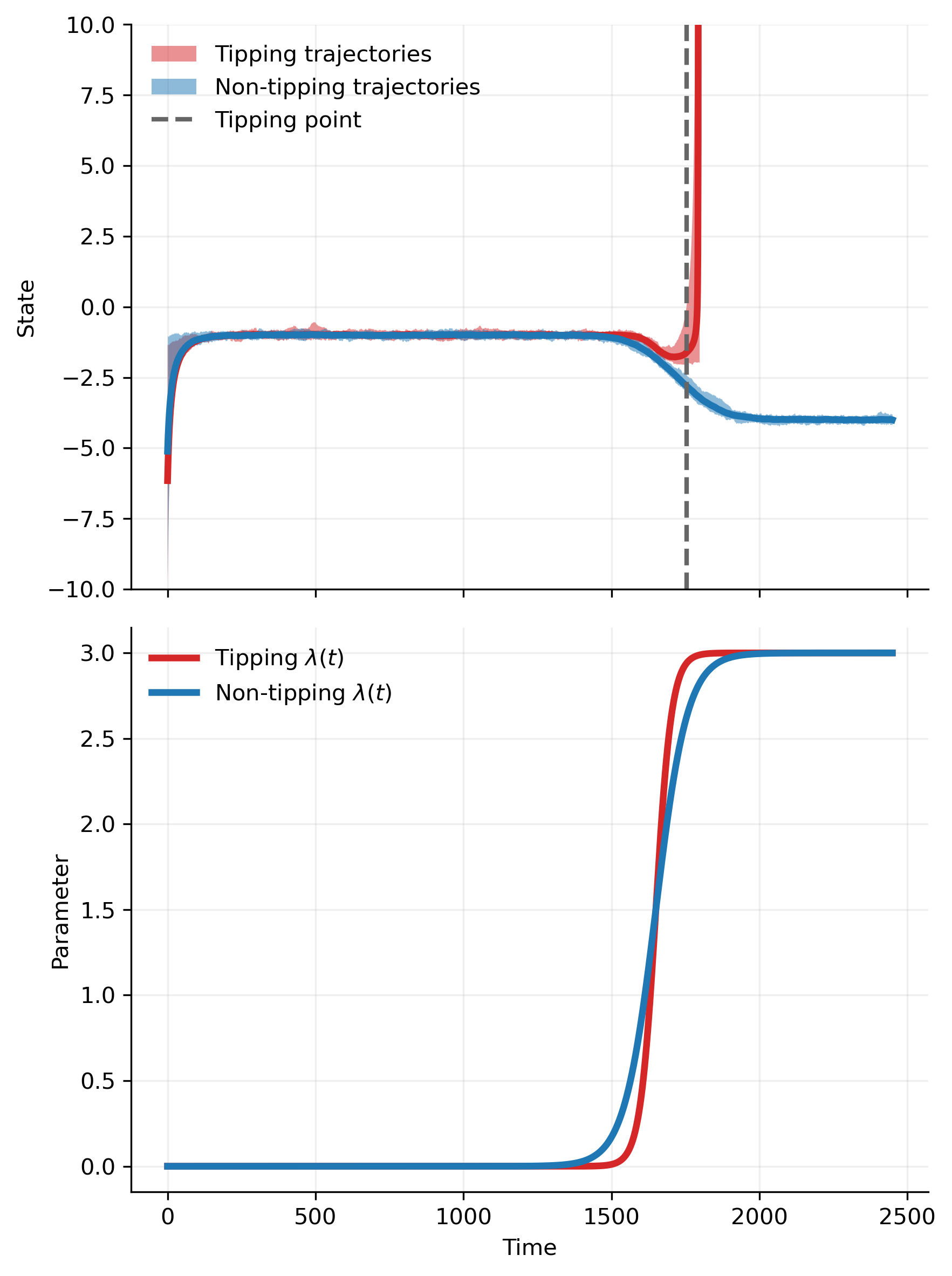}
        \caption{Saddle-node}
        \label{fig:rate_examples_saddle_node}
    \end{subfigure}
    \hfill
    \begin{subfigure}[h]{0.24\linewidth}
        \centering
        \includegraphics[width=\linewidth]{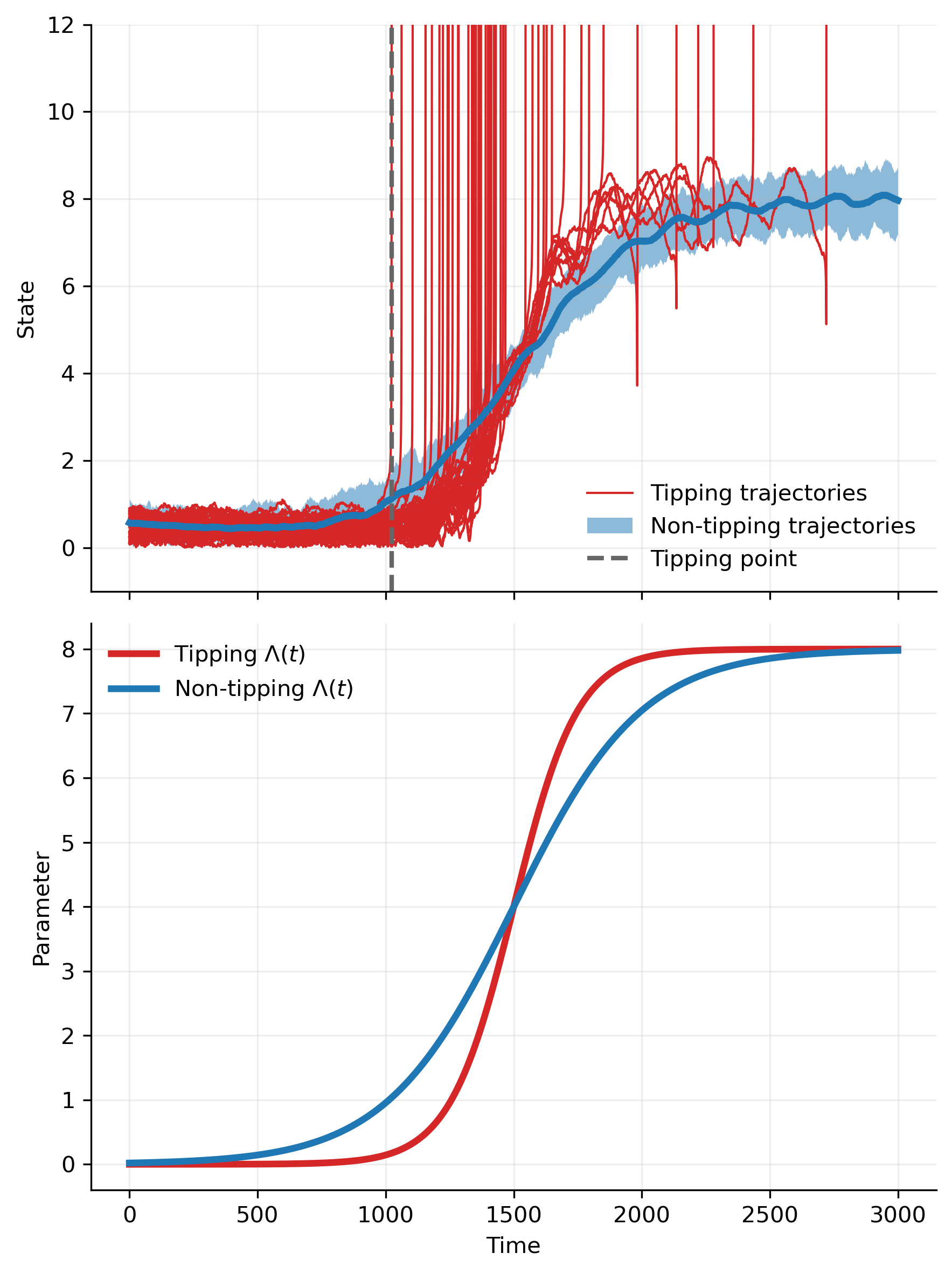}
        \caption{Bautin}
        \label{fig:rate_examples_bautin}
    \end{subfigure}
    \caption{
        Representative B-tipping (\textbf{a}, \textbf{b}) and R-tipping (\textbf{c}, \textbf{d}) examples.
        In each panel, the upper plot shows examples of tipping (red) and non-tipping (blue) trajectories, and the lower plot shows the corresponding parameter schedule.
        Note that the validation datasets use a more diverse set of forcing schedules, see Figures~\ref{fig:forcing-synth} and~\ref{fig:forcing-systems-bif}, which vary the initial value and slope (for b-type systems only) as well as the timing of the critical transition (for all systems, including r-type).
    }
    \label{fig:bifurcation_examples}
\end{figure}

\begin{figure}[h!]
    \centering
    \begin{subfigure}[h]{\linewidth}
        \centering
        \includegraphics[width=\linewidth]{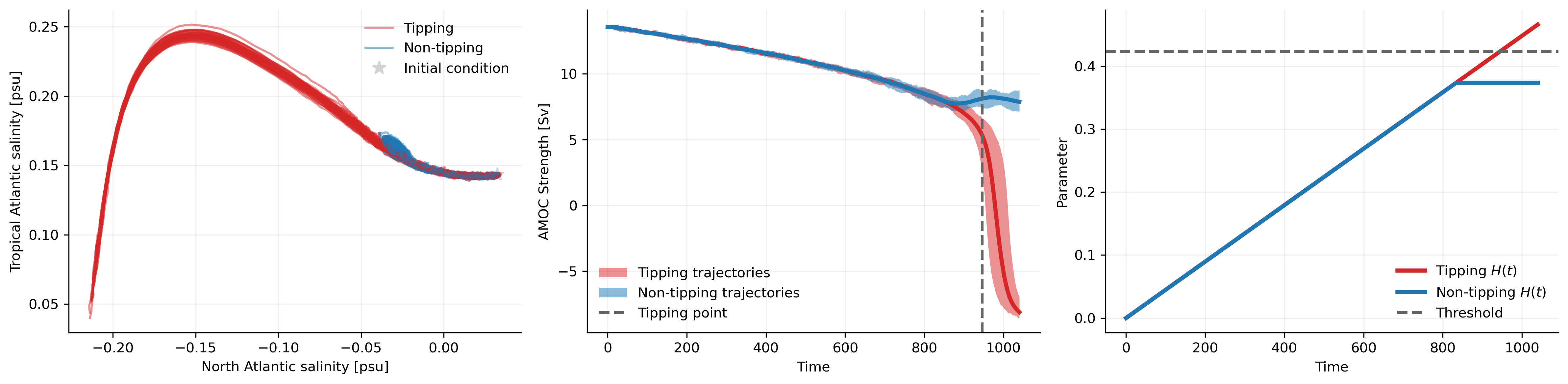}
        \caption{AMOC box-model (B-tipping)}
        \label{fig:bifurcation_examples_amoc3b}
    \end{subfigure}
    \hfill
    \begin{subfigure}[h]{\linewidth}
        \centering
        \includegraphics[width=\linewidth]{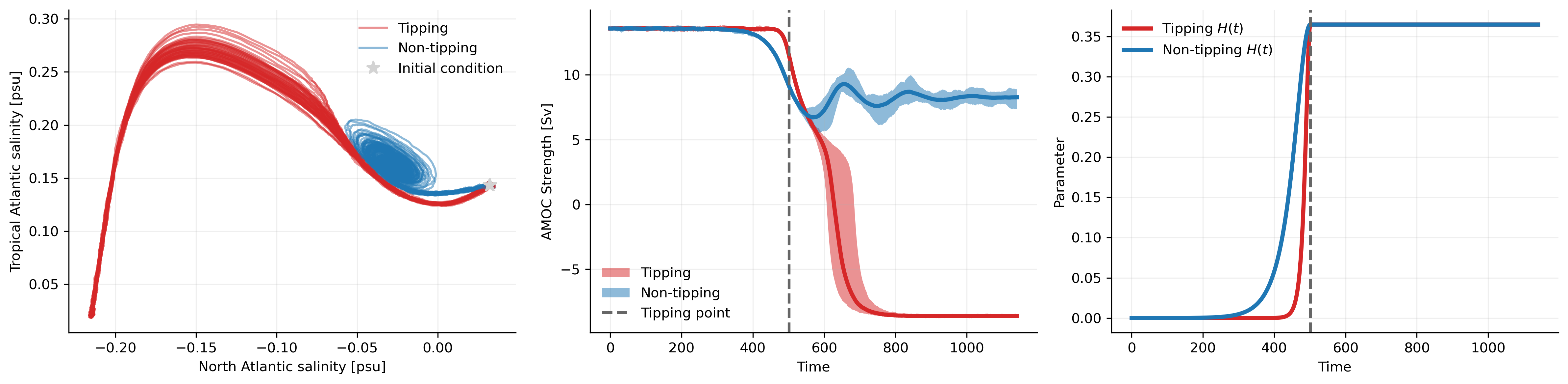}
        \caption{AMOC box-model (R-tipping)}
        \label{fig:rate_examples_amoc3b}
    \end{subfigure}
    \caption{
    Examples of idealized AMOC box-model trajectories.
    (\textbf{a}) B-tipping and (\textbf{b}) R-tipping.
    Each panel shows \textbf{left}: phase-space evolution in $(S_T,S_N)$, \textbf{center}: AMOC strength $Q$, and \textbf{right}: the corresponding hosing or control schedule.
    Note that for B-tipping, we use a more diverse set of forcing schedules than shown here, see Figures~\ref{fig:forcing-synth} and~\ref{fig:forcing-systems-bif}, which vary the initial value and slope as well as the critical time.
    }
    \label{fig:amoc3box}
\end{figure}

\paragraph{Summary} Table~\ref{tab:tipping} summarizes the datasets we simulated with TipBox\footnote{\url{https://github.com/juannat7/tipbox}} as well as additional benchmark datasets for evaluation on real-world examples.
Sample tipping and non-tipping trajectories, along with their specific instantiation of the bifurcating parameter or forcing rate schedules are shown in Figures~\ref{fig:bifurcation_examples} and ~\ref{fig:amoc3box}.

\begin{table}[h]
\centering
\small

\caption{Datasets used for evaluating tipping point prediction across b-tipping (B), r-tipping (R), and n-tipping (N), spanning canonical, semi-real, sim-to-real, and real-world systems.}
\begin{tabularx}{\columnwidth}{l l X l X}
\toprule
\textbf{Dataset} & \textbf{System} & \textbf{Description} & \textbf{Type} & \textbf{References} \\
\midrule
Canonical & \texttt{b\_fold} & Fold (saddle-node) bifurcation & B & \cite{bury2021deep} \\
Canonical & \texttt{b\_hopf} & Hopf bifurcation & B & \cite{bury2021deep} \\
Canonical & \texttt{b\_transcritical} & Transcritical bifurcation & B & \cite{bury2021deep} \\
\midrule
Semi-real & \texttt{b\_harvesting} & May's harvesting model & B & \cite{may1977thresholds} \\
Semi-real & \texttt{b\_rosenzweig\_macarthur\_tc} & Consumer-resource (transcritical) & B & \cite{bury2021deep} \\
Semi-real & \texttt{b\_rosenzweig\_macarthur\_hopf} & Consumer-resource (Hopf) & B & \cite{bury2021deep} \\
Semi-real & \texttt{b\_seirx\_tc} & SEIRx disease-vaccination & B & \cite{bury2021deep} \\
Semi-real & \texttt{b\_amoc} & AMOC 2-box model & B & \cite{ritchie2023rate,alkhayuon2019basin} \\
Semi-real & \texttt{r\_bautin} & Bautin normal form & R & \cite{ritchie2023rate} \\
Semi-real & \texttt{r\_compost\_bomb} & Compost bomb instability & R & \cite{luke2011soil,wieczorek2011excitability} \\
Semi-real & \texttt{r\_saddle\_node} & Saddle-node normal form & R & \cite{ashwin2012tipping} \\
Semi-real & \texttt{r\_amoc} & AMOC 2-box model & R & \cite{ritchie2023rate,alkhayuon2019basin} \\
\midrule
Sim-to-real & \texttt{TAC} & Thermoacoustic instability experiments & B & \cite{bonciolini2018experiments}\\
Sim-to-real & \texttt{DaphniaExt} & Ecological extinction events & N & \cite{drake2010early} \\
\midrule
Real & \texttt{SWEC-iEEG} & Epileptic seizure onset (iEEG) & B & \cite{burrello2019laelaps,jirsa2014nature} \\
Real & \texttt{microcosm} & Cyanobacteria under light stress & B/N & \cite{veraart2012recovery} \\
Real & \texttt{voice} & Phonation onset under increasing flow (Hopf) & B & \cite{mergell1998phonation,murray2012vibratory,shimamura2016effect} \\
Real & \texttt{cellular\_atp} & Cellular energy status under hypoxia & B/N & \cite{wagner2019multiparametric} \\
Real & \texttt{greenhouse\_earth} & Calcium carbonate in the end of greenhouse Earth & B & \cite{dakos2008slowing} \\
Real & \texttt{blackout\_frequency} & Bus voltage frequency before power grid failure & N & \cite{council1996western} \\
Real & \texttt{AMOC} & Observed variability of the meridional overturning circulation (MOC) & R/B & \cite{moat2024atlantic} \\
\bottomrule
\end{tabularx}
\label{tab:tipping}
\end{table}

\newpage
\paragraph{b\_fold} The fold (saddle-node) bifurcation is described by the normal form
\begin{equation}
\frac{dx}{dt} = \mu - x^2,
\end{equation}
where two equilibria collide and annihilate at $\mu = 0$, leading to an abrupt transition. This bifurcation is associated with catastrophic shifts, where the system loses stability and jumps to a distant attractor.

\paragraph{b\_hopf} The Hopf bifurcation involves a two-dimensional system of the form
\begin{equation}
\begin{aligned}
\frac{dx}{dt} &= \mu x - y - x(x^2 + y^2), \\
\frac{dy}{dt} &= x + \mu y - y(x^2 + y^2),
\end{aligned}
\end{equation}
where a pair of complex conjugate eigenvalues crosses the imaginary axis at $\mu = 0$, leading to the emergence of a stable or unstable limit cycle. This transition results in oscillatory dynamics rather than a shift between steady states.

\paragraph{b\_transcritical} The transcritical bifurcation is given by
\begin{equation}
\frac{dx}{dt} = \mu x - x^2,
\end{equation}
in which two equilibria exist for all parameter values and exchange stability at $\mu = 0$. In contrast to the fold case, the transition is continuous and does not involve the disappearance of equilibria, but rather a reorganization of stability.

\paragraph{b\_harvesting} We consider May's single-species harvesting model \cite{may1977thresholds} describing the dynamics of a biomass variable $x(t) \in \mathbb{R}_+$. The stochastic dynamics are given by
\begin{equation}
\frac{dx}{dt} = r x\left(1-\frac{x}{k}\right) - h(t)\frac{x^2}{s^2+x^2} + \sigma_x \,\xi(t),
\end{equation}
where $r$ is the intrinsic growth rate, $k$ is the carrying capacity, $s$ controls the saturation scale of harvesting, $h(t)$ is the harvesting rate, and $\xi(t)$ denotes Gaussian white noise with amplitude $\sigma_x$.

The logistic growth term promotes population persistence, while the nonlinear harvesting term induces a destabilizing feedback. As the harvesting rate $h$ increases, the system undergoes a fold (saddle-node) bifurcation at a critical value $h^\star$, beyond which the stable positive-biomass equilibrium disappears, leading to an abrupt collapse of the population. 

We use parameters $r=1$, $k=1$, $s=0.1$, and $\sigma_x = 0.01$, for which the deterministic system exhibits a fold bifurcation at $h^\star \approx 0.26$. Tipping trajectories are generated by ramping the harvesting rate from $h=0.15$ to $h=0.27$, crossing the bifurcation threshold, while non-tipping trajectories follow the same initial ramp but are capped at $h=0.25$, remaining below the critical point. Tipping is identified as the collapse of the biomass toward the low-density state.

\paragraph{b\_rosenzweig\_macarthur\_tc and b\_rosenzweig\_macarthur\_hopf} We consider the Rosenzweig--MacArthur consumer--resource model \cite{rosenzweig1963graphical,bury2021deep} with state $\mathbf{x}(t) = (x(t), y(t))^\top \in \mathbb{R}_+^2$, where $x(t)$ denotes the resource population and $y(t)$ the consumer population. The stochastic dynamics are given by
\begin{align}
\frac{dx}{dt} &= r x\left(1-\frac{x}{k}\right) - \frac{a(t)\,x y}{1 + a(t) h x} + \sigma_x \,\xi_x(t), \\
\frac{dy}{dt} &= \frac{e\,a(t)\,x y}{1 + a(t) h x} - m y + \sigma_y \,\xi_y(t),
\end{align}
where $r$ is the intrinsic growth rate of the resource, $k$ its carrying capacity, $a(t)$ the consumer attack rate, $e$ the conversion efficiency, $h$ the handling time, and $m$ the consumer mortality rate. The noise terms $\xi_x(t)$ and $\xi_y(t)$ are independent Gaussian white noise processes with amplitudes $\sigma_x$ and $\sigma_y$.

As the attack rate $a$ increases, the system exhibits two bifurcations that structure its dynamical regimes. At $a \approx 5.60$, a transcritical bifurcation marks the consumer invasion threshold: for $a < 5.60$, the system converges to a consumer-free equilibrium $(x^\ast, 0)$, while for $a > 5.60$, a stable coexistence equilibrium $(x^\ast, y^\ast)$ emerges. As $a$ increases further, the coexistence equilibrium remains stable until a Hopf bifurcation at $a^\star \approx 15.69$, where a pair of complex conjugate eigenvalues crosses the imaginary axis, leading to the onset of predator--prey cycles. For $a > a^\star$, the system exhibits stable limit cycle behaviour.

Initial conditions are sampled near the equilibrium corresponding to the initial parameter value. In the transcritical regime, trajectories are initialized near the resource-only equilibrium, with $x(0)$ close to carrying capacity and a small but positive consumer density $y(0) > 0$, since $y=0$ is always an invariant equilibrium and would otherwise prevent consumer invasion even after crossing the transcritical threshold. In the Hopf regime, initial conditions are sampled near the stable coexistence equilibrium at the ramp start, ensuring trajectories begin on the equilibrium branch that later destabilizes.

We use parameters $r=4$, $k=1.7$, $e=0.5$, $h=0.15$, and $m=2$. For transcritical tipping, we generate trajectories by ramping $a$ from below to above $a \approx 5.60$, enabling the transition from a consumer-free to a coexistence state. For Hopf tipping, trajectories are generated by ramping $a$ from $12$ to $17$, thereby crossing the Hopf threshold, while non-tipping trajectories are capped below the respective bifurcation points. Tipping is identified either as successful consumer invasion (transcritical) or as the transition from equilibrium dynamics to persistent oscillations (Hopf).

\paragraph{b\_seirx\_tc} We consider a stochastic SEIRx model \cite{oraby2014influence,pananos2017critical,bury2021deep} capturing the coupled dynamics of infectious disease transmission and vaccination behaviour. The state is given by $\mathbf{x}(t) = (S(t), E(t), I(t), x(t))^\top$, where $S$, $E$, and $I$ denote the susceptible, exposed, and infectious populations, respectively, and $x(t)$ represents the fraction of individuals with provaccine sentiment. The recovered population is given by $R(t) = N - S - E - I$. The dynamics are
\begin{align}
\frac{dS}{dt} &= \mu N(1 - x) - \beta \frac{SI}{N} - \mu S + \sigma_S \,\xi_S(t), \\
\frac{dE}{dt} &= \beta \frac{SI}{N} - (\epsilon + \mu) E + \sigma_E \,\xi_E(t), \\
\frac{dI}{dt} &= \epsilon E - (\gamma + \mu) I + \sigma_I \,\xi_I(t), \\
\frac{dx}{dt} &= \kappa x(1 - x)\bigl(-\omega(t) + I + \delta(2x - 1)\bigr) + \sigma_x \,\xi_x(t),
\end{align}
where $\mu$ is the birth/death rate, $\beta$ the transmission rate, $\epsilon$ the exposed-to-infectious rate, $\gamma$ the recovery rate, $\kappa$ the social learning rate, $\delta$ the strength of injunctive social norms, and $\omega(t)$ the perceived risk of vaccination relative to infection. The noise terms $\xi_S, \xi_E, \xi_I, \xi_x$ are independent Gaussian white noise processes.

As the perceived vaccination risk $\omega$ increases, the system undergoes a transcritical bifurcation at $\omega = \delta$, corresponding to a loss of stability of the disease-free, high-vaccination equilibrium. Below the threshold ($\omega < \delta$), the system remains in a regime with high vaccination uptake ($x \approx 1$) and low infection prevalence. Above the threshold ($\omega > \delta$), vaccination sentiment declines, leading to the emergence of endemic infection. This transition from disease-free to endemic dynamics constitutes bifurcation-induced tipping.

We use parameters $N = 100{,}000$, $\mu = 0.02/52$, $\beta = 10.5$, $\epsilon = 0.7$, $\gamma = 0.7$, $\kappa = 0.007$, and $\delta = 50$, corresponding to a typical pediatric infectious disease. Noise amplitudes are $\sigma_S = \sigma_E = \sigma_I = 5$ and $\sigma_x = 5 \times 10^{-4}$. Tipping trajectories are generated by ramping $\omega$ from $0$ to $100$, thereby crossing the bifurcation threshold, while non-tipping trajectories remain below $\omega = \delta$. Tipping is identified as a collapse in vaccination sentiment ($x(t)$ decreasing) accompanied by a resurgence of infection ($I(t)$ increasing).

\paragraph{Non-tipping r-tipping episodes} Non-tipping counterparts of r-tipping episodes can be constructed in several ways, for example with a slower forcing rate from the start of the episode, as in the sets used here, or with forcing that follows the tipping episodes up to a later, slower ramp.
With the first construction, the two classes may already differ before the forcing ramp; Appendix~\ref{si-sec:aux-drive-sensitivity} reports results for the second.

\paragraph{r\_saddle\_node} We consider a stochastic saddle-node normal form with scalar state $x(t) \in \mathbb{R}$ and time-dependent forcing $\lambda(t)$. The dynamics are given by
\begin{equation}
\frac{dx}{dt} = (x + \lambda(t))^2 - 1 + \sqrt{2\sigma_x}\,\xi(t),
\end{equation}
where $\sigma_x$ controls the noise magnitude and $\xi(t)$ denotes Gaussian white noise. The external forcing evolves according to
\begin{equation}
\lambda(t) = \frac{\lambda_{\max}}{2}
\left[
\tanh\left(\frac{\lambda_{\max} \epsilon t}{2}\right) + 1
\right],
\end{equation}
where $\lambda_{\max}$ sets the forcing amplitude and $\epsilon$ determines the rate of change.

For each fixed $\lambda$, the frozen system exhibits the characteristic saddle-node structure with a stable and unstable equilibrium that collide at a critical parameter value. For sufficiently slow forcing ($\epsilon$ small), trajectories track the attracting equilibrium branch. However, when the forcing rate exceeds a critical threshold, the system fails to follow this quasi-static equilibrium and undergoes a rapid transition, even though the instantaneous system remains locally stable. This behaviour constitutes rate-induced tipping.

We use $\lambda_{\max}=3$ and compare a tipping regime with $\epsilon=1.25$ to a non-tipping regime with $\epsilon=0.625$, with noise level $\sigma_x = 0.008$. For analysis, we consider the scalar observable $x(t)$ and define tipping as the event $x(t) \geq 0$.

\paragraph{r\_bautin} We consider a stochastic Bautin system with complex state $z(t) = x(t) + i y(t)$ and time-dependent forcing $\Lambda(t)$. The dynamics follow a shifted Bautin normal form
\begin{equation}
\frac{dz}{dt} = (a + i\omega)\bigl(z - \Lambda(t)\bigr)
- b\,|z - \Lambda(t)|^2 \bigl(z - \Lambda(t)\bigr)
+ |z - \Lambda(t)|^4 \bigl(z - \Lambda(t)\bigr)
+ \sigma_z \,\xi_z(t),
\end{equation}
where $a$ controls linear growth, $\omega$ is the angular frequency, $b$ is the cubic coefficient, $\sigma_z$ sets the noise amplitude, and $\xi_z(t)$ is complex-valued white noise. The external forcing evolves according to
\begin{equation}
\Lambda(t) = \frac{\Lambda_{\max}}{2}
\left[
\tanh\left(\frac{\Lambda_{\max} r t}{2}\right) + 1
\right],
\end{equation}
where $\Lambda_{\max}$ determines the forcing magnitude and $r$ controls the rate of change.

The forcing $\Lambda(t)$ translates the underlying Bautin system through state space. For sufficiently large $r$, the system fails to track the quasi-static attractor and undergoes rate-induced tipping, even though the instantaneous dynamics remain locally stable. We use parameters $a=0.1$, $\omega=3$, $b=1$, $\sigma_z=0.2$, and $\Lambda_{\max}=8$. Tipping runs are generated with $r=0.10$, while non-tipping runs use $r=0.05$.

For analysis, we consider the scalar observable
\begin{equation}
\rho(t) = \sqrt{x(t)^2 + y(t)^2},
\end{equation}
and define tipping as the event $\rho(t) \geq 10$.

\paragraph{r\_compost\_bomb} We consider a stochastic compost-bomb model describing thermally driven instability in a coupled temperature–carbon system~\cite{luke2011soil,wieczorek2011excitability}. The dynamics are given by
\begin{align}
\mu \frac{dT_s}{dt} &= -\lambda (T_s - T_a(t)) + A C_s r_0 e^{(\alpha (T_s - T_{\mathrm{ref}}))} + D_3 \,\xi(t), \\
\frac{dC_s}{dt} &= \Pi - C_s r_0 e^{(\alpha (T_s - T_{\mathrm{ref}}))},
\end{align}
where $T_s(t)$ denotes soil temperature and $C_s(t)$ the soil carbon content. The external forcing is given by a linearly increasing atmospheric temperature
\begin{equation}
T_a(t) = v t,
\end{equation}
where $v$ controls the rate of forcing. The nonlinear term
\[
r(T_s) = r_0 e^{(\alpha (T_s - T_{\mathrm{ref}}))}
\]
represents temperature-dependent microbial respiration, giving rise to a positive feedback loop in which increasing temperature accelerates respiration, releasing heat and further increasing temperature.

For sufficiently small forcing rates $v$, the system tracks a quasi-static equilibrium corresponding to stable temperature–carbon states. However, when $v$ exceeds a critical threshold, the system fails to track this equilibrium branch and undergoes rapid thermal runaway, despite remaining locally stable at each instantaneous forcing level. This behaviour constitutes rate-induced tipping.

We use parameters $\mu = 2.5 \times 10^6$, $\lambda = 5.049 \times 10^6$, $A = 3.9 \times 10^7$, $\Pi = 1.055$, $r_0 = 0.01$, and $\alpha = \log(2.5)/10$, with noise magnitude $D_3 = 50$. Tipping and non-tipping regimes are generated by varying the forcing rate $v$. Tipping is identified when $T_s > 30$, indicating entry into the runaway regime.

\paragraph{b\_amoc and r\_amoc} We consider a reduced Atlantic Meridional Overturning Circulation (AMOC) box model derived from the five-box formulation in~\cite{ritchie2023rate, alkhayuon2019basin}. By treating the Southern Ocean and bottom water salinities as slow variables and enforcing salt conservation, the system reduces to a two-dimensional stochastic dynamical system with state $\mathbf{x}(t) = (S_N(t), S_T(t))^\top \in \mathbb{R}^2$, where $S_N$ and $S_T$ denote North Atlantic and tropical Atlantic salinity, respectively. The AMOC strength is diagnosed as
\begin{equation}
Q = \frac{\lambda\left[\alpha(T_S - T_0) + \beta(S_N - S_S)\right]}{1 + \lambda \alpha \mu},
\end{equation}
and serves as the observable for tipping detection.

The salinity dynamics are piecewise-defined depending on the sign of $Q$. For $Q \geq 0$,
\begin{align}
V_N \frac{dS_N}{dt} &= Q(S_T - S_N) + K_N(S_T - S_N) - F_N(H(t)) S_0 + \sigma_N \xi_N(t), \\
V_T \frac{dS_T}{dt} &= Q\big(\gamma S_S + (1-\gamma)S_{IP} - S_T\big)
+ K_S(S_S - S_T) \\
&+ K_N(S_N - S_T) - F_T(H(t)) S_0 + \sigma_T \xi_T(t),
\end{align}
while for $Q < 0$,
\begin{align}
V_N \frac{dS_N}{dt} &= |Q|(S_B - S_N) + K_N(S_T - S_N) - F_N(H(t)) S_0 + \sigma_N \xi_N(t), \\
V_T \frac{dS_T}{dt} &= |Q|(S_N - S_T)
+ K_S(S_S - S_T) + K_N(S_N - S_T) - F_T(H(t)) S_0 + \sigma_T \xi_T(t).
\end{align}
Here, $F_N$ and $F_T$ denote freshwater fluxes induced by a hosing forcing $H(t)$, and $\xi_N(t), \xi_T(t)$ are independent Gaussian white noise processes with amplitudes $\sigma_N = \sigma_T = 1.0$.

For quasi-static forcing, the system exhibits b-tipping (fold bifurcation) at a critical freshwater forcing $H^\star \approx 0.4236$, corresponding to the collapse of the overturning circulation~\cite{alkhayuon2019basin}. Tipping trajectories are generated by linearly ramping $H$ from $0.0$ to $0.4736$, crossing the bifurcation threshold, while non-tipping trajectories follow the same ramp but are capped at $H=0.3736$, remaining on the stable branch.

To isolate r-tipping, we use a transient forcing protocol
\begin{equation}
H(r,t)=
\begin{cases}
H_0 + \Delta H\,\mathrm{sech}\!\big(r(t-t_{\mathrm{crit}})\big), & t<t_{\mathrm{crit}},\\
H_0 + \Delta H, & t\ge t_{\mathrm{crit}},
\end{cases}
\end{equation}
such that tipping and non-tipping trajectories differ only through the forcing rate $r$. For sufficiently large $r$, the system fails to track the quasi-static equilibrium and undergoes collapse even though the instantaneous system remains locally stable. We use $r=0.017$ for tipping and $r=0.005$ for non-tipping, with $t_{\mathrm{crit}}=500$.

Tipping is identified as a rapid decline in AMOC strength $Q(t)$, indicating a transition from the strong to the weak circulation state.

\paragraph{SWEC-iEEG} We analyze continuous intracranial electroencephalography (iEEG) recordings from the SWEC-ETHZ database~\cite{burrello2019laelaps}, which comprises 116 expert-annotated seizures across 2,656 hours of multi-channel iEEG from 18 patients with pharmacoresistant epilepsy. Signals were recorded intracranially via strip, grid, and depth electrodes, band-pass filtered between 0.5 and 120\,Hz, and digitized at 512 or 1024\,Hz depending on the patient.
Seizure onset and end times were determined by a board-certified EEG epileptologist. Seizure onset can be interpreted as a bifurcation-induced tipping event: in the Epileptor framework~\cite{jirsa2014nature}, the transition from interictal to ictal dynamics corresponds to a saddle-node bifurcation driven by a slow permittivity variable that modulates neural excitability.
The system remains near a stable resting-state equilibrium during interictal periods, and seizure onset occurs when this equilibrium is annihilated through the bifurcation, triggering a rapid transition to high-amplitude ictal oscillations.
For evaluation, we extract fixed-length segments preceding each annotated seizure onset as critical trajectories, paired with multiple interictal segments, at least 1h away from any seizure, as non-critical controls. %

\begin{figure}
    \centering
    \includegraphics[width=1\linewidth]{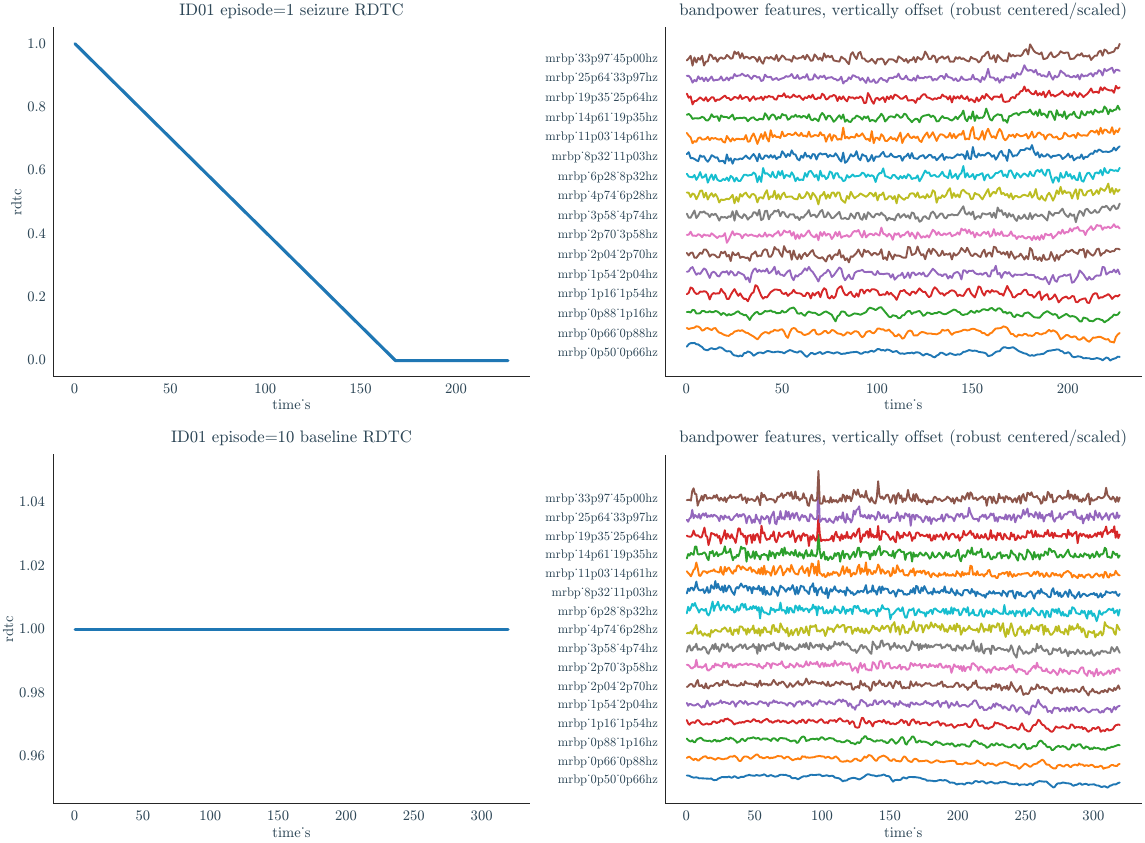}
    \caption{Representative SWEC-iEEG traces around seizure onset and baseline.
    Right-hand side plot shows vertically stacked sub-rows of the $K{=}16$ multi-resolution log-power features, one band per row, with seizure onset at 168\,s.
    The expert-annotated seizure time is used for the surrogate RDTC definition (top row).}
    \label{fig:swec-representative-traces}
\end{figure}

We do not feed raw iEEG into our framework, but a multi-resolution band-power representation that summarises spectral content on a common 1\,s grid, see Fig.~\ref{fig:swec-representative-traces}.
For each patient we down-sample to $C{=}16$ evenly-spaced electrodes and compute log-power in $K{=}16$ contiguous frequency bands whose edges are geometrically spaced from $0.5$ to $45$\,Hz ($\mathrm{geomspace}(0.5, 45, 17)$).
Each band's power is estimated with Welch's method on a causal-trailing window of $\max(2\,\mathrm{s},\, 6/f_{\mathrm{lo}})$ seconds, where $f_{\mathrm{lo}}$ is the band's lower edge (so low-frequency bands integrate longer windows for adequate spectral resolution); we cap \texttt{nperseg} at $1024$ samples and zero-pad the FFT to $2048$ to fix the frequency-grid spacing across bands.
Band power is then obtained by trapezoidal integration of the PSD over the band edges, $\log_{10}$-transformed, and aggregated across the $16$ channels via median.
The resulting $K{=}16$-dimensional time series, sampled at $1$\,Hz, is the input to all methods on SWEC-iEEG.

This dataset provides a challenging real-world benchmark with high dimensionality (36--100 electrodes per patient or representative bandpowers), substantial inter-patient variability, and clinically defined transition times.

\paragraph{TAC} We consider a stochastic thermoacoustic system exhibiting a subcritical Hopf bifurcation, following the experimental and modelling study of Bonciolini et al.~\cite{bonciolini2018experiments}. The system describes the evolution of acoustic pressure amplitude $A(t)$ in a combustion chamber, where thermoacoustic feedback can lead to large-amplitude oscillations. Near the bifurcation, the dynamics can be approximated by a stochastic normal form
\begin{equation}
\frac{dA}{dt} = \mu(t) A - \alpha A^3 + \beta A^5 + \sigma \xi(t),
\end{equation}
where $A(t)$ denotes the oscillation amplitude, $\mu(t)$ is a time-dependent control parameter, $\alpha,\beta > 0$ determine the nonlinear saturation, and $\xi(t)$ is Gaussian white noise. The subcritical nature of the bifurcation implies bistability between a low-amplitude (stable) state and a high-amplitude limit cycle.

The control parameter $\mu(t)$ is ramped in time, corresponding to changes in operating conditions of the combustor. In the quasi-static case, the system undergoes a bifurcation-induced transition when the stable equilibrium disappears. However, for finite ramping rates, the system exhibits a rate-dependent delay: the transition to large-amplitude oscillations occurs beyond the quasi-static bifurcation point, with the delay increasing as the rate of change increases~\cite{bonciolini2018experiments}. This phenomenon leads to dynamic hysteresis when the parameter is ramped forward and backward.

\begin{figure}
    \centering
    \includegraphics[width=1\linewidth]{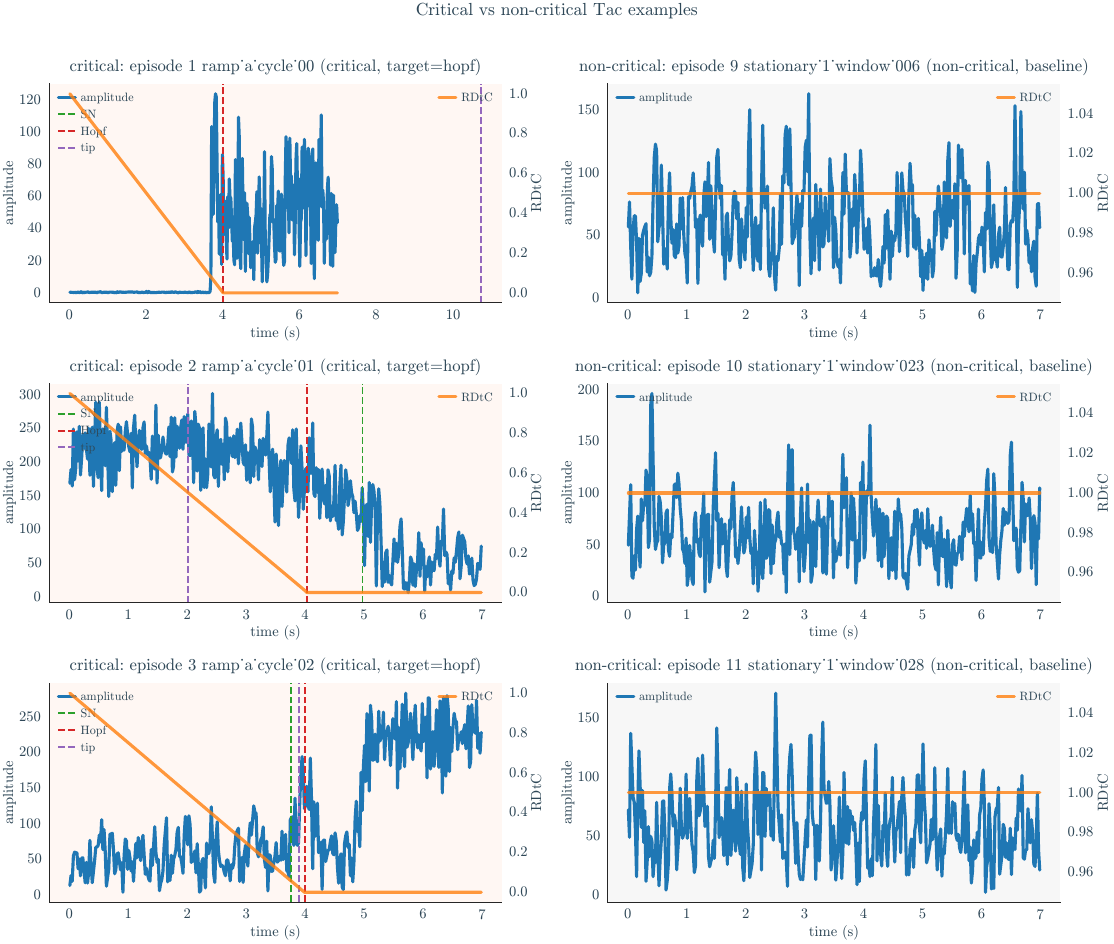}
    \caption{Examples of critical and non-critical \texttt{TAC} datasets trajectories. Dashed lines indicate the varios transitions, of which we use the Hopf bifurcation as surrogate RDTC target.}
    \label{fig:tac-examples}
\end{figure}

For the TipPFN benchmark, the query trajectories are real experimental episodes extracted from the Bonciolini pressure recordings, rather than simulated test trajectories. We use the Mic1 pressure channel sampled at 2\,kHz, extract a 100--200\,Hz band-pass Hilbert amplitude envelope, and construct a balanced real-query pool of ramp-up tipping episodes and low-state stationary baseline episodes; see Fig.~\ref{fig:tac-examples} for example trajectories.
The additional three microphone channels are processed analogously and provided as features.
We then provide real and simulated context data for up to four context episodes.
RDTC is defined with respect to the fitted Hopf crossing; the actual forcing $\mu(t)$ is not provided to the models.
This setting combines bifurcation structure, finite-rate forcing, and stochastic transition variability, making it a stringent sim-to-real test of whether TipPFN can use limited context to forecast criticality in noisy laboratory trajectories.

\paragraph{DaphniaExt} We use experimental observations of population collapse from controlled microcosm experiments reported in~\cite{drake2010early}. The system consists of replicate populations of \textit{Daphnia magna} subjected to gradually deteriorating environmental conditions through sustained reductions in food availability. The resulting population dynamics can be interpreted through a minimal stochastic growth model with time-dependent parameters,
\begin{equation}
    \frac{dN}{dt} = r(t)\,N + \sigma \xi(t),
\end{equation}
where $N(t)$ denotes population size, $r(t)$ is a time-varying growth rate reflecting environmental deterioration, and $\xi(t)$ is Gaussian white noise~\cite{ricker1954logistic}.

As environmental conditions worsen, $r(t)$ decreases and crosses zero, corresponding to a transcritical bifurcation in which the stable positive population state loses stability and extinction becomes inevitable. The observed dynamics exhibit a transition from stationary fluctuations around a stable equilibrium to a persistent decline toward extinction. Tipping is identified as the onset of irreversible population decline.

\begin{figure}
    \centering
    \includegraphics[width=1\linewidth]{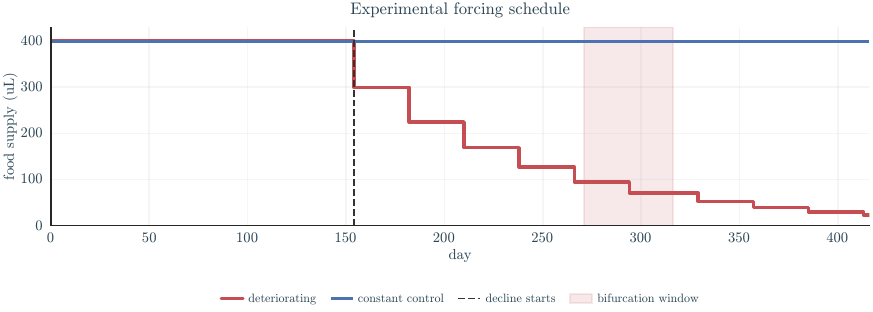}
    \caption{Experimental forcing schedule and estimated bifurcation range~\cite{drake2010early} for the \texttt{DaphniaExt} dataset.}
    \label{fig:daphnia-forcing-schedule}
\end{figure}

\begin{figure}
    \centering
    \includegraphics[width=1\linewidth]{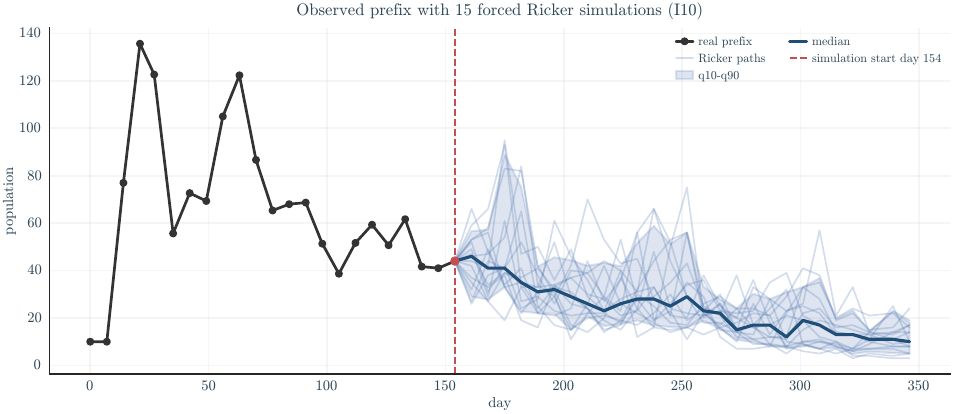}
    \caption{Initial empirical \texttt{DaphniaExt} observations, extended with simulated stochastic growth model trajectories which are used for synthetic context trajectories.}
    \label{fig:daphnia-ricker-forced}
\end{figure}
\begin{figure}
    \centering
    \includegraphics[width=1\linewidth]{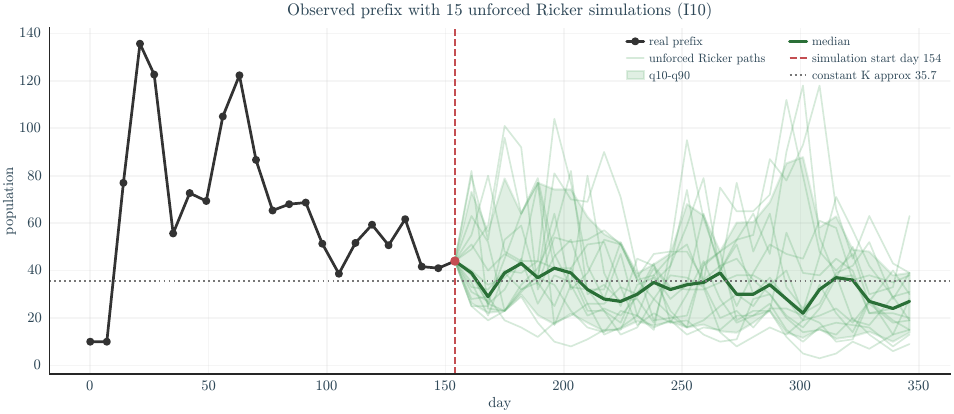}
    \caption{Like above, but without forcing, corresponding to a non-critical trajectory.}
    \label{fig:daphnia-ricker-unforced}
\end{figure}

\begin{figure}[p]
    \centering
    \includegraphics[width=1\linewidth]{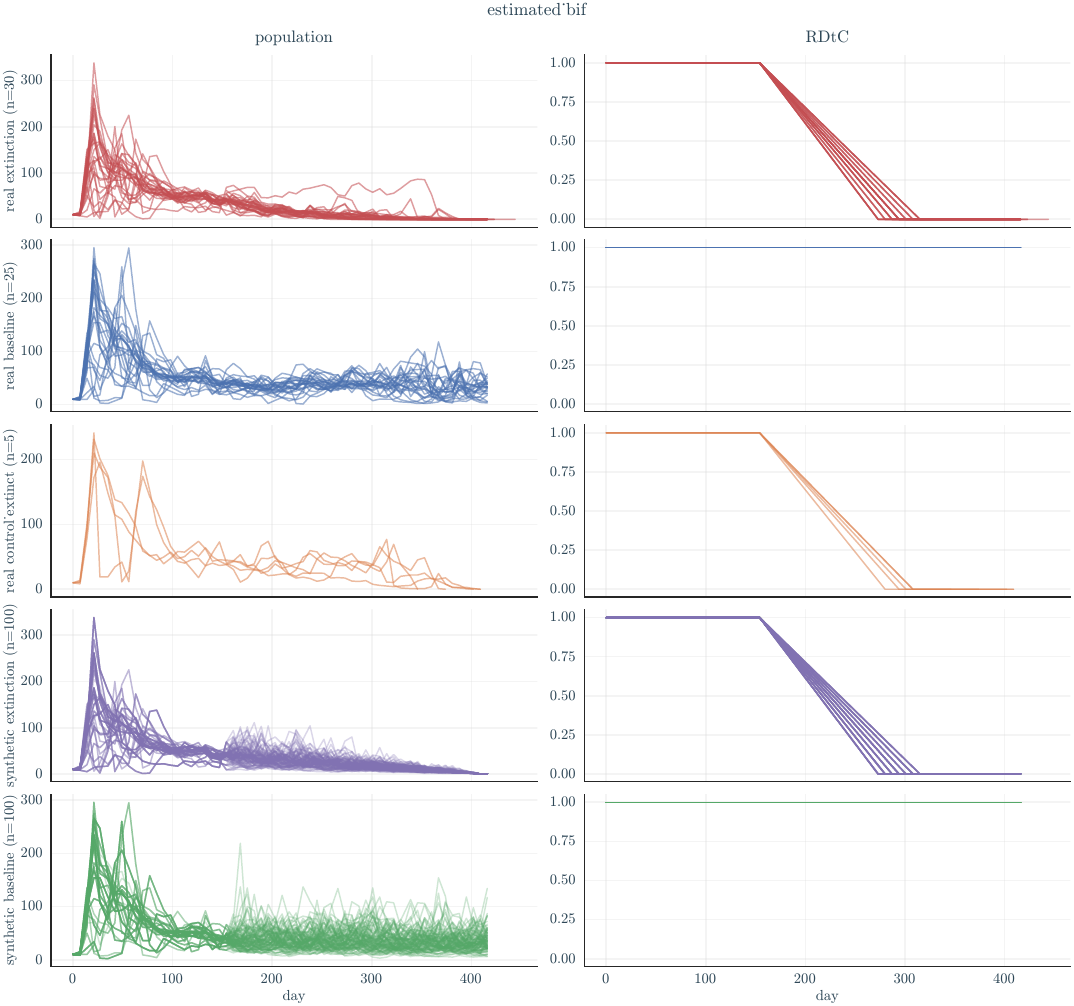}
    \caption{Real and synthetic \texttt{DaphniaExt} episodes with corresponding heuristic RDTC ramps, anchored at a random point within the estimated bifurcation range, see Fig.~\ref{fig:daphnia-forcing-schedule}.}
    \label{fig:daphnia-estimated-bif}
\end{figure}
\begin{figure}[p]
    \centering
    \includegraphics[width=1\linewidth]{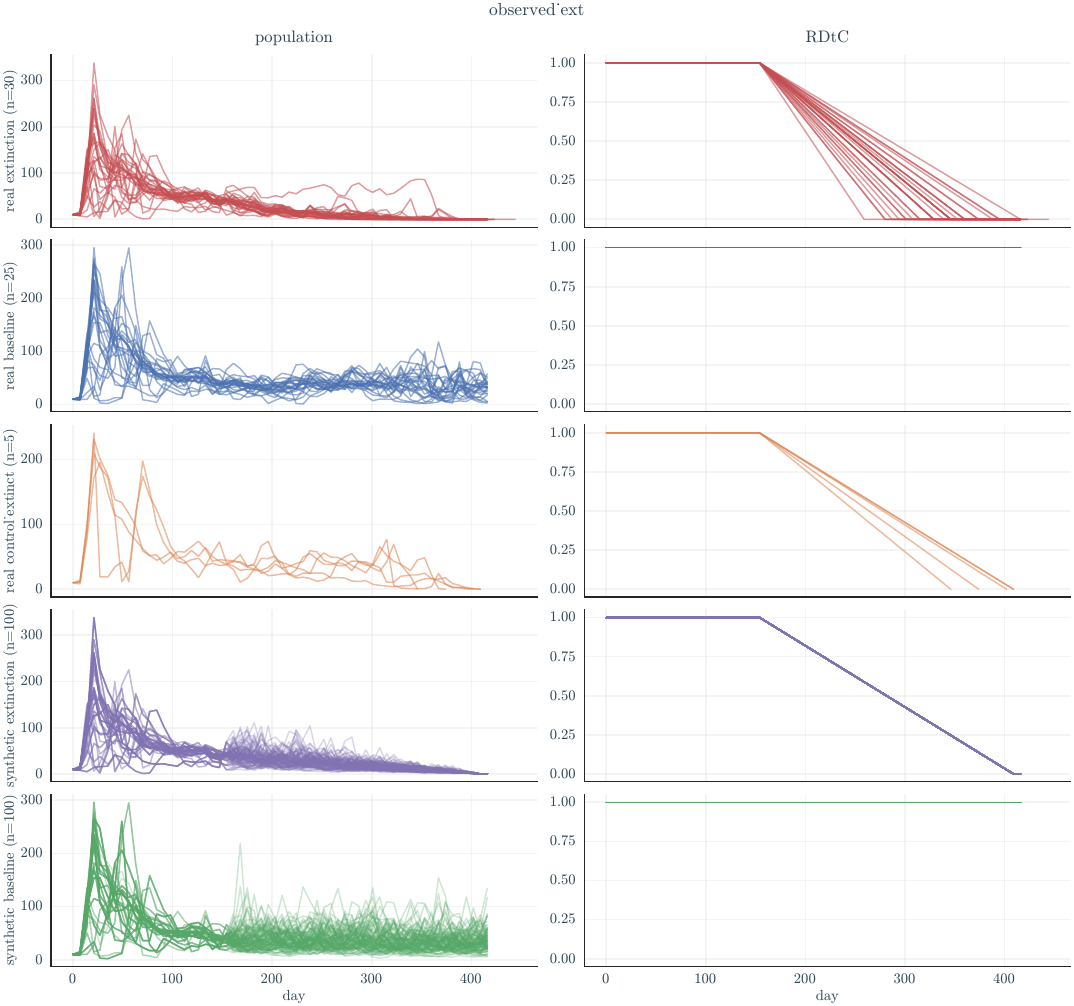}
    \caption{Like above, but with the $\Lambda = 0$ point set to the actual extinction.}
    \label{fig:daphnia-observed-ext}
\end{figure}

\FloatBarrier
\newpage
\paragraph{microcosm} We analyze a cyanobacteria population in chemostats subjected to dilution events under gradually increasing light levels~\cite{veraart2012recovery}, using the data compiled by~\cite{grziwotz2023anticipating}. Population density is inferred from the light attenuation coefficient, computed from continuous measurements of outgoing light intensity. The full dataset comprises 7,784 observations over 28.86 days at a sampling interval of 5 minutes. The time series is divided into six segments separated by dilution events, which are external perturbations and not part of the intrinsic population dynamics. To ensure consistency, we restrict analysis to recovery phases between dilution events. Specifically, for each segment, we use the final 250 time points (approximately one day) prior to the next dilution event, yielding a total of 1,500 data points across all segments.

\paragraph{voice} We analyze experimental time series of phonation onset, corresponding to the emergence of vocal fold oscillations, which can be interpreted as a Hopf bifurcation~\cite{mergell1998phonation,murray2012vibratory}. The data are obtained from a physical replica of human vocal folds (EPI model) with a multilayer body–cover structure that reproduces realistic tissue mechanics. Oscillations are induced by gradually increasing airflow, while subglottal pressure is recorded using a pressure transducer. The available time series, from the data compiled by~\cite{grziwotz2023anticipating}, consists of 7,501 data points over 0.375\,s with a sampling interval of $5 \times 10^{-5}$\,s. Tipping corresponds to the transition from a stable, non-oscillatory state to sustained oscillations as the flow rate crosses a critical threshold. We use the measured pressure signal as the observable for tipping detection.

\paragraph{cellular\_atp} We analyze experimental time series of cytosolic ATP (i.e., the readily available energy pool) dynamics in living plant cells, measured via a genetically encoded FRET sensor sensitive to MgATP$^{2-}$~\cite{wagner2019multiparametric}, using the data compiled by~\cite{grziwotz2023anticipating}. The data capture the response of leaf tissue under gradually increasing hypoxia, induced by sealing samples in a dark environment where respiration depletes available oxygen. ATP levels remain relatively stable during oxygen decline and then undergo a sudden collapse, indicating a critical transition in cellular energy state. The dataset consists of 271 observations over 3 minutes with a sampling interval of 0.011 minutes. We use the fluorescence ratio signal as the observable, with tipping identified as the abrupt drop in ATP concentration.

\paragraph{greenhouse\_earth} We analyze a paleoclimate time series of calcium carbonate (CaCO$_3$) associated with the end of greenhouse Earth, marking the transition from an ice-free state to the formation of polar ice caps~\cite{dakos2008slowing}, using the data compiled by~\cite{grziwotz2023anticipating}. The data exhibit increasing autocorrelation prior to the climate shift, consistent with critical slowing down. The dataset consists of 462 observations spanning 5.9 million years with a sampling interval of 0.013 million years. Tipping is identified as the transition to a glaciated climate state.

\paragraph{blackout\_frequency} We analyze bus voltage frequency data preceding the Western Interconnect blackout of August 1996, measured within the Bonneville Power Administration network~\cite{council1996western}, using the data compiled by~\cite{grziwotz2023anticipating}. The time series captures system dynamics leading up to grid separation and exhibits critical fluctuations prior to the blackout, consistent with EWS of instability. The dataset consists of 11,301 observations over 565 seconds with a sampling interval of 0.05\,s. Tipping is identified as the transition to system-wide failure, i.e., blackout.

\paragraph{AMOC} The RAPID-MOCHA-WBTS dataset~\cite{moat2024atlantic} provides continuous measurements of the AMOC at 26.5°N, derived from current velocity, temperature, salinity, and pressure observations collected by a trans-basin mooring array across the Atlantic and cable measurements across the Florida Straits. These variables are used to compute oceanic volume and heat transports in both depth and density space, yielding a 12-hourly, 10-day low-pass filtered transport time series spanning April 2nd 2004 to March 27th 2024.

\subsection{Baselines}\label{si-sec:baselines}
In addition to the classical indicators such as AR1 and variance, we provide additional details on the comprehensive set of DL-based baselines included in this work.

\paragraph{TabPFN (v2.6)} We include TabPFN~\cite{hollmann2023tabpfntransformersolvessmall,hollmann2025accurate} as a state-of-the-art ICL baseline. TabPFN is a Prior-Data Fitted Network (PFN), a transformer model trained on large amounts of synthetically generated tabular data to approximate Bayesian inference via in-context learning. In our setting, we adapt TabPFN to time series by representing each trajectory segment as a tabular sample with engineered features (e.g., recent observations or summary statistics), and evaluate its ability to predict RDTC. We evaluate TabPFN's performance for an increasing number of context episodes (1, 2, 3). 
Zero-context datasets are skipped for
TabPFN because the row-wise regressor requires observed target rows.

The context information is transformed into a tabular regression problem suitable for TabPFN in the following way: Let the processed single-sample batch have row index $r=1,\ldots,R$, identifier
columns $\iota_r=(e_r,t_r)$, selected feature vector $x_r$, action vector
$y_r=(y_{r,1},\ldots,y_{r,A})$, action mask
$m_r=(m_{r,1},\ldots,m_{r,A})$, context indicator $c_r$, and valid-row indicator
$v_r$. The first action slot is transformed RDTC, $y_{r,1}=\Lambda^*_r$. The
row-wise TabPFN feature vector is
\begin{equation}
    X^{\mathrm{Tab}}_r
    =
    \left[
        \iota_r,\,
        x_r,\,
        y_{r,2:A},\,
        m_{r,2:A},\,
        c_r,\,
        v_r
    \right],
    \qquad
    Y^{\mathrm{Tab}}_r = y_{r,1}.
\label{eq:tabpfn-row-layout}
\end{equation}
The supervised fit set and prediction set are
\begin{equation}
    \mathcal{T}
    =
    \{r: v_r=1,\ m_{r,1}=0,\ Y^{\mathrm{Tab}}_r\ \mathrm{finite}\},
    \qquad
    \mathcal{Q}
    =
    \{r: v_r=1,\ c_r=0,\ m_{r,1}=1\}.
\label{eq:tabpfn-fit-query-sets}
\end{equation}
Non-finite entries in $X^{\mathrm{Tab}}$ are passed as missing values. TabPFN is
fit separately for each query episode on $\mathcal{T}$ and predicts only at
$\mathcal{Q}$.
The model is evaluated using default parameters. It returns logits
$\ell_{q,j}$ over histogram bins and bin borders $b_0,\ldots,b_B$. With
$p_{q,j}=\operatorname{softmax}(\ell_q)_j$,
$C_{q,j}=\sum_{k=1}^j p_{q,k}$, and $C_{q,0}=0$, the inverse-CDF quantile is
\begin{equation}
    \hat Q_q(\alpha)
    =
    b_{j-1}
    +
    \frac{\alpha-C_{q,j-1}}{\max(p_{q,j},10^{-12})}
    (b_j-b_{j-1}),
    \qquad
    j=\min\{k:C_{q,k}\geq\alpha\}.
\label{eq:tabpfn-inverse-cdf}
\end{equation}

\paragraph{Bury} We include the deep learning approach of \cite{bury2021deep}, which uses a CNN-LSTM architecture trained on simulated time series from canonical dynamical systems (Hopf, transcritial and fold). The model predicts the probability of an upcoming tipping point directly from observed trajectories in a single forward pass, without relying on handcrafted indicators. As it is trained on bifurcation normal forms, the method is primarily designed for b-tipping and has previously not been generalize to r- or n-induced transitions.

\paragraph{Zhuge} We include the deep learning approach of \cite{zhuge2025deep}, which builds on \cite{bury2021deep} using a CNN-LSTM architecture trained on simulated time series from canonical bifurcation normal forms. The model learns a direct mapping from observed trajectories (and control parameters) to the predicted tipping point, and can handle irregularly sampled data. As it is trained on normal-form dynamics, the method (like \cite{bury2021deep}) is primarily designed for B-tipping and relies on shared features of local bifurcation structure.

\paragraph{Huang}
We include the deep learning approach of \cite{huang2024deep}, which develops a CNN-based classifier for predicting R-tipping under time-varying forcing and stochastic perturbations. Unlike Bury and Zhuge, this method is explicitly designed for r-tipping and predicts the probability that an observed trajectory will tip, rather than relying on CSD indicators. The original study trains and evaluates separate models on three prototypical R-tipping systems: a saddle-node system, a Bautin system, and a compost-bomb model. It further considers extrapolation to out-of-sample forcing rates, showing that models trained at one forcing rate can generalize to unseen rates in some systems, while performance is system-dependent. In our experiments, we use this method in a stronger cross-system setting: the model is trained only on the saddle-node r-tipping system and then evaluated on other R-tipping systems. This differs from the original setup, where separate system-specific training was used. We adopt this protocol because the publicly released code and data~\cite{huang2024code} primarily provide the saddle-node training workflow, limiting direct replication of the full multi-system training procedure for our benchmark.

\paragraph{Comparison of the learned methods} Table~\ref{tab:si-method-comparison} summarizes how the learned methods differ in training data, task and inference setting.

\begin{table}[h]
\centering
\scriptsize
\setlength{\tabcolsep}{3pt}
\caption{\textbf{Comparison of the learned methods.}
\emph{Inference setting} describes the use in this work; in Table~\ref{tab:auroc-results-best-col} all methods receive the same query window of at most 128 steps, which Bury, Zhuge and Huang bring to their native input length.
Parameter counts are learned parameters per network, read from the checkpoints.}
\label{tab:si-method-comparison}
\begin{tabularx}{\linewidth}{@{}>{\raggedright\arraybackslash}p{1.45cm}*{5}{>{\raggedright\arraybackslash}X}@{}}
\toprule
& Bury~\cite{bury2021deep} & Zhuge~\cite{zhuge2025deep} & Huang~\cite{huang2024deep} & TabPFN 2.6~\cite{hollmann2025accurate} & TipPFN \\
\midrule
Driver construction & Random 2-D polynomial systems, bifurcation parameter increased towards a fold, Hopf or transcritical point; null runs & Bifurcation normal forms with a varying control parameter & Saddle-node system with forcing ramped at different rates & Generic synthetic tabular prior; no time series & Random 2-D polynomial systems as in Bury (fold, Hopf, transcritical); Bezier, ramp or constant forcing (App.~\ref{si-sec:forcing_datasets}, \ref{si-sec:synthetic-forcing-schedules}) \\
\addlinespace
System embedding & None & None & None & -- & 14-variable random SDE network driven by the driver through external inputs (App.~\ref{si-sec:synthetic-augmenting-sde}) \\
\addlinespace
Observation model & One state variable, detrended & One state variable and the control parameter, irregular sampling & One state variable & Table rows of per-time-step features (Eq.~\eqref{eq:tabpfn-row-layout}) & Random subset of 1--16 driver and auxiliary variables, resampled in time \\
\addlinespace
Task & Classification & Regression & Binary classification & In-context regression & In-context distributional regression \\
\addlinespace
Target & Upcoming bifurcation type or none & Control value at the transition & Tipping vs.\ no tipping & Transformed RDTC & Quantiles of transformed RDTC at every time step \\
\addlinespace
Inference setting & Univariate, 500 steps; no context; mean of 6 networks & Univariate plus forcing parameter (privileged), 500 steps; no context; mean of 10 networks & Univariate, 680 steps; no context; one network & Fitted per query episode on 1--2 context episodes & Up to 16 features, $\le$128 steps; 0--2 context episodes; one forward pass \\
\addlinespace
Parameters & 23{,}334 & 41{,}721 & 915{,}074 & 12.9M & 210.7M \\
\bottomrule
\end{tabularx}
\end{table}

\subsection{Metrics}\label{si-sec:metrics}
We describe the metrics used in this work. For tipping point detection, we consider a binary classification setting where each sample is labeled as either tipping ($y=1$) or non-tipping ($y=0$). Let $\hat{s}: \mathcal{D} \to \mathbb{R}$ denote a scoring function that assigns a confidence score to each observed trajectory, where higher values indicate a higher likelihood of an impending tipping event. A prediction $\hat{y}=1$ is made whenever $\hat{s} > \tau$ for a threshold $\tau$.

For each threshold $\tau$, we define the true positive rate (TPR) and false positive rate (FPR) as
\begin{equation}
\text{TPR}(\tau) = \frac{|\{i : y_i = 1,\ \hat{s}_i > \tau\}|}{|\{i : y_i = 1\}|}, \quad
\text{FPR}(\tau) = \frac{|\{i : y_i = 0,\ \hat{s}_i > \tau\}|}{|\{i : y_i = 0\}|}.
\end{equation}

The receiver operating characteristic (ROC) curve \cite{fawcett2006introduction,hanley1982meaning} is obtained by plotting $\text{TPR}(\tau)$ against $\text{FPR}(\tau)$ as the threshold $\tau$ varies over all possible values. The ROC curve characterizes the trade-off between correctly detecting tipping events and incorrectly classifying non-tipping trajectories.

The Area Under the ROC Curve (AUROC) \cite{peterson1954theory,bradley1997use} is defined as
\begin{equation}
\text{AUROC} = \int_0^1 \text{TPR}\bigl(\text{FPR}^{-1}(u)\bigr)\,du,
\end{equation}
which corresponds to the probability that a randomly chosen tipping trajectory is assigned a higher score than a randomly chosen non-tipping trajectory. An AUROC of $0.5$ indicates random performance, while $1.0$ corresponds to perfect discrimination.

\newpage
\subsubsection{Scores}
Table~\ref{tab:si-score-definitions} gives an overview of computed scores for each model and baseline.

\begin{table}[h]
\centering
\caption{Score definitions across all methods. TipPFN and TabPFN expose six heads per scope: two distributional summaries of the predicted RDtC quantile distribution ($\texttt{1-mean}$, $\texttt{1-median}$) and four CDF probabilities at thresholds $\tau \in \{0.05, 0.1, 0.2, 0.3\}$. The model emits RDtC in $\tanh$-bounded coordinates; raw thresholds $\tau$ in physical RDtC units are mapped by $\tilde{\tau} = \tanh(\tau / 0.2)$. Both scopes use the last query position in their respective scope (forecast horizon aggregation if a horizon row is present).
Baseline score names follow $\texttt{<method>:<short\_mode>:<scope>:<head>}$ where $\texttt{<scope>}$ is $\texttt{w}N$ for a fixed query window of length $N$ or $\texttt{whist}$ for the whole available history; $\texttt{<short\_mode>} \in \{\texttt{pad}, \texttt{backfill}, \texttt{resample}\}$ controls how short signals are extended to the model's input length. EWS does not consume $\texttt{short\_mode}$; Zhuge additionally carries a $\texttt{<control\_source>} \in \{\texttt{param}, \texttt{rdtc}\}$ slot. Higher score $\Rightarrow$ closer to tipping for every entry except where noted.}
\label{tab:si-score-definitions}
\resizebox{\columnwidth}{!}{%
\begin{tabular}{@{}l l p{6.5cm}@{}}
\toprule
Score name & Definition & Notes \\
\midrule
\multicolumn{3}{@{}l}{\emph{TipPFN, TabPFN — nowcast scope (last nowcast query position)}}\\
\texttt{<m>:nc:1-median} & $1 - \mathrm{median}(Q_{\mathrm{nc}})$ & $Q_{\mathrm{nc}}$: predicted RDtC quantile distribution at the last nowcast query \\
\texttt{<m>:nc:1-mean} & $1 - \mathbb{E}[Q_{\mathrm{nc}}]$ & Trapezoidal integration of the predicted quantiles over $p \in [0.01, 0.99]$, without the tail mass \\
\texttt{<m>:nc:P(rdtc<$\tau$)} & $\Pr(Q_{\mathrm{nc}} < \tilde{\tau})$ & $\tilde{\tau} = \tanh(\tau/0.2)$, $\tau \in \{0.05, 0.1, 0.2, 0.3\}$ \\[2pt]
\multicolumn{3}{@{}l}{\emph{TipPFN, TabPFN — forecast scope (forecast query position(s))}}\\
\texttt{<m>:fc:1-median} & $1 - \mathrm{median}(Q_{\mathrm{fc}})$ & Last forecast position; multi-horizon runs aggregate per-horizon scores \\
\texttt{<m>:fc:1-mean} & $1 - \mathbb{E}[Q_{\mathrm{fc}}]$ & As above; same scope rules as $\texttt{fc:1-median}$ \\
\texttt{<m>:fc:P(rdtc<$\tau$)} & $\Pr(Q_{\mathrm{fc}} < \tilde{\tau})$ & $\tilde{\tau}$, $\tau$ as in the nowcast row \\
\midrule
\multicolumn{3}{@{}l}{\emph{EWS\cite{dakos2008slowing,dakos2012robustness}} — rolling-window indicator with Kendall-$\tau$ trend test}\\
\texttt{dews:w$N$:var\_tau} & $\tau\bigl(\mathrm{Var}_{w}(x_t)\bigr)$ & Kendall-$\tau$ of rolling variance over window of length $N$ \\
\texttt{dews:w$N$:ar1\_tau} & $\tau\bigl(\mathrm{AR}(1)_{w}(x_t)\bigr)$ & Kendall-$\tau$ of rolling lag-1 autocorrelation \\
\texttt{dews:w$N$:acf\_tau} & $\tau\bigl(\mathrm{ACF}_{w}(x_t)\bigr)$ & Kendall-$\tau$ of rolling autocorrelation function (lag 1, window) \\
\texttt{dews:w$N$:skw\_tau} & $\tau\bigl(\mathrm{Skew}_{w}(x_t)\bigr)$ & Kendall-$\tau$ of rolling skewness \\
\texttt{dews:w$N$:lambd\_tau} & $\tau\bigl(\hat{\lambda}_{w}(x_t)\bigr)$ & Kendall-$\tau$ of rolling local return rate $\hat{\lambda}$ \\
\midrule
\multicolumn{3}{@{}l}{\emph{Bury \cite{bury2021deep}} — CNN classifier over $\{\text{fold}, \text{hopf}, \text{transcritical}, \text{null}\}$}\\
\texttt{bury:<sm>:w$N$:prob\_tip} & $\Pr(\text{tip}) = 1 - \Pr(\text{null})$ & Sum of fold + hopf + transcritical class probabilities \\
\texttt{bury:<sm>:w$N$:prob\_fold} & $\Pr(\text{fold})$ & Per-class softmax output \\
\texttt{bury:<sm>:w$N$:prob\_hopf} & $\Pr(\text{hopf})$ & Per-class softmax output \\
\texttt{bury:<sm>:w$N$:prob\_transcritical} & $\Pr(\text{transcritical})$ & Per-class softmax output \\
\midrule
\multicolumn{3}{@{}l}{\emph{Huang \cite{huang2024deep}} — CNN classifier (tip vs.\ no-tip)}\\
\texttt{huang:<sm>:w$N$:prob\_tip} & $\Pr(\text{tip})$ & Single-output sigmoid probability of an upcoming tipping event \\
\midrule
\multicolumn{3}{@{}l}{\emph{Zhuge \cite{zhuge2025deep}} — neural regressor for the tipping bifurcation parameter}\\
\texttt{zhuge:<cs>:<sm>:w$N$:signed\_tip\_margin} & $\Delta_{\mathrm{tip}}^{\pm} = -\hat{p}_{\mathrm{rem}} / |p_{\Delta}|$ & Normalized signed margin to the bifurcation parameter; identical for $\texttt{<cs>} \in \{\texttt{param}, \texttt{rdtc}\}$ \\
\bottomrule
\end{tabular}%
}
\end{table}

\FloatBarrier
\newpage
\subsection{Further Results}\label{si-sec:results}
In this section, we provide additional information to the conducted experiments and their results.

\subsubsection{Auxiliary variables driven by the state variables only}\label{si-sec:aux-drive-sensitivity}
In the training set and in the synthetic evaluation sets, the 14 auxiliary variables are driven by the state variables $x$, $y$ of the canonical driver, by its forcing parameter $\lambda$ and by the RDTC $\Lambda$ (Appendix~\ref{si-sec:synthetic-augmenting-sde}).
They therefore carry a noisy, filtered readout of the forcing level: a linear regression on the auxiliary variables of one sample recovers $\Lambda$ with a median $R^2$ of .93 on the semi-real b-tipping set and .96 on the canonical set, compared with .18 and .57 from $x$ and $y$.
Methods that observe the auxiliary variables, TipPFN and TabPFN, thus receive information about the forcing that the univariate baselines do not, as Zhuge does through its forcing input.

To quantify the effect, we regenerated the three synthetic evaluation sets with the auxiliary variables driven by $x$ and $y$ only, keeping systems, parameters, seeds and forcing schedules, and repeated the evaluation of Table~\ref{tab:auroc-results-best-col} with the released model (Table~\ref{tab:si-aux-drive}).
Averaged over all datasets, TipPFN drops from .661/.782/.838 to .586/.606/.644 (0/1/2 context episodes) and TabPFN from .523/.667 to .500/.535, while the baselines, which observe only $x$, change by at most .03.
TipPFN with two context episodes remains the best column overall and on the three canonical systems, but its lead over the best baseline shrinks from .23 to .04, and baselines score higher on all five semi-real b-tipping systems and on the four r-tipping systems.
In the regenerated r-tipping set, the non-critical episodes also follow the critical forcing profile until their ramp onset, unlike the schedules of Appendix~\ref{si-sec:forcing}, so these rows differ from Table~\ref{tab:auroc-results-best-col} in both the auxiliary drive and the forcing; with non-critical windows placed at the same position relative to their ramp onset, the averages over all datasets are .638/.614/.651 for TipPFN and .644 for Bury.
Because the released model was trained on data with the same wiring, these numbers measure how much it relies on the forcing information in the auxiliary variables, not the accuracy a model trained without it would reach.

\begin{table}[h]
\centering
\caption{\textbf{AUROC with auxiliary variables driven by the state variables only.} Balanced AUROC averaged over $\Delta>0$ on the synthetic evaluation sets regenerated with the auxiliary variables driven by $x$ and $y$ only, with the protocol, model and column selection of Table~\ref{tab:auroc-results-best-col} (EWS now selects $W=128$ instead of $W=64$). The real-world rows are unchanged (EWS at $W=128$: SWEC-iEEG .537, TAC .568) and enter the \emph{All datasets} row. The r-tipping rows use the regenerated forcing design described in the text. Best value per row in bold.}
\label{tab:si-aux-drive}
\resizebox{\columnwidth}{!}{%
\begin{tabular}{lccccccccc}
\toprule
System & \shortstack[c]{\strut TipPFN\\\strut 0c} & \shortstack[c]{\strut TipPFN\\\strut 1c} & \shortstack[c]{\strut TipPFN\\\strut 2c} & \shortstack[c]{\strut TabPFN\\\strut 1c} & \shortstack[c]{\strut TabPFN\\\strut 2c} & \shortstack[c]{\strut EWS\\\strut $\tau_{\mathrm{var}}$} & \shortstack[c]{\strut Bury~\cite{bury2021deep}\\\strut $P_{\mathrm{tip}}$} & \shortstack[c]{\strut Huang~\cite{huang2024deep}\\\strut $P_{\mathrm{tip}}$} & \shortstack[c]{\strut Zhuge~\cite{zhuge2025deep}\\\strut $\Delta_{\mathrm{tip}}^{\pm}$} \\
\midrule
\multicolumn{10}{l}{\emph{Canonical}} \\
B-Fold & .862 & .863 & \textbf{.902} & .509 & .569 & .655 & .704 & .452 & .524 \\
B-Hopf & .823 & .861 & \textbf{.934} & .568 & .685 & .769 & .818 & .484 & .645 \\
B-Trans. & .709 & .716 & \textbf{.744} & .508 & .599 & .639 & .721 & .461 & .453 \\
\midrule
\multicolumn{10}{l}{\emph{Semi-real}} \\
B-Harv. & .837 & .862 & .922 & .619 & .667 & \textbf{.991} & .967 & .874 & .734 \\
B-RM TC & .630 & .550 & .644 & .411 & .434 & .732 & \textbf{.789} & .615 & .573 \\
B-RM Hopf & .570 & .533 & .565 & .414 & .406 & .616 & \textbf{.641} & .486 & .422 \\
B-SEIRx & .476 & .500 & .508 & .506 & .506 & \textbf{.540} & .473 & .508 & .377 \\
B-AMOC & .869 & .893 & .934 & .619 & .829 & .935 & \textbf{.948} & .509 & .679 \\
R-Bautin & .334 & .416 & .410 & .453 & .411 & .339 & .458 & \textbf{.580} & .208 \\
R-SN & .299 & .382 & .368 & .442 & .393 & .393 & .291 & \textbf{.615} & .100 \\
R-Compost & .542 & .424 & .448 & .514 & .517 & .477 & \textbf{.573} & .433 & .360 \\
R-AMOC & .231 & .407 & .413 & .416 & .389 & .251 & .249 & \textbf{.589} & .147 \\
\midrule
\emph{All datasets} & .586 & .606 & \textbf{.644} & .500 & .535 & .599 & .609 & .535 & .435 \\
\emph{All datasets, Table~\ref{tab:auroc-results-best-col}} & .670 & .779 & \textbf{.838} & .522 & .666 & .572 & .604 & .534 & .443 \\
\bottomrule
\end{tabular}
}
\end{table}

\begin{figure}[h]
    \centering
    \includegraphics[width=1\linewidth]{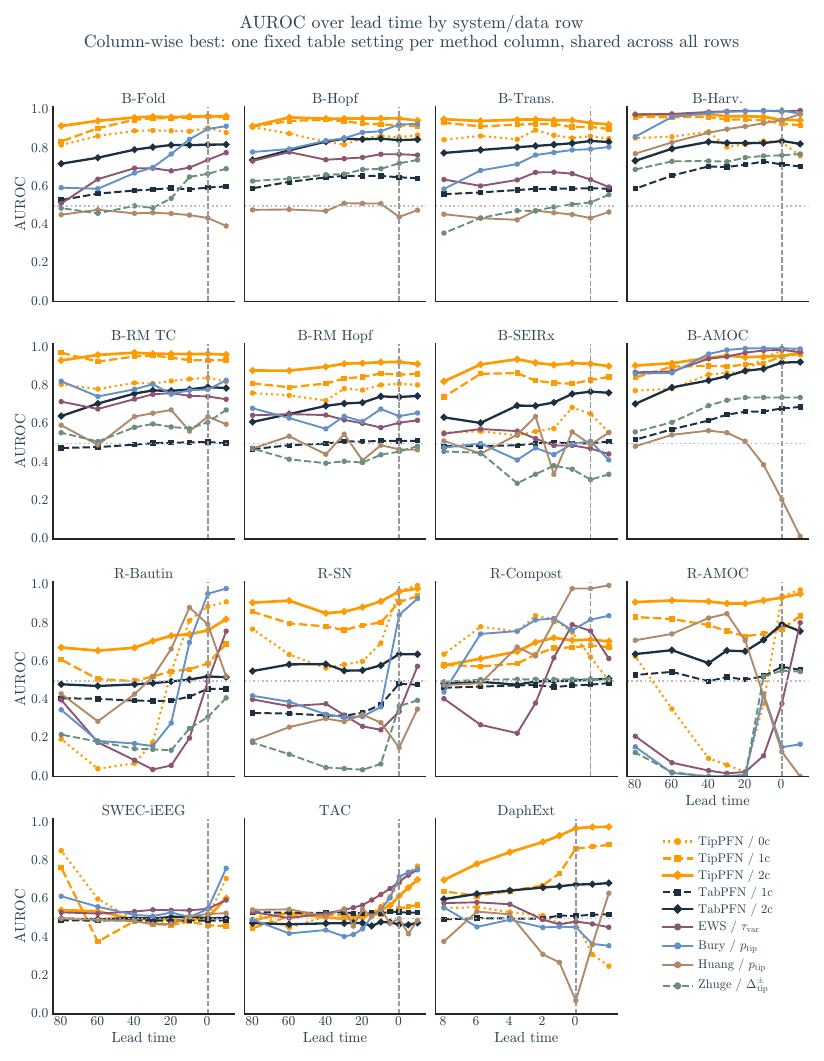}
    \caption{
        \textbf{AUROC over lead time for all systems}, selecting each model's score variant that performed best across all these systems (same as in Table~\ref{tab:auroc-results-best-col} for $\Delta > 0$.
    }
    \label{fig:auroc-all-colwise-best}
\end{figure}

\begin{figure}[p]
    \centering
    \includegraphics[width=1\linewidth]{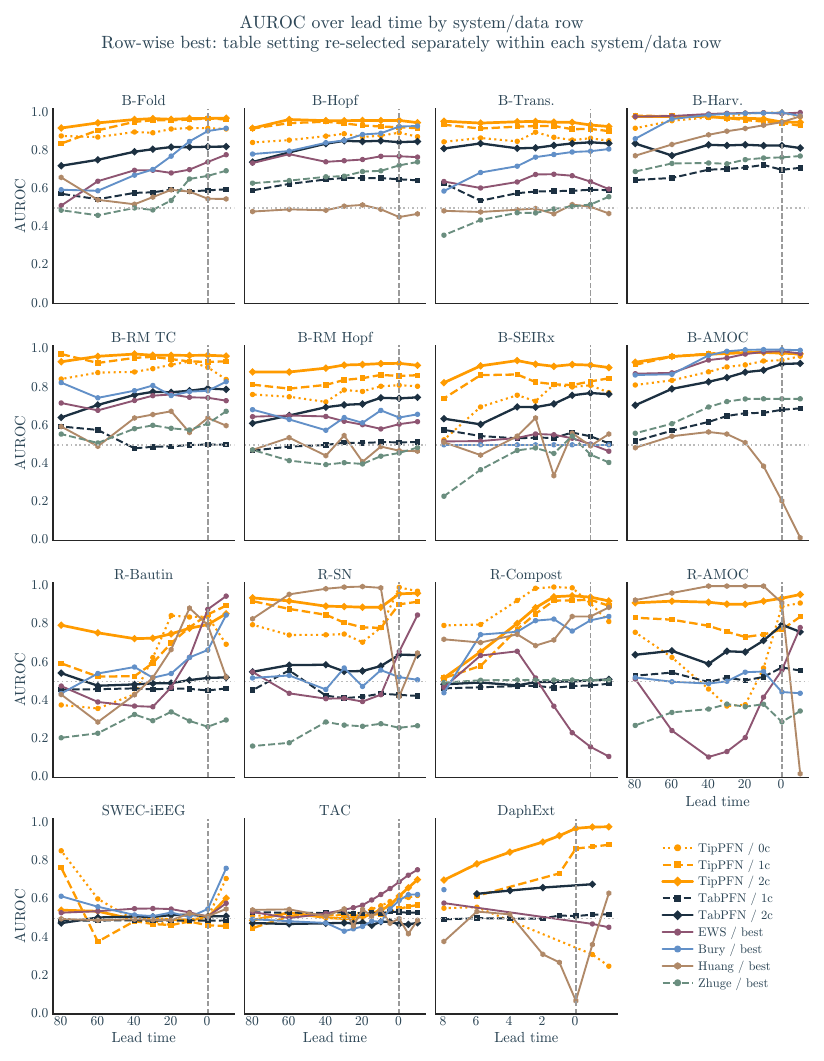}
    \caption{\textbf{AUROC over lead time for all systems}, selecting each model's score variant that performed best on that particular system (corresponding to Table~\ref{tab:auroc-results-best-per-system}) for $\Delta > 0$.}
    \label{fig:auroc-all-rowwise-best}
\end{figure}

\FloatBarrier
\subsubsection{Held-out configuration selection}\label{si-sec:heldout-selection}
Each column of Table~\ref{tab:auroc-results-best-col} uses the configuration (query-window length and, for the ML baselines, the length adaptation) with the highest mean AUROC over all rows, and Table~\ref{tab:auroc-results-best-per-system} selects a configuration, including the TipPFN and TabPFN score head, separately for each row.
Because the configuration is selected on the data on which it is reported, the reported values can be optimistic.
To estimate this effect, we repeat the selection on held-out episodes.
For each of $R=200$ repeats, we split the evaluation episodes of each system or dataset, stratified by class, into a 20\% selection fold and an 80\% report fold.
An episode is one source trajectory and is assigned to one fold as a whole, so all its lead times and query windows stay on the same side. Episodes of one synthetic sample share the system draw and serve as each other's context, so the folds are not independent at the level of system draws.
The SWEC-iEEG row pools all patients, so its episodes are split over all patients rather than per patient.
On the selection fold, each column (column-wise) or each cell (row-wise) selects its configuration with the rule of the published table; the selected configuration is then evaluated on the report fold with the balanced AUROC averaged over $\Delta>0$.
All configurations are scored from the existing per-query scores; no model is re-run.
Table~\ref{tab:si-heldout-selection} reports means over the 12 non-canonical rows (9 for Zhuge, which has no real-world rows); the canonical systems share their bifurcation types with the training prior and are left out.

With column-wise selection, the held-out means differ from the post hoc values by at most .02, and TipPFN with two context episodes is unchanged (.809).
Including the canonical systems, its held-out mean is .838 [.832, .844], equal to Table~\ref{tab:auroc-results-best-col}.
Row-wise selection is more optimistic, by up to .045 for TabPFN and .017 for TipPFN with two context episodes, because each cell chooses among several configurations on few episodes.
In both variants, TipPFN with two context episodes remains the best method, .18 (column-wise) and .19 (row-wise) above the best other method (TabPFN with two context episodes and Bury, respectively).

The selected configuration is stable for TipPFN with one and two context episodes (128-step window in 191 and 196 of 200 repeats).
Without context, the selection folds pick the 96- or 64-step window in 195 of 200 repeats, whereas all data give the 128-step window.
The held-out mean over the non-canonical rows is then higher (.641 against .610).

\begin{table}[t]
\centering
\caption{\textbf{Row-wise best pre-tip AUROC by system and dataset.} Balanced AUROC for critical-vs-all tipping detection, averaged over positive lead times ($\Delta > 0$). Each cell re-selects the best score/configuration within that system/data row and method column. The small line below each AUROC is the picked configuration: \texttt{w<N>} for query-window length, \texttt{res} = \textit{resample}, \texttt{bf} = \textit{backfill}, \texttt{pad} = \textit{pad}; for TipPFN/TabPFN it also lists the selected score head (e.g.\ \texttt{fc:1-median} or \texttt{fc:P(rdtc<0.05)}).
Feature budget is fixed at 16 within the TipPFN/TabPFN columns; baseline score heads are those of Table~\ref{tab:auroc-column-choices}, and all score heads are defined in Table~\ref{tab:si-score-definitions}.
\\
For the \texttt{TAC} dataset, higher performance can be achieved by selective context composition, see Fig.~\ref{fig:auroc-tac-cnc}.
}
\label{tab:auroc-results-best-per-system}
\resizebox{\columnwidth}{!}{%
\begin{tabular}{lccccccccc}
\toprule
System/data & \shortstack[c]{\strut TipPFN\\\strut 0c} & \shortstack[c]{\strut TipPFN\\\strut 1c} & \shortstack[c]{\strut TipPFN\\\strut 2c} & \shortstack[c]{\strut TabPFN\\\strut 1c} & \shortstack[c]{\strut TabPFN\\\strut 2c} & \shortstack[c]{\strut EWS\\\strut best} & \shortstack[c]{\strut Bury~\cite{bury2021deep}\\\strut best} & \shortstack[c]{\strut Huang~\cite{huang2024deep}\\\strut best} & \shortstack[c]{\strut Zhuge~\cite{zhuge2025deep}\\\strut best} \\
\midrule
\multicolumn{10}{l}{\emph{Canonical}} \\
B-Fold & \shortstack[c]{.935\\{\scriptsize w96~/~fc:P(rdtc<0.2)}} & \shortstack[c]{.929\\{\scriptsize w128~/~fc:1-median}} & \shortstack[c]{\textbf{.955}\\{\scriptsize w128~/~fc:1-median}} & \shortstack[c]{.584\\{\scriptsize w64~/~fc:P(rdtc<0.3)}} & \shortstack[c]{.786\\{\scriptsize w64~/~fc:1-median}} & \shortstack[c]{.655\\{\scriptsize w128}} & \shortstack[c]{.697\\{\scriptsize res~/~w128}} & \shortstack[c]{.577\\{\scriptsize pad~/~w128}} & \shortstack[c]{.522\\{\scriptsize res~/~w128}} \\
B-Hopf & \shortstack[c]{.893\\{\scriptsize w128~/~fc:P(rdtc<0.1)}} & \shortstack[c]{.935\\{\scriptsize w128~/~fc:1-median}} & \shortstack[c]{\textbf{.952}\\{\scriptsize w128~/~fc:P(rdtc<0.05)}} & \shortstack[c]{.640\\{\scriptsize w64~/~fc:1-median}} & \shortstack[c]{.820\\{\scriptsize w64~/~fc:1-median}} & \shortstack[c]{.755\\{\scriptsize w128}} & \shortstack[c]{.842\\{\scriptsize res~/~w128}} & \shortstack[c]{.497\\{\scriptsize res~/~w96}} & \shortstack[c]{.665\\{\scriptsize res~/~w128}} \\
B-Trans. & \shortstack[c]{.874\\{\scriptsize w128~/~fc:P(rdtc<0.3)}} & \shortstack[c]{.925\\{\scriptsize w128~/~fc:1-median}} & \shortstack[c]{\textbf{.950}\\{\scriptsize w128~/~fc:1-median}} & \shortstack[c]{.601\\{\scriptsize w128~/~fc:P(rdtc<0.3)}} & \shortstack[c]{.824\\{\scriptsize w128~/~fc:1-median}} & \shortstack[c]{.650\\{\scriptsize w128}} & \shortstack[c]{.722\\{\scriptsize res~/~w128}} & \shortstack[c]{.490\\{\scriptsize bf~/~w64}} & \shortstack[c]{.458\\{\scriptsize res~/~w128}} \\
\midrule
\multicolumn{10}{l}{\emph{Semi-real}} \\
B-Harv. & \shortstack[c]{.970\\{\scriptsize w128~/~fc:1-median}} & \shortstack[c]{.971\\{\scriptsize w128~/~fc:1-median}} & \shortstack[c]{.980\\{\scriptsize w128~/~fc:1-mean}} & \shortstack[c]{.691\\{\scriptsize w96~/~fc:1-median}} & \shortstack[c]{.847\\{\scriptsize w96~/~fc:P(rdtc<0.3)}} & \shortstack[c]{\textbf{.991}\\{\scriptsize w128}} & \shortstack[c]{.967\\{\scriptsize res~/~w128}} & \shortstack[c]{.873\\{\scriptsize res~/~w64}} & \shortstack[c]{.734\\{\scriptsize res~/~w128}} \\
B-RM TC & \shortstack[c]{.901\\{\scriptsize w96~/~fc:P(rdtc<0.2)}} & \shortstack[c]{.952\\{\scriptsize w128~/~fc:P(rdtc<0.05)}} & \shortstack[c]{\textbf{.963}\\{\scriptsize w128~/~fc:P(rdtc<0.1)}} & \shortstack[c]{.569\\{\scriptsize w128~/~fc:P(rdtc<0.3)}} & \shortstack[c]{.741\\{\scriptsize w64~/~fc:1-median}} & \shortstack[c]{.733\\{\scriptsize w128}} & \shortstack[c]{.784\\{\scriptsize res~/~w128}} & \shortstack[c]{.604\\{\scriptsize res~/~w64}} & \shortstack[c]{.570\\{\scriptsize res~/~w128}} \\
B-RM Hopf & \shortstack[c]{.770\\{\scriptsize w128~/~fc:P(rdtc<0.3)}} & \shortstack[c]{.845\\{\scriptsize w128~/~fc:P(rdtc<0.1)}} & \shortstack[c]{\textbf{.912}\\{\scriptsize w128~/~fc:P(rdtc<0.05)}} & \shortstack[c]{.529\\{\scriptsize w128~/~fc:P(rdtc<0.3)}} & \shortstack[c]{.689\\{\scriptsize w64~/~fc:1-median}} & \shortstack[c]{.627\\{\scriptsize w128}} & \shortstack[c]{.638\\{\scriptsize res~/~w128}} & \shortstack[c]{.484\\{\scriptsize res~/~w64}} & \shortstack[c]{.422\\{\scriptsize res~/~w128}} \\
B-SEIRx & \shortstack[c]{.723\\{\scriptsize w128~/~fc:P(rdtc<0.3)}} & \shortstack[c]{.831\\{\scriptsize w128~/~fc:1-mean}} & \shortstack[c]{\textbf{.907}\\{\scriptsize w128~/~fc:1-mean}} & \shortstack[c]{.583\\{\scriptsize w64~/~fc:P(rdtc<0.05)}} & \shortstack[c]{.699\\{\scriptsize w64~/~fc:P(rdtc<0.05)}} & \shortstack[c]{.536\\{\scriptsize w96}} & \shortstack[c]{.500\\{\scriptsize pad~/~w128}} & \shortstack[c]{.507\\{\scriptsize res~/~w64}} & \shortstack[c]{.426\\{\scriptsize bf~/~w96}} \\
B-AMOC & \shortstack[c]{.906\\{\scriptsize w96~/~fc:P(rdtc<0.3)}} & \shortstack[c]{.969\\{\scriptsize w128~/~fc:1-median}} & \shortstack[c]{\textbf{.972}\\{\scriptsize w128~/~fc:P(rdtc<0.05)}} & \shortstack[c]{.632\\{\scriptsize w64~/~fc:P(rdtc<0.3)}} & \shortstack[c]{.829\\{\scriptsize w64~/~fc:P(rdtc<0.3)}} & \shortstack[c]{.935\\{\scriptsize w128}} & \shortstack[c]{.948\\{\scriptsize res~/~w128}} & \shortstack[c]{.509\\{\scriptsize res~/~w64}} & \shortstack[c]{.679\\{\scriptsize res~/~w128}} \\
R-Bautin & \shortstack[c]{.580\\{\scriptsize w64~/~fc:1-median}} & \shortstack[c]{.643\\{\scriptsize w64~/~fc:P(rdtc<0.1)}} & \shortstack[c]{\textbf{.776}\\{\scriptsize w64~/~fc:P(rdtc<0.05)}} & \shortstack[c]{.468\\{\scriptsize w64~/~fc:P(rdtc<0.05)}} & \shortstack[c]{.501\\{\scriptsize w96~/~fc:P(rdtc<0.3)}} & \shortstack[c]{.451\\{\scriptsize w64}} & \shortstack[c]{.541\\{\scriptsize res~/~w64}} & \shortstack[c]{.538\\{\scriptsize res~/~w64}} & \shortstack[c]{.283\\{\scriptsize bf~/~w96}} \\
R-SN & \shortstack[c]{.754\\{\scriptsize w64~/~fc:1-median}} & \shortstack[c]{.864\\{\scriptsize w96~/~fc:P(rdtc<0.3)}} & \shortstack[c]{.931\\{\scriptsize w128~/~fc:P(rdtc<0.3)}} & \shortstack[c]{.501\\{\scriptsize w64~/~fc:P(rdtc<0.3)}} & \shortstack[c]{.592\\{\scriptsize w64~/~fc:P(rdtc<0.3)}} & \shortstack[c]{.440\\{\scriptsize w64}} & \shortstack[c]{.520\\{\scriptsize bf~/~w96}} & \shortstack[c]{\textbf{.959}\\{\scriptsize pad~/~w128}} & \shortstack[c]{.242\\{\scriptsize bf~/~w96}} \\
R-Compost & \shortstack[c]{\textbf{.927}\\{\scriptsize w128~/~fc:P(rdtc<0.05)}} & \shortstack[c]{.763\\{\scriptsize w128~/~fc:1-median}} & \shortstack[c]{.793\\{\scriptsize w128~/~fc:1-median}} & \shortstack[c]{.500\\{\scriptsize w64~/~fc:P(rdtc<0.3)}} & \shortstack[c]{.505\\{\scriptsize w64~/~fc:P(rdtc<0.3)}} & \shortstack[c]{.481\\{\scriptsize w64}} & \shortstack[c]{.727\\{\scriptsize res~/~w128}} & \shortstack[c]{.737\\{\scriptsize res~/~w128}} & \shortstack[c]{.506\\{\scriptsize res~/~w128}} \\
R-AMOC & \shortstack[c]{.584\\{\scriptsize w64~/~fc:P(rdtc<0.3)}} & \shortstack[c]{.798\\{\scriptsize w128~/~fc:P(rdtc<0.3)}} & \shortstack[c]{.924\\{\scriptsize w128~/~fc:P(rdtc<0.05)}} & \shortstack[c]{.616\\{\scriptsize w64~/~fc:P(rdtc<0.3)}} & \shortstack[c]{.699\\{\scriptsize w64~/~fc:P(rdtc<0.3)}} & \shortstack[c]{.270\\{\scriptsize w64}} & \shortstack[c]{.519\\{\scriptsize bf~/~w96}} & \shortstack[c]{\textbf{.982}\\{\scriptsize bf~/~w128}} & \shortstack[c]{.350\\{\scriptsize bf~/~w96}} \\
\midrule
\multicolumn{10}{l}{\emph{Real-world \& Sim-to-Real}} \\
SWEC-iEEG & \shortstack[c]{\textbf{.580}\\{\scriptsize w128~/~fc:P(rdtc<0.2)}} & \shortstack[c]{.512\\{\scriptsize w128~/~fc:1-mean}} & \shortstack[c]{.563\\{\scriptsize w128~/~fc:P(rdtc<0.3)}} & \shortstack[c]{.489\\{\scriptsize w64~/~fc:1-median}} & \shortstack[c]{.516\\{\scriptsize w128~/~fc:P(rdtc<0.3)}} & \shortstack[c]{.542\\{\scriptsize w96}} & \shortstack[c]{.541\\{\scriptsize res~/~w128}} & \shortstack[c]{.499\\{\scriptsize res~/~w128}} & -- \\
TAC & \shortstack[c]{.536\\{\scriptsize w96~/~fc:P(rdtc<0.05)}} & \shortstack[c]{.526\\{\scriptsize w128~/~fc:P(rdtc<0.3)}} & \shortstack[c]{.554\\{\scriptsize w128~/~fc:P(rdtc<0.3)}} & \shortstack[c]{.533\\{\scriptsize w64~/~fc:P(rdtc<0.1)}} & \shortstack[c]{.476\\{\scriptsize w128~/~fc:P(rdtc<0.1)}} & \shortstack[c]{\textbf{.568}\\{\scriptsize w128}} & \shortstack[c]{.479\\{\scriptsize res~/~w64}} & \shortstack[c]{.522\\{\scriptsize res~/~w64}} & -- \\
DaphniaExt & \shortstack[c]{.556\\{\scriptsize w16~/~fc:1-median}} & \shortstack[c]{.736\\{\scriptsize w16~/~fc:P(rdtc<0.05)}} & \shortstack[c]{\textbf{.833}\\{\scriptsize w16~/~fc:1-median}} & \shortstack[c]{.514\\{\scriptsize w16~/~fc:P(rdtc<0.1)}} & \shortstack[c]{.672\\{\scriptsize w8~/~fc:1-mean}} & \shortstack[c]{.580\\{\scriptsize w12}} & \shortstack[c]{.651\\{\scriptsize pad~/~w12}} & \shortstack[c]{.404\\{\scriptsize res~/~w8}} & -- \\
\midrule
\emph{All datasets} & .766 & .813 & \textbf{.864} & .563 & .680 & .614 & .672 & .612 & .488 \\
\bottomrule
\end{tabular}
}
\end{table}

\begin{table}[h]
\centering
\small
\caption{\textbf{Held-out configuration selection.} Mean balanced AUROC ($\Delta>0$) over the 12 non-canonical rows (Zhuge: 9 rows).
\emph{Post hoc}: configuration selected on all episodes, per column as in Table~\ref{tab:auroc-results-best-col} or per cell as in Table~\ref{tab:auroc-results-best-per-system}.
\emph{Held-out}: configuration selected on a 20\% episode fold and evaluated on the other 80\%; mean over $R=200$ splits, 5th--95th percentile in brackets.}
\label{tab:si-heldout-selection}
\begin{tabular}{lcccc}
\toprule
 & \multicolumn{2}{c}{Column-wise} & \multicolumn{2}{c}{Row-wise} \\
\cmidrule(lr){2-3}\cmidrule(lr){4-5}
Method & Post hoc & Held-out [p5, p95] & Post hoc & Held-out [p5, p95] \\
\midrule
TipPFN 0c & .621 & .641 [.631, .649] & .732 & .707 [.684, .730] \\
TipPFN 1c & .741 & .740 [.732, .749] & .784 & .764 [.750, .777] \\
TipPFN 2c & \textbf{.810} & \textbf{.810} [.803, .816] & \textbf{.842} & \textbf{.825} [.811, .837] \\
TabPFN 1c & .503 & .499 [.482, .515] & .552 & .507 [.487, .528] \\
TabPFN 2c & .631 & .627 [.609, .644] & .647 & .605 [.583, .628] \\
EWS $\tau_{\mathrm{var}}$ & .564 & .552 [.537, .569] & .596 & .588 [.576, .598] \\
Bury~\cite{bury2021deep} & .566 & .568 [.562, .576] & .651 & .635 [.622, .647] \\
Huang~\cite{huang2024deep} & .549 & .549 [.541, .559] & .635 & .627 [.612, .649] \\
Zhuge~\cite{zhuge2025deep} & .407 & .406 [.393, .420] & .468 & .451 [.431, .469] \\
\bottomrule
\end{tabular}
\end{table}

\begin{table}[t]
\centering
\caption{
    Source-trajectory support for the AUROC evaluation. For each system or empirical benchmark row, we report the number of unique source samples and query episodes represented in the result tables, split by real versus synthetic source origin and by critical versus non-critical labels. Counts are deduplicated over lead-time offsets and model-configuration sweeps.
    Canonical and semi-real systems are synthetic simulation benchmarks by construction.
}
\label{tab:si-system-sample-episode-counts}
\begingroup
\renewcommand{\arraystretch}{1.2}
\resizebox{\columnwidth}{!}{%
\begin{tabular}{lrrrrrrl}
\toprule
System/data & samples & episodes & \shortstack[c]{real\\$n_{\mathrm{pos}}$} & \shortstack[c]{real\\$n_{\mathrm{neg}}$} & \shortstack[c]{synth.\\$n_{\mathrm{pos}}$} & \shortstack[c]{synth.\\$n_{\mathrm{neg}}$} & \shortstack[c]{$n_{\mathrm{neg}}$\\split} \\
\midrule
\multicolumn{8}{l}{\emph{Canonical}}\\
B-Fold & 41 & 164 & 0 & 0 & 80 & 84 & \shortstack[l]{19 equilibrium, 15 flat\\31 receding, 19 approaching} \\[1pt]
B-Hopf & 46 & 184 & 0 & 0 & 93 & 91 & \shortstack[l]{18 equilibrium, 21 flat\\29 receding, 23 approaching} \\[1pt]
B-Trans. & 41 & 164 & 0 & 0 & 82 & 82 & \shortstack[l]{28 equilibrium, 17 flat\\17 receding, 20 approaching} \\[1pt]
\midrule
\multicolumn{8}{l}{\emph{Semi-real}}\\
B-Harv. & 20 & 80 & 0 & 0 & 40 & 40 & \shortstack[l]{9 equilibrium, 13 flat\\7 receding, 11 approaching} \\[1pt]
B-RM TC & 31 & 124 & 0 & 0 & 62 & 62 & \shortstack[l]{15 equilibrium, 18 flat\\16 receding, 13 approaching} \\[1pt]
B-RM Hopf & 33 & 132 & 0 & 0 & 65 & 67 & \shortstack[l]{17 equilibrium, 16 flat\\15 receding, 19 approaching} \\[1pt]
B-SEIRx & 17 & 68 & 0 & 0 & 35 & 33 & \shortstack[l]{11 equilibrium, 7 flat\\5 receding, 10 approaching} \\[1pt]
B-AMOC & 27 & 108 & 0 & 0 & 54 & 54 & \shortstack[l]{14 equilibrium, 12 flat\\13 receding, 15 approaching} \\[1pt]
R-Bautin & 34 & 136 & 0 & 0 & 66 & 70 & \shortstack[l]{38 equilibrium, 32 flat} \\[1pt]
R-SN & 35 & 140 & 0 & 0 & 71 & 69 & \shortstack[l]{41 equilibrium, 28 flat} \\[1pt]
R-Compost & 31 & 124 & 0 & 0 & 63 & 61 & \shortstack[l]{30 equilibrium, 31 flat} \\[1pt]
R-AMOC & 28 & 112 & 0 & 0 & 56 & 56 & \shortstack[l]{25 equilibrium, 31 flat} \\[1pt]
\midrule
\multicolumn{8}{l}{\emph{Others}}\\
SWEC-iEEG & 18 & 807 & 116 & 691 & 0 & 0 & \shortstack[l]{691 baseline} \\[1pt]
TAC & 2 & 320 & 80 & 80 & 80 & 80 & \shortstack[l]{80 real baseline, 80 synth. baseline} \\[1pt]
DaphniaExt & 2 & 110 & 30 & 25 & 30 & 25 & \shortstack[l]{25 real baseline, 25 synth. baseline} \\[1pt]
\midrule
\multicolumn{8}{l}{\emph{Summary}}\\
Total & 406 & 2{,}773 & 226 & 796 & 877 & 874 & \shortstack[l]{796 real baseline, 105 synth. baseline, 265 synth. equilibrium\\241 synth. flat, 133 synth. receding, 130 synth. approaching} \\[1pt]
\bottomrule
\end{tabular}%
}
\endgroup
\end{table}

\subsubsection{Native-window baselines}\label{si-sec:native-window}
Table~\ref{tab:auroc-results-best-col} restricts every method to query windows of at most 128 steps, although the ML baselines were designed for longer inputs.
For a comparison at each method's native window, we re-simulated the canonical and semi-real b-tipping validation systems with 1024 steps per episode.
The baselines run at full resolution with their native windows (Bury and Zhuge $W=500$, Huang $W=680$, TabPFN $W=768$ with two context episodes; EWS uses the variance trend over the available history), while TipPFN observes every fourth step with $W=128$, a comparable span of 512 steps.
Lead times are matched in time: $\Delta$ TipPFN steps correspond to $4\Delta$ full-resolution steps.
TipPFN with two context episodes leads on 6 of 8 systems (Table~\ref{tab:si-native-window}); Bury and EWS score higher on B-Harvesting, and Bury on B-AMOC.
The TipPFN runs use all 16 input variables, and the re-simulated sets keep the auxiliary-variable wiring of Table~\ref{tab:auroc-results-best-col}, so this lead is subject to the same caveat (Appendix~\ref{si-sec:aux-drive-sensitivity}).
R-tipping systems are not included: with the construction of their non-tipping episodes used in these sets (Appendix~\ref{si-sec:datasets}), the two classes may already differ before the forcing ramp; Appendix~\ref{si-sec:aux-drive-sensitivity} reports r-tipping results for the second construction.

\paragraph{Real-world data}
On SWEC-iEEG and TAC, all methods read the same time grid (1\,s and 0.02\,s), so lead times need no conversion, and all methods are scored on the same critical episodes and lead times.
For SWEC-iEEG, we use a version of the evaluation set with an 872\,s look-back before each of the 116 seizures and 529 interictal segments, so every native window fits at all lead times $\Delta\in\{10,\dots,80\}$\,s (three seeds; 345--348 critical and 1{,}581 non-critical windows per lead time); its TipPFN values therefore differ from Table~\ref{tab:auroc-results-best-col}.
The univariate baselines are averaged over three band-power channels.
The TAC episodes of Table~\ref{tab:auroc-results-best-col} hold only about 200 steps before the transition.
We re-extracted the critical cycles from the raw recordings with a longer segment; the usable history before a cycle ends at the oscillating state of the previous cycle (median 670 steps).
Of the 80 cycles of Table~\ref{tab:auroc-results-best-col}, eight at the start of a recording and one with a contaminated history are dropped, and the remaining 71 have at least 500 clean steps, so Bury, Zhuge and EWS run at $W=500$ for all lead times $\Delta\in\{5,\dots,80\}$ steps (five seeds; 355 critical and 400 stationary windows per lead time).
Huang's $W=680$ fits in 9--33 cycles for $\Delta\leq25$ and in none for $\Delta\geq40$; we average it over $\Delta\leq25$ and report TipPFN on the same items.
No cycle has 768 clean steps, so TabPFN runs at $W=500$ with two context episodes; its context pool also contains 18 cycles that are not in the pool of Table~\ref{tab:auroc-results-best-col}.
TipPFN is scored on the same 71 cycles.
Zhuge takes the control-parameter trajectory as a second input, which the real-world data do not provide.
The RDTC and the TAC fuel flow cannot serve as this input, because they ramp only in critical episodes and would reveal the label; we feed elapsed time, which is identical for both classes.
DaphniaExt is not included, as its windows have at most 16 steps.

No method separates the classes at native windows either (Table~\ref{tab:si-native-window}, bottom).
On SWEC-iEEG, all means lie between .455 and .519.
On TAC, they lie between .474 and .542, except for Zhuge with the time input (.369, below .5 at every lead time; we do not flip its sign, as that would be fitted on the test data) and Huang on its $\Delta\leq25$ subset (.574; TipPFN without context reaches .578 on the same items).

\begin{table}[h]
\centering
\caption{\textbf{Native-window comparison} on re-simulated canonical and semi-real b-tipping systems (1024 steps per episode) and on two real-world datasets.
Balanced AUROC averaged over matched lead times.
Synthetic rows: $\Delta\in\{10,\dots,80\}$ TipPFN steps, i.e.\ $4\Delta$ full-resolution steps; baselines use their native window $W$ at full resolution, TipPFN uses $W=128$ on every fourth step.
Huang and TabPFN have no feasible critical windows at the longest lead times and are averaged over $\Delta\leq40$ and $\Delta\leq30$ (TipPFN 2c mean on these subsets: .959); TabPFN cells rest on 1--12 critical episodes at their largest lead time.
Real-world rows: all methods use the recorded time grid, the baselines with the window in the column header and TipPFN with $W=128$; $\Delta\in\{10,\dots,80\}$\,s (SWEC-iEEG) and $\Delta\in\{5,\dots,80\}$ steps (TAC); Zhuge receives elapsed time as its control input.
$^{\dagger}$Huang's output saturates on B-AMOC (all scores below $10^{-14}$), so this value is degenerate.
$^{\ddagger}$Only $\Delta\leq25$ (9--33 cycles per lead time); TipPFN 0c/1c/2c on the same items: .578/.521/.559.
$^{\S}$Run at $W=500$, since no TAC cycle has 768 clean steps; the context pool also contains 18 cycles outside the pool of Table~\ref{tab:auroc-results-best-col}.
Best value per synthetic row in bold.}
\label{tab:si-native-window}
\resizebox{\columnwidth}{!}{%
\begin{tabular}{lcccccccc}
\toprule
System & \shortstack[c]{\strut Bury\\\strut $W{=}500$} & \shortstack[c]{\strut EWS $\tau_{\mathrm{var}}$\\\strut history} & \shortstack[c]{\strut Huang\\\strut $W{=}680$} & \shortstack[c]{\strut Zhuge\\\strut $W{=}500$} & \shortstack[c]{\strut TabPFN 2c\\\strut $W{=}768$} & \shortstack[c]{\strut TipPFN\\\strut 0c} & \shortstack[c]{\strut TipPFN\\\strut 1c} & \shortstack[c]{\strut TipPFN\\\strut 2c} \\
\midrule
\multicolumn{9}{l}{\emph{Canonical}} \\
B-Fold & .805 & .748 & .508 & .515 & .830 & .843 & .931 & \textbf{.949} \\
B-Hopf & .864 & .807 & .501 & .578 & .820 & .864 & .939 & \textbf{.973} \\
B-Trans. & .740 & .664 & .522 & .513 & .857 & .845 & .928 & \textbf{.946} \\
\midrule
\multicolumn{9}{l}{\emph{Semi-real}} \\
B-Harv. & \textbf{.994} & .982 & .471 & .656 & .798 & .869 & .940 & .968 \\
B-RM TC & .846 & .879 & .505 & .409 & .900 & .787 & .952 & \textbf{.981} \\
B-RM Hopf & .550 & .732 & .482 & .216 & .861 & .752 & .849 & \textbf{.895} \\
B-SEIRx & .511 & .559 & .503 & .394 & .801 & .596 & .831 & \textbf{.939} \\
B-AMOC & \textbf{.989} & .935 & .004$^{\dagger}$ & .684 & .791 & .950 & .963 & .965 \\
\midrule
\emph{Mean (8 systems)} & .787 & .788 & .437 & .496 & .832 & .813 & .917 & \textbf{.952} \\
\midrule
\multicolumn{9}{l}{\emph{Real-world}} \\
SWEC-iEEG & .459 & .509 & .473 & .499 & .513 & .455 & .501 & .519 \\
TAC & .542 & .532 & .574$^{\ddagger}$ & .369 & .474$^{\S}$ & .513 & .510 & .513 \\
\bottomrule
\end{tabular}
}
\end{table}

\FloatBarrier
\subsubsection{Ablations and sensitivity analysis}\label{si-sec:ablations}
This section details the ablations summarized in Sec.~\ref{sec:ablations}.
All of them use the evaluation sets of Table~\ref{tab:auroc-results-best-col}, and both arms of every comparison are evaluated on the same data; SWEC-iEEG is not included.
Unless stated otherwise, the score is the forecast $1-\mathrm{median}(\Lambda)$ with $W=128$ and 16 observed features, averaged over $\Delta>0$ as in Table~\ref{tab:auroc-results-best-col}; family values are unweighted means over systems, and \emph{simulated} denotes the 3 canonical and 9 semi-real systems.
The two target ablations need evaluation options that the code behind Table~\ref{tab:auroc-results-best-col} does not have; they were run with later versions of our evaluation code, with the RDTC reference re-evaluated in the same version, so their RDTC family means differ from Table~\ref{tab:auroc-results-best-col} by up to .03.

\paragraph{Training target}
We compare RDTC with two other training targets, each used to train a model with the architecture of TipPFN on the same prior.
The \emph{single-label classifier} predicts one label per episode, whether it eventually tips, read at the final forecast step and trained with a cross-entropy loss.
Its training contexts end before the transition (non-tipping contexts at a pseudo-anchor drawn from the transition times), and we truncate the evaluation contexts in the same way; without truncation its context results are lower.
The \emph{dense binary target} keeps the loss, masking and context interface of TipPFN and replaces the graded RDTC at every time step by the episode outcome ($\pm0.95$ for tipping and non-tipping episodes), in the query and in the context episodes.
It is the closer control: only the information about the distance to the transition is removed, whereas the classifier also changes the loss, the output and the context format.

On identical query rows and context episodes (Table~\ref{tab:si-abl-classifier}), the classifier is close to RDTC without context ($-.031$ on the simulated systems; it is ahead on the canonical systems), but context lowers its AUROC below its own zero-context value, so RDTC leads by .213 and .245 with one and two context episodes.
The classifier checkpoint (28k steps, selected on validation data) was trained longer than the released model (22k steps); after 22k steps the classifier reaches .583 without context, against .724 for RDTC.
The dense binary target (Table~\ref{tab:si-abl-binary}) is within .02 of RDTC without context on the simulated systems, benefits from context, and stays below RDTC by .144--.167 with two context episodes; the graded target is better in every cell except r-tipping without context at $d=96$.
The paper-size binary model was trained for fewer steps (9k) than the released model (22k), which favors RDTC; the $d=96$ pair (32k binary and 37k RDTC steps) shows the same pattern.
On TAC, all four models of Table~\ref{tab:si-abl-binary} are near chance (.40--.56); TAC and DaphniaExt are not included in the table.
Dense supervision is therefore needed to learn from one or two context episodes, and the graded RDTC target adds to it.

\begin{table}[h]
\centering
\small
\caption{\textbf{RDTC versus a single-label classifier.} AUROC averaged over $\Delta>0$ for the released model (RDTC) and a classifier with the same architecture trained on one eventual-tipping label per episode, on identical query rows and context episodes. Classifier contexts are truncated before the transition, as in its training. The last row gives the paired difference with a 95\% cluster-bootstrap interval (2{,}000 replicates over source samples).}
\label{tab:si-abl-classifier}
\begin{tabular}{llccc}
\toprule
Systems & Target & 0c & 1c & 2c \\
\midrule
Canonical (3) & RDTC & .872 & .931 & .950 \\
 & Single label & .887 & .832 & .842 \\
\midrule
Semi-real (9) & RDTC & .675 & .824 & .899 \\
 & Single label & .628 & .573 & .608 \\
\midrule
Simulated (12) & RDTC & .724 & .851 & .911 \\
 & Single label & .693 & .638 & .666 \\
 & Difference & $+.031$ {\scriptsize [$+.019$, $+.042$]} & $+.213$ {\scriptsize [$+.197$, $+.229$]} & $+.245$ {\scriptsize [$+.229$, $+.262$]} \\
\bottomrule
\end{tabular}
\end{table}
\begin{table}[h]
\centering
\small
\caption{\textbf{Graded RDTC versus a dense binary target.} AUROC averaged over $\Delta>0$ for models trained on the graded RDTC (as TipPFN) or on the episode outcome at every time step (binary), at the size of the released model ($d=1280$) and at model dimension 96. The binary models see the outcome encoding in their context episodes, as in training. Training steps: $d=1280$ RDTC 22k (released model), binary 9k; $d=96$ RDTC 37k, binary 32k.}
\label{tab:si-abl-binary}
\begin{tabular}{llcccccc}
\toprule
& & \multicolumn{3}{c}{$d=1280$} & \multicolumn{3}{c}{$d=96$} \\
\cmidrule(lr){3-5}\cmidrule(lr){6-8}
Systems & Target & 0c & 1c & 2c & 0c & 1c & 2c \\
\midrule
Canonical (3) & RDTC & .876 & .927 & .951 & .867 & .899 & .923 \\
 & Binary & .838 & .856 & .859 & .666 & .749 & .826 \\
\midrule
Semi-real b-tipping (5) & RDTC & .790 & .898 & .939 & .696 & .866 & .910 \\
 & Binary & .783 & .785 & .840 & .673 & .774 & .799 \\
\midrule
Semi-real r-tipping (4) & RDTC & .544 & .701 & .812 & .488 & .657 & .795 \\
 & Binary & .527 & .444 & .501 & .616 & .536 & .575 \\
\midrule
Simulated (12) & RDTC & .730 & .839 & .899 & .670 & .805 & .875 \\
 & Binary & .712 & .689 & .732 & .652 & .688 & .731 \\
\bottomrule
\end{tabular}
\end{table}
\paragraph{Bury et al.\ retrained on the TipPFN prior}
To separate the effect of the training prior from that of the target, task construction and architecture, we retrained the CNN-LSTM classifier of~\cite{bury2021deep} on TipPFN's training data.
As in~\cite{bury2021deep}, it receives the univariate driver signal $x$ of each training episode, detrended with a LOWESS filter (span 0.2), as a sequence of length 256, and is trained for a four-way classification into fold, Hopf, transcritical and null, where null includes forced but subcritical episodes; the score is one minus the predicted null probability.
We kept the original architecture (one convolutional and two LSTM layers, about 23k parameters per network) and trained an ensemble of 20 networks for 300 epochs, ten for each of the two padding variants of~\cite{bury2021deep}, with 15\% of the samples held out each for validation and testing.
Evaluation follows Table~\ref{tab:auroc-results-best-col}; with the original weights, the same pipeline reproduces the Bury column of Table~\ref{tab:auroc-results-best-col} within .012 on 13 of 14 rows (R-Compost: $+.060$).

Compared with the original networks, the retrained ensemble is better on B-SEIRx, the r-tipping systems except R-AMOC, TAC and DaphniaExt, and about .1 worse on the canonical systems, B-Harv.\ and B-AMOC (Table~\ref{tab:si-abl-bury}).
TipPFN with two context episodes is above it on 12 of the 14 rows (all except R-Compost and TAC), and zero-context TipPFN on 9.
On the simulated systems, TipPFN's auxiliary variables also carry information about the forcing (Appendix~\ref{si-sec:aux-drive-sensitivity}); observing only $x$, as the retrained model does, zero-context TipPFN is above it on 7 of the 12 simulated systems, mainly canonical and b-tipping ones.
The training prior alone therefore does not explain TipPFN's performance.

\begin{table}[h]
\centering
\small
\caption{\textbf{Bury et al.\ retrained on the TipPFN prior.} AUROC averaged over $\Delta>0$. \emph{Change} is the difference to the original pretrained model (Bury column of Table~\ref{tab:auroc-results-best-col}). TipPFN columns give the number of context episodes (c) and observed auxiliary features (f); with 0f TipPFN observes only $x$, as the retrained model does. For TAC and DaphniaExt, the TipPFN columns are those of Table~\ref{tab:auroc-results-best-col}.}
\label{tab:si-abl-bury}
\begin{tabular}{lcccccc}
\toprule
& \multicolumn{2}{c}{Bury~\cite{bury2021deep}, retrained} & \multicolumn{3}{c}{TipPFN} \\
\cmidrule(lr){2-3}\cmidrule(lr){4-6}
System/data & AUROC & Change & 0c/0f & 0c/16f & 2c/16f \\
\midrule
\multicolumn{6}{l}{\emph{Canonical}} \\
B-Fold & .587 & $-.109$ & .818 & .876 & .955 \\
B-Hopf & .731 & $-.111$ & .816 & .861 & .952 \\
B-Trans. & .620 & $-.102$ & .747 & .864 & .950 \\
\midrule
\multicolumn{6}{l}{\emph{Semi-real}} \\
B-Harv. & .863 & $-.104$ & .970 & .846 & .974 \\
B-RM TC & .714 & $-.069$ & .791 & .815 & .963 \\
B-RM Hopf & .585 & $-.054$ & .582 & .769 & .904 \\
B-SEIRx & .520 & $+.052$ & .539 & .582 & .905 \\
B-AMOC & .858 & $-.090$ & .783 & .853 & .940 \\
R-Bautin & .508 & $+.200$ & .086 & .308 & .700 \\
R-SN & .381 & $+.026$ & .348 & .644 & .892 \\
R-Compost & .877 & $+.150$ & .686 & .767 & .665 \\
R-AMOC & .094 & $-.009$ & .171 & .257 & .913 \\
\midrule
\multicolumn{6}{l}{\emph{Real-world \& Sim-to-Real}} \\
TAC & .534 & $+.057$ & -- & .521 & .519 \\
DaphniaExt & .514 & $+.030$ & -- & .527 & .833 \\
\bottomrule
\end{tabular}
\end{table}
\paragraph{Model size}
Table~\ref{tab:si-abl-scaling} compares the released model ($d=1280$) with variants of model dimension 256, 96, 32 and 16, trained on the same prior; the smaller variants were trained on a subset of 200{,}000 training draws.
Averaged over zero to two context episodes on the 12 simulated systems, the $d=96$ and $d=32$ variants are .034 below the released model and $d=16$ is .078 below; with two context episodes, $d=32$ reaches .857 against .893 (Sec.~\ref{sec:ablations}).
All $d=256$ runs with learning rates of $6.25\cdot10^{-4}$ and above diverged during training; we report the checkpoint with the lowest validation loss before divergence (17k steps), which is below $d=96$ and $d=32$ and reflects this instability rather than a capacity limit.
On DaphniaExt, the small models are better than the released model with two context episodes and worse with one; with a single dataset, this is not evidence of a trend.
Scale is therefore not required for most of the performance.

\begin{table}[h]
\centering
\small
\caption{\textbf{Model size.} AUROC averaged over $\Delta>0$ for the released model ($d=1280$) and variants of smaller model dimension $d$, for 0, 1 and 2 context episodes (c). The $d=256$ checkpoint precedes a training divergence (see text). \emph{Mean} averages the simulated rows over 0--2 context episodes.}
\label{tab:si-abl-scaling}
\begin{tabular}{llccccc}
\toprule
& & $d=1280$ & $d=256$ & $d=96$ & $d=32$ & $d=16$ \\
\midrule
\multicolumn{2}{l}{Parameters} & 210.7M & 8.44M & 1.25M & 137k & 35.9k \\
\multicolumn{2}{l}{Training steps} & 22k & 17k & 37k & 83k & 39k \\
\midrule
Canonical (3) & 0c & .865 & .880 & .861 & .874 & .865 \\
 & 1c & .928 & .873 & .905 & .906 & .856 \\
 & 2c & .951 & .899 & .919 & .939 & .892 \\
\midrule
Semi-real (9) & 0c & .645 & .616 & .608 & .634 & .610 \\
 & 1c & .800 & .718 & .758 & .725 & .669 \\
 & 2c & .874 & .804 & .834 & .830 & .770 \\
\midrule
Simulated (12) & 0c & .700 & .682 & .671 & .694 & .674 \\
 & 1c & .832 & .757 & .795 & .770 & .716 \\
 & 2c & .893 & .828 & .855 & .857 & .800 \\
 & Mean & .808 & .756 & .774 & .774 & .730 \\
\midrule
TAC & 0c & .521 & .573 & .509 & .542 & .497 \\
 & 1c & .512 & .520 & .524 & .516 & .507 \\
 & 2c & .519 & .541 & .534 & .561 & .556 \\
\midrule
DaphniaExt & 0c & .529 & .355 & .583 & .437 & .373 \\
 & 1c & .661 & .572 & .533 & .591 & .596 \\
 & 2c & .829 & .822 & .897 & .895 & .894 \\
\bottomrule
\end{tabular}
\end{table}
\paragraph{Context noise and mismatch}
We perturb the context episodes in two ways (Table~\ref{tab:si-abl-context}).
For context noise, we add zero-mean Gaussian noise with standard deviation $\sigma$ times the standard deviation of each context channel to the observed context signals; time stamps and RDTC are unchanged, and the query is not perturbed.
The unperturbed runs reproduce Table~\ref{tab:auroc-results-best-col} within .02 in 44 of 45 cells (R-AMOC without context: .233 instead of .257).
For mismatched context, each query receives all its context episodes from one other sample, drawn at random per query either from the same system as the query, with other parameters and auxiliary dynamics, or from a different system; within that sample, the episodes are selected as for matched context.
AUROC falls monotonically with noise on the simulated systems, mildly up to $\sigma=.5$ and clearly at $\sigma=1$ (two context episodes: from .893 to .802); DaphniaExt degrades in the same way (.829 to .730), and TAC stays at chance.
Mismatched context lowers AUROC on the simulated systems below the zero-context value of .700, both from the same system (.552 and .565 with one and two context episodes) and from a different system (.528 and .532).
With two same-system context episodes, the paired difference is $-.135$ to zero context and $-.328$ to matched context (95\% cluster-bootstrap intervals over query samples, 2{,}000 replicates: $[-.164, -.102]$ and $[-.356, -.299]$).
R-Compost is the only system where mismatched context scores about as high as matched context (same system: .684 and .662, against .620 and .669); there, zero-context TipPFN (.772) is also better than matched context, as in Table~\ref{tab:auroc-results-best-col}.
TAC and DaphniaExt offer no other sample to draw mismatched context from.
TipPFN's context gains therefore require context from the same generating draw $\psi$ as the query; context from another draw of the same system does not transfer.

\begin{table}[h]
\centering
\small
\caption{\textbf{Context noise and mismatch.} AUROC averaged over $\Delta>0$ with 1 or 2 context episodes (c). Noise: zero-mean Gaussian noise with $\sigma$ times the standard deviation of each context channel added to the context signals. Mismatched: all context episodes from one other sample, drawn at random per query, of the same system (other parameters and auxiliary dynamics) or of a different system, without noise; not defined for TAC and DaphniaExt.}
\label{tab:si-abl-context}
\begin{tabular}{llcccccc}
\toprule
& & \multicolumn{4}{c}{Context noise $\sigma$} & \multicolumn{2}{c}{Mismatched} \\
\cmidrule(lr){3-6}\cmidrule(lr){7-8}
Systems & & 0 & .25 & .5 & 1 & Same system & Other system \\
\midrule
Canonical (3) & 1c & .928 & .922 & .915 & .875 & .584 & .576 \\
 & 2c & .951 & .942 & .936 & .916 & .615 & .586 \\
\midrule
Semi-real (9) & 1c & .800 & .775 & .758 & .718 & .541 & .512 \\
 & 2c & .874 & .853 & .823 & .764 & .549 & .514 \\
\midrule
Simulated (12) & 1c & .832 & .812 & .797 & .757 & .552 & .528 \\
 & 2c & .893 & .875 & .851 & .802 & .565 & .532 \\
\midrule
TAC & 1c & .512 & .504 & .490 & .479 & -- & -- \\
 & 2c & .519 & .514 & .510 & .504 & -- & -- \\
\midrule
DaphniaExt & 1c & .661 & .627 & .598 & .526 & -- & -- \\
 & 2c & .829 & .813 & .791 & .730 & -- & -- \\
\bottomrule
\end{tabular}
\end{table}
\paragraph{Sample efficiency (indicative)}
To estimate how much training data TipPFN needs, we evaluated intermediate checkpoints of the $d=96$ variant, which was trained on 200{,}000 distinct training draws per epoch with 16 samples per step (Table~\ref{tab:si-abl-sample-eff}).
With two context episodes, it reaches 90\% of its final AUROC (37k steps, about three epochs) after 8\% of one epoch on the canonical and b-tipping systems and on DaphniaExt, and after 32\% on the r-tipping systems; TAC stays near chance at every checkpoint.
The ratio is taken on the raw AUROC scale, which flatters weak cells, and the curves are not monotone.
A checkpoint saved after a fraction of the data has also received fewer optimizer steps, so this analysis does not separate data volume from optimization progress.
We therefore report it as an indicative estimate, not as a controlled dataset-size ablation, which would require training separate models to convergence on subsets of the prior.

\begin{table}[h]
\centering
\small
\caption{\textbf{Sample efficiency (indicative).} AUROC with two context episodes, averaged over $\Delta>0$, for checkpoints of the $d=96$ variant after the given fraction of one epoch (200{,}000 training draws); \emph{final} is the evaluated end of training (37k steps, 2.96 epochs). The last column gives the first checkpoint at which AUROC reaches 90\% of the final value.}
\label{tab:si-abl-sample-eff}
\begin{tabular}{lccccccc}
\toprule
Systems & 4\% & 8\% & 16\% & 32\% & 96\% & Final & 90\% at \\
\midrule
Canonical (3) & .792 & .894 & .923 & .913 & .922 & .919 & 8\% \\
Semi-real b-tipping (5) & .649 & .827 & .884 & .897 & .907 & .896 & 8\% \\
Semi-real r-tipping (4) & .402 & .468 & .575 & .691 & .683 & .757 & 32\% \\
DaphniaExt & .463 & .772 & .873 & .819 & .821 & .836 & 8\% \\
TAC & .512 & .495 & .548 & .537 & .541 & .535 & -- \\
\bottomrule
\end{tabular}
\end{table}

\FloatBarrier
\subsubsection{SWEC-iEEG}
\label{si-sec:swec-results}
Seizure onset prediction is a challenging and clinically relevant test case for transition forecasting, as it requires anticipating a rapid shift from interictal to ictal dynamics.
In SWEC-iEEG, the model operates on multichannel iEEG bandpower trajectories.
We interpret seizure onset as the critical event and evaluate prediction on preictal segments relative to interictal control segments from the same patient, drawn from periods at least one hour away from any seizure episode.
No method separates pre-ictal from interictal segments: in Table~\ref{tab:auroc-results-best-col}, all columns lie between .49 and .56, and at lead times of 10--40\,s between .46 and .55 (Fig.~\ref{fig:auroc-all-colwise-best}).
The higher TipPFN zero-context value comes from $\Delta = 60$ and $80$\,s, where the recordings before onset are too short for a full window, so critical query windows hold 89--109 observed steps against 128 for interictal ones.

\begin{figure}[bth]
  \centering
  \includegraphics[width=\linewidth]{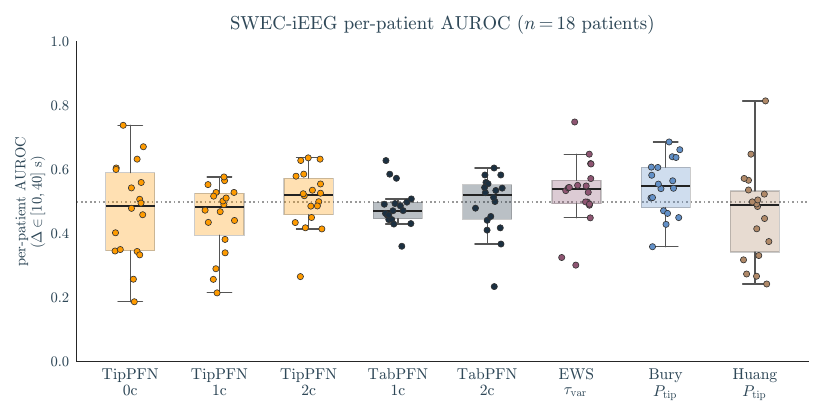}
  \caption{
    \textbf{SWEC per-patient AUROC distribution}.
    Each dot is one patient's AUROC computed within that patient ($n=18$) and averaged over $\Delta = 10$--$40$\,s, at the setting of each column in Table~\ref{tab:auroc-results-best-col}.
    For the uni-variate baseline models, this includes a per-signal-band selection: the bands
    \texttt{mrbp\_0p50\_0p66hz} (delta), \texttt{mrbp\_3p58\_4p74hz}
    (theta), and \texttt{mrbp\_19p35\_25p64hz} (beta) are scored
    separately and the best band is used, matching the choice in Table~\ref{tab:auroc-results-best-col}) and Fig.~\ref{fig:auroc-all-colwise-best}.
}
  \label{fig:si-swec-per-patient-distribution}
\end{figure}

With neurological EEG patterns being highly patient-specific, one interesting use case for TipPFN is to evaluate individual seizure risk by including only same-patient observations into the context.
Fig.~\ref{fig:si-swec-per-patient-distribution} shows the resulting patient-specific AUROC scores.
They scatter around chance for every method (medians .47--.55); with 2--23 seizures per patient, individual values are noisy.

\FloatBarrier
\newpage
\subsubsection{Daphnia Extinction}
\label{si-sec:daphnia-results}
The controlled \textit{Daphnia magna} extinction experiment of Drake and Griffen~\cite{drake2010early} is a valuable real-world transfer benchmark because it combines empirical observations with experimentally controlled deterioration, a well-characterized transcritical transition, and allows to match a Ricker-map simulation~\cite{ricker1954logistic} to generate simulated context.

\begin{figure}[h]
    \centering
    \includegraphics[width=1\linewidth]{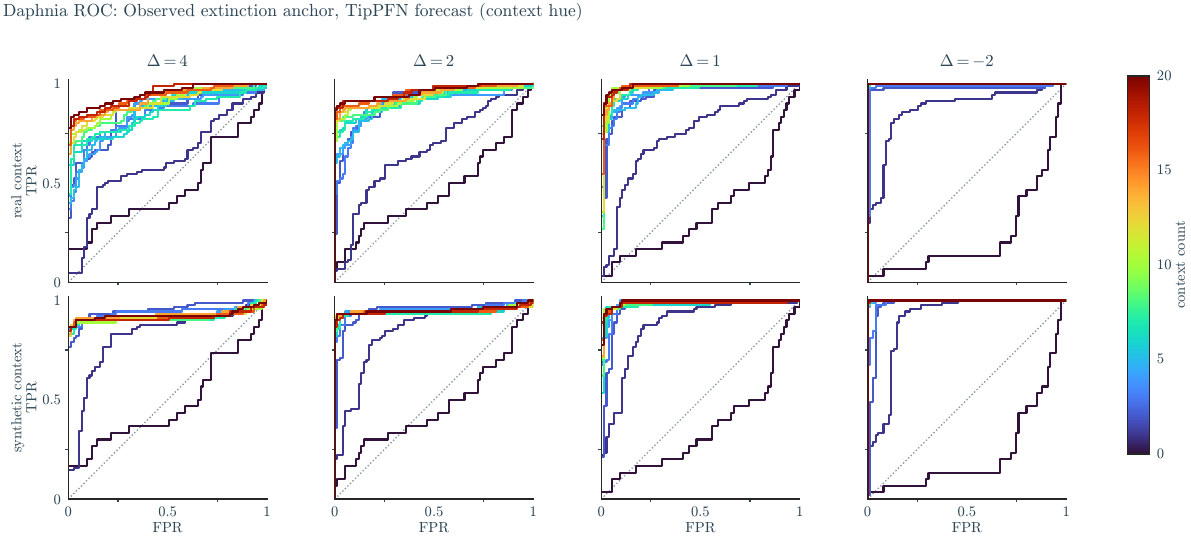}
    \caption{
        \texttt{DaphniaExt} dataset ROC curves for $W=16$, different $\Delta$ and increasing context episodes, using real observations as context (top row) or purely simulated context (bottom row).
        Surrogate $\Lambda$ was condition on $\Lambda = 0$ at the extinction time of the respective population.
    }
    \label{fig:daphnia-roc-observed-ext}
    \includegraphics[width=1\linewidth]{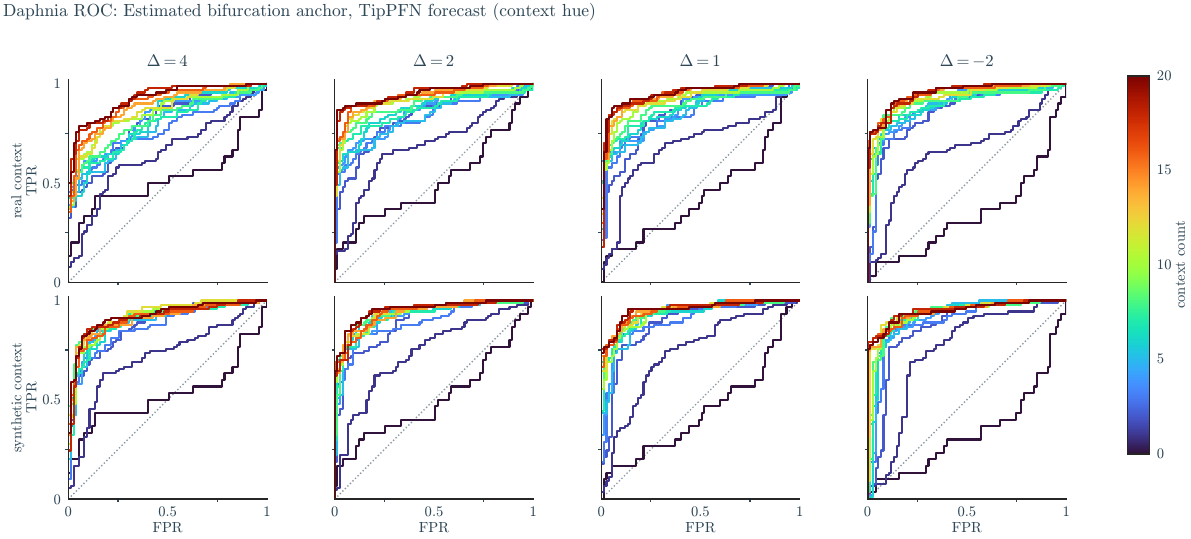}
    \caption{
        Like above, but surrogate $\Lambda$ was conditioned on $\Lambda = 0$ at a random time within the bifurcation-time interval $[271, 316]$ days estimated in~\cite{drake2010early}, aiming to estimate the unterlying transition.
    }
    \label{fig:daphnia-roc-estimated-bif}
\end{figure}

To construct context, we consider two surrogate RDTC choices: either setting $\Lambda=0$ at the observed extinction event, or sampling it from the bifurcation-time interval estimated in~\cite{drake2010early}.
Both observed and simulated contexts can yield comparable predictive performance, suggesting that matched simulations can act as a practical surrogate when repeated real trajectories are limited.
Simulated context tends to peak already with fewer episodes, see Fig.~\ref{fig:daphnia-roc-observed-ext} and~\ref{fig:daphnia-roc-estimated-bif}.

\begin{figure}[h]
    \centering
    \includegraphics[width=1\linewidth]{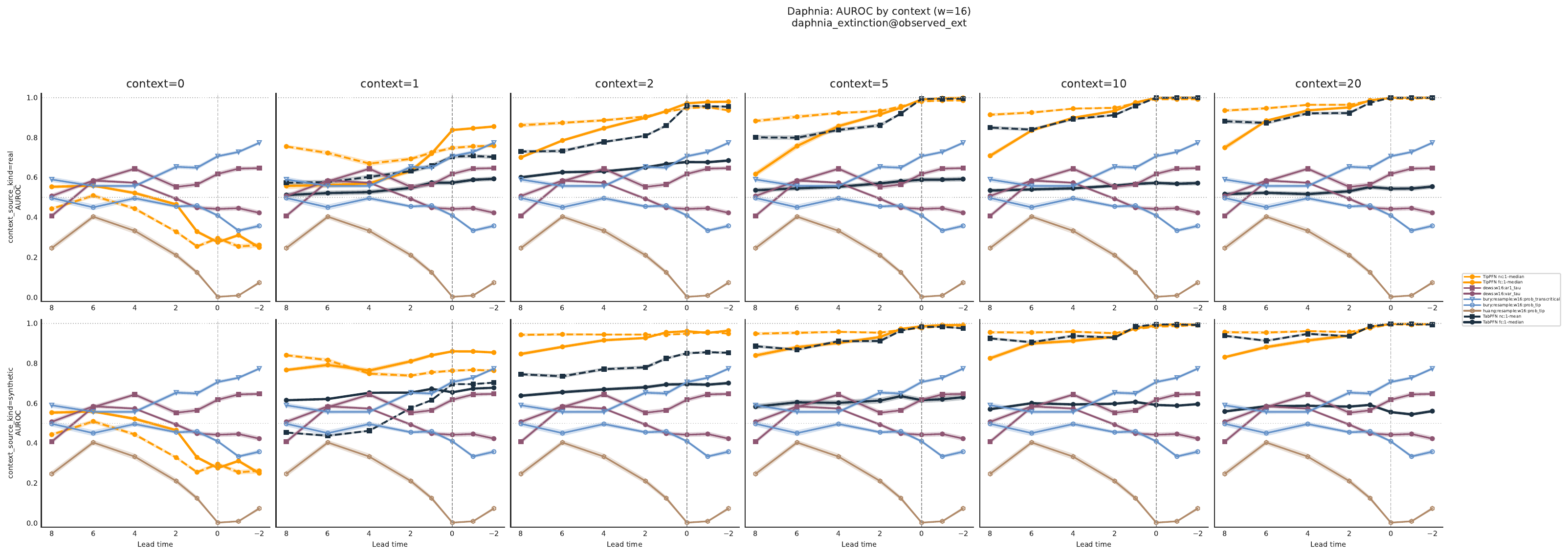}
    \caption{
        \texttt{DaphniaExt} dataset AUROC curves for $W=16$, different context sizes, using real observations as context (top row) or purely simulated context (bottom row).
        Surrogate $\Lambda$ was conditioned on $\Lambda = 0$ at the extinction time of the respective population.
    }
    \includegraphics[width=1\linewidth]{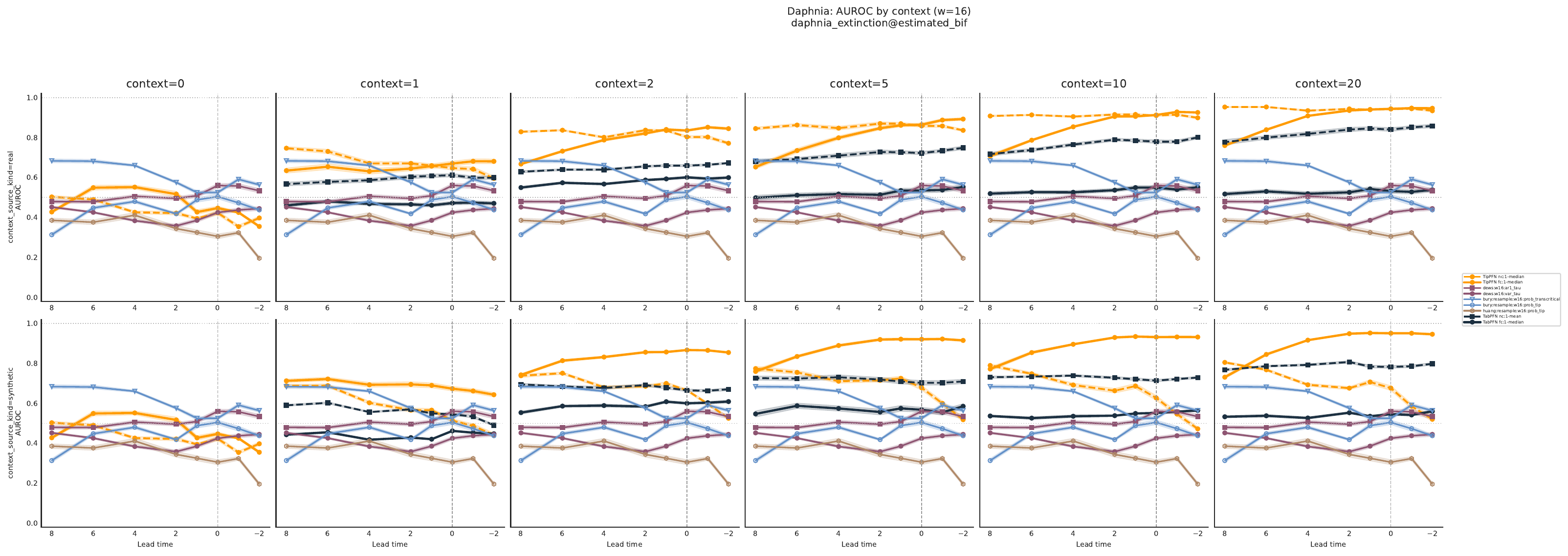}
    \caption{
        Like above, but with surrogate $\Lambda$ on the estimated bifurcation, the systems hidden critical transition.
    }
    \label{fig:auroc-daphnia-rdtc-choice}
\end{figure}

The two $\Lambda$ target constructions also reveal a qualitative difference between PFN-style models: TabPFN scores en par with TipPFN when it comes to distinguishing extinction; however, the arguably more difficult detection is that of the underlying bifurcation, which TipPFN discerns more clearly than other baselines, see Fig.~\ref{fig:auroc-daphnia-rdtc-choice}.

\FloatBarrier
\newpage
\subsubsection{TAC}
\label{si-sec:tac-results}
\begin{figure}[h]
    \centering
    \includegraphics[width=1\linewidth]{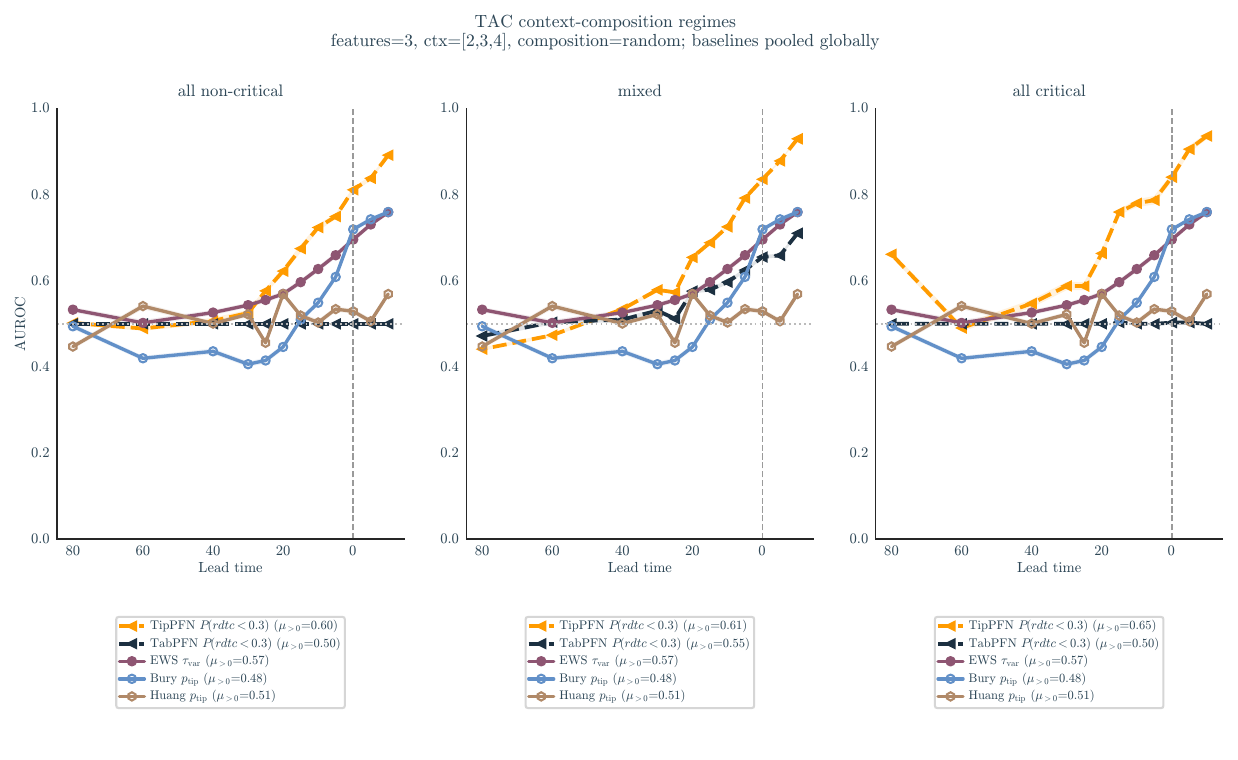}
    \caption{
        \texttt{TAC} dataset AUROC scores over lead time for contexts of different composition.
    }
    \label{fig:auroc-tac-cnc}
\end{figure}

The thermo-acoustic combustor dataset provides a particularly noisy real-world benchmark built around a stochastic subcritical Hopf transition~\cite{bonciolini2018experiments}.
We formulate the task as prediction of the Hopf bifurcation and assign surrogate $\Lambda$ labels by anchoring a linear ramp at the known critical point.
In the basic setting, the Hopf-specific Bury model performs very well, consistent with its close match to the target mechanism.
TipPFN performance is higher for an increased context size of 4; it additionally increases when composing the context with exclusively critical context episodes, presumably because they are the more informative and distinguishing signal.
Fig.~\ref{fig:auroc-tac-cnc} shows the effect on AUROC when filling the context with only non-critical, only critical, or random episodes, causing TipPFN to outperform Bury.

\FloatBarrier
\newpage
\subsubsection{Zero-shot datasets}
\begin{figure}[hb]
    \centering
    \includegraphics[width=1\linewidth]{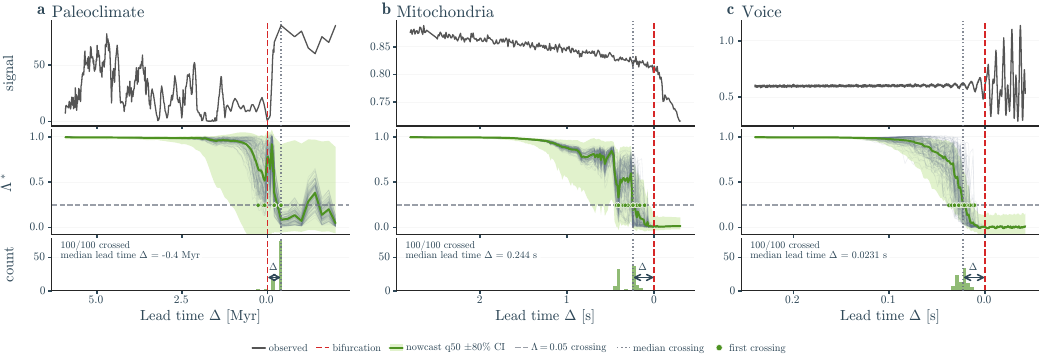}
    \caption{
        \textbf{Zero-shot TipPFN RDTC nowcasts} on three additional uni-variate real-world time series (see Fig.~\ref{fig:zero-shot-selected}).
        Note the crossing threshold $\Lambda^* = \tanh(5\Lambda) \approx 0.245$. 
    }
    \label{fig:zero-shot-si}
\end{figure}

In addition to Fig.~\ref{fig:zero-shot-selected}, we study three further zero-shot time series, depicted in Fig.~\ref{fig:zero-shot-si}.
While we find early warning ($\Delta > 0$) for the \texttt{mitochondria} and \texttt{voice} dataset, the RDTC prediction for the paleoclimate (\texttt{greenhouse\_earth}) dataset only crossed the threshold  after the critical transition.

\FloatBarrier
\subsubsection{Proximity of real-world inputs to the training prior}\label{si-sec:prior-coverage}
PFN predictions approximate the posterior predictive only for inputs supported by the training prior.
We therefore measure how close the real-world inputs of TAC and DaphniaExt lie to inputs seen in training.
For every evaluation query we take one segment, the last $L\in\{8,16,32,64,128\}$ steps of the model-visible query window after input normalization, with all input channels.
The reference consists of 2{,}195{,}904 prior segments of the same length and number of channels, eight per training draw (all 274{,}488 draws), constructed like training queries: resampled in time, with randomly selected observed variables, and normalized in the same way as the evaluation input.
The distance between two segments is the root-mean-square difference over all entries; RDTC does not enter it.
A segment counts as covered if the distance to its 20th-nearest prior segment is at most the 95th percentile of the same quantity for 19{,}668 held-out prior segments (one per draw from 12 of the 167 blocks of training draws, searched against the other 155 blocks), calibrated separately for each $L$ and number of channels.
Table~\ref{tab:si-prior-coverage} reports the covered fraction with 95\% intervals from 2{,}000 bootstrap replicates over source recordings (TAC) or segments (DaphniaExt).

Except for short TAC segments ($L\le16$: 51\% and 77\%), at least 94\% of the segments are covered, and all TAC segments with $L\ge64$; by construction, 95\% of held-out prior segments are covered.
Restricting the neighbors to prior segments whose RDTC is close to that of the query (band width 0.025--0.1) does not lower any of these fractions.
The analysis measures proximity in the space of normalized observations: it shows that the real-world inputs resemble inputs TipPFN was trained on, not that the evaluation data follow the prior distribution or that the physical systems belong to the model class of the prior.

\begin{table}[h]
\centering
\footnotesize
\setlength{\tabcolsep}{4pt}
\caption{\textbf{Coverage of real-world evaluation segments by the training prior.}
Percentage of segments of length $L$ whose distance to their 20th-nearest prior segment is at most the 95th percentile for held-out prior segments (Appendix~\ref{si-sec:prior-coverage}), with 95\% bootstrap interval; no interval where all segments are covered.
DaphniaExt windows have at most 16 steps; its two rows place $\Lambda=0$ at the observed extinction or within the estimated bifurcation interval.}
\label{tab:si-prior-coverage}
\begin{tabular}{lrccccc}
\toprule
Dataset & Segments & $L=8$ & $L=16$ & $L=32$ & $L=64$ & $L=128$ \\
\midrule
TAC & 160 & 51.3 {\scriptsize [42.5, 64.6]} & 76.9 {\scriptsize [66.7, 81.7]} & 94.4 {\scriptsize [92.4, 99.0]} & 100.0 & 100.0 \\
DaphniaExt, extinction & 55 & 94.5 {\scriptsize [87.3, 100.0]} & 94.5 {\scriptsize [89.0, 100.0]} & -- & -- & -- \\
DaphniaExt, bifurcation & 55 & 100.0 & 100.0 & -- & -- & -- \\
\bottomrule
\end{tabular}
\end{table}

\FloatBarrier
\newpage
\subsection{AUROC Uncertainty}\label{sec:si-auroc-uncertainty}

\begin{table}[h]
\centering
\caption{
    Per-system standard errors for the macro AUROC over $\Delta > 0$ for the values reported in Tables~\ref{tab:auroc-results-best-col} and~\ref{tab:auroc-results-best-per-system}.
    \emph{Total samples} is $\sum_c (n_{\text{pos}}^{(c)} + n_{\text{neg}}^{(c)})$ across all contributing $(\Delta, \text{system})$ points
    \emph{pooled $n_{\min}$} is $\sum_c \min(n_{\text{pos}}^{(c)}, n_{\text{neg}}^{(c)})$,  the effective balanced-sample count entering equation~\eqref{eq:si-hm-se}.
    $\mathrm{SE}_{\text{HM}}$ is reported at $A=0.8$; values at $A=0.5$ are $\sim 25\%$ larger.
    $\mathrm{SE}_{\text{BS}}$ is the propagated balanced-subsample uncertainty from equation~\eqref{eq:si-bs-se}; for TAC it is effectively zero because $n_{\text{pos}}{=}n_{\text{neg}}$ at every contributing point and no subsampling occurs.
}
\label{tab:si-auroc-se-per-system}
\begin{tabular}{lrrrr}
\toprule
System / dataset & total samples & pooled $n_{\min}$ & $\mathrm{SE}_{\text{HM}}$ at $A{=}0.8$ & $\mathrm{SE}_{\text{BS}}$ \\
\midrule
\multicolumn{5}{l}{\emph{Canonical}}\\
B-Fold & 911 & 407 & 0.020 & 0.0009 \\
B-Hopf & 1{,}019 & 465 & 0.019 & 0.0008 \\
B-Trans. & 910 & 418 & 0.020 & 0.0005 \\
\midrule
\multicolumn{5}{l}{\emph{Semi-real}}\\
B-Harv. & 434 & 194 & 0.029 & 0.0008 \\
B-RM TC & 672 & 300 & 0.023 & 0.0021 \\
B-RM Hopf & 737 & 335 & 0.022 & 0.0013 \\
B-SEIRx & 370 & 164 & 0.031 & 0.0022 \\
B-AMOC & 597 & 273 & 0.024 & 0.0019 \\
R-Bautin & 738 & 318 & 0.022 & 0.0042 \\
R-SN & 768 & 346 & 0.022 & 0.0017 \\
R-Compost & 682 & 308 & 0.023 & 0.0023 \\
R-AMOC & 608 & 272 & 0.024 & 0.0029 \\
\midrule
\multicolumn{5}{l}{\emph{Others}}\\
SWEC-iEEG & 14{,}526 & 2{,}088 & 0.009 & 0.0013 \\
TAC & 21{,}600 & 10{,}800 & 0.004 & $\sim 0$ \\
DaphniaExt & 4{,}400 & 2{,}000 & 0.009 & 0.0006 \\
\midrule
\multicolumn{5}{l}{\emph{Summary}}\\
All systems & 49{,}070 & 18{,}688 & 0.003 & 0.0005 \\
\bottomrule
\end{tabular}
\end{table}

For each AUROC value reported, we compute a balanced AUROC by drawing $K=10$ class-balanced subsamples without replacement from the available critical and non-critical query windows, evaluating \texttt{sklearn.metrics.roc\_auc\_score} on each, and reporting the mean.
Each subsample contains $n_{\min} = \min(n_{\text{pos}}, n_{\text{neg}})$ examples per class.

We attach a sample-size standard error using a simplified form of the Hanley–McNeil approximation,
\begin{equation}
  \mathrm{SE}_{\text{HM}} \;=\; \sqrt{\frac{A\,(1-A)}{n_{\min}^{\text{pooled}}}}\,,
  \label{eq:si-hm-se}
\end{equation}
where $n_{\min}^{\text{pooled}} = \sum_{c} \min(n_{\text{pos}}^{(c)}, n_{\text{neg}}^{(c)})$ sums per-point minority-class counts across all lead times $\Delta > 0$ contributing to the reported macro AUROC.
Equation~\eqref{eq:si-hm-se} ignores the correlation between lead times and between overlapping windows of the same episode, so it understates the uncertainty; the cluster bootstrap in Appendix~\ref{si-sec:bootstrap-ci} resamples whole source samples and accounts for this dependence.
AUROC differences below $2\,\mathrm{SE}_{\text{HM}}$ should be regarded as not statistically resolved.

We additionally report the propagated standard error of the macro AUROC from the balanced-subsampling procedure.
For each contributing $(\Delta, \text{system})$ point we have an empirical std $\sigma_{\text{BS}}^{(c)}$ across the $K{=}10$ subsamples; pooling these across the $K_{\text{macro}}$ points contributing to a system's macro AUROC gives
\begin{equation}
  \mathrm{SE}_{\text{BS}} \;=\; \frac{1}{K_{\text{macro}}}\sqrt{\frac{1}{K}\sum_{c=1}^{K_{\text{macro}}}\bigl(\sigma_{\text{BS}}^{(c)}\bigr)^2}\,.
  \label{eq:si-bs-se}
\end{equation}
This quantity captures sensitivity of the AUROC estimator to which $n_{\min}$ examples are chosen for a given imbalance ratio at each point.
It does not capture sampling variation in the underlying episode population, and is therefore consistently $5$--$30\times$ smaller than $\mathrm{SE}_{\text{HM}}$ in our data.
We report it as a separate diagnostic rather than combining it with $\mathrm{SE}_{\text{HM}}$, since the two estimate related but distinct variances and the second-order contribution would not move the displayed precision.

Per-system pooled $n_{\min}$, $\mathrm{SE}_{\text{HM}}$, $\mathrm{SE}_{\text{BS}}$, and the underlying total sample count are shown in Table~\ref{tab:si-auroc-se-per-system}.

\paragraph{Selection bias from best-of-$K$ picking.}
For columns whose value is the best across a family of configurations (e.g. any column that selects across window size, feature-budget, or score-head combinations within a method), the reported AUROC is the maximum over $K$ correlated noisy estimates and is therefore positively biased relative to the true expected AUROC at the chosen setting.
Under the null hypothesis that all candidate settings have equal expected AUROC, the expected maximum exceeds the per-cell mean by approximately $\sigma \cdot \Phi^{-1}(1 - 1/K)$, where $\sigma$ is the per-cell sampling std and $\Phi^{-1}$ is the inverse normal CDF.
For our typical $K \in [10, 150]$ candidate cells per (method, system) and $\sigma \approx \mathrm{SE}_{\text{HM}}$, the per-cell inflation is $0.02$--$0.06$ in the iid-null limit; correlation across candidates pulls this back to a typical $0.01$--$0.03$.

Appendix~\ref{si-sec:heldout-selection} repeats the column-wise selection on held-out data; the column means change by at most .02.
Differences between methods at the per-system level should therefore be interpreted as upper bounds on the true gap rather than unbiased point estimates.

\subsubsection{Per-system bootstrap confidence intervals}\label{si-sec:bootstrap-ci}
The standard errors above treat query windows as independent and ignore the correlation between lead times.
Here we instead resample the independent units from which the query windows are drawn and report confidence intervals for the TipPFN columns of Table~\ref{tab:auroc-results-best-col} and for the change in AUROC with context.
We use the scores and settings of Table~\ref{tab:auroc-results-best-col}; the point estimate is its balanced AUROC averaged over $\Delta>0$ and reproduces the table.
The cluster bootstrap~\cite{efron1993introduction} ($B=10{,}000$) resamples the generating draw $\psi$ for the simulated systems, the recording for TAC and the population episode for DaphniaExt; each replicate uses the same units for all lead times and context counts, so differences between context counts are paired.
We report two-sided 90\% percentile intervals and bootstrap $p$-values, Holm-corrected~\cite{holm1979simple} across the 14 rows within each column (cells against 0.5, differences against 0); the smallest attainable corrected $p$-value is about 0.003.
SWEC-iEEG is not included (Appendix~\ref{si-sec:swec-results}).

With two context episodes, TipPFN is significantly above chance on all 14 rows (Table~\ref{tab:si-bootstrap-ci}).
At zero context, B-SEIRx, TAC and DaphniaExt are not distinguishable from chance, and R-Bautin and R-AMOC are significantly below chance.
The change from zero to two context episodes is significant on every row except TAC and positive on all of them except R-Compost, where zero-context TipPFN performs best ($-.102$, 90\% CI [$-.159$, $-.038$]); the first context episode significantly lowers AUROC on R-Compost.
For the real-world datasets, the intervals do not account for the post hoc choice of window and are therefore optimistic; for TAC, 1{,}100 of the 10{,}000 replicates contain a lead time with only one class, which makes its intervals approximate.

\begin{table}[h]
\centering
\caption{\textbf{Cluster-bootstrap confidence intervals for TipPFN.}
AUROC averaged over $\Delta>0$ with 90\% confidence interval for 0, 1 and 2 context episodes (c), and paired changes from 0c.
\emph{Units} is the number of independent units resampled.
Daggers mark values whose Holm-corrected $p$-value exceeds 0.05 (cells tested against 0.5, changes against 0); underlined cells lie significantly below 0.5.}
\label{tab:si-bootstrap-ci}
\resizebox{\columnwidth}{!}{%
\begin{tabular}{lrccccc}
\toprule
System/data & Units & 0c & 1c & 2c & 0c$\to$1c & 0c$\to$2c \\
\midrule
\multicolumn{7}{l}{\emph{Canonical}} \\
B-Fold & 41 & .876 [.839, .914] & .929 [.906, .953] & .955 [.936, .971] & $+.053$ [$+.014$, $+.096$]$^\dagger$ & $+.078$ [$+.039$, $+.118$] \\
B-Hopf & 46 & .861 [.822, .897] & .935 [.898, .964] & .952 [.913, .982] & $+.074$ [$+.029$, $+.119$] & $+.091$ [$+.046$, $+.136$] \\
B-Trans. & 41 & .864 [.819, .902] & .925 [.886, .956] & .950 [.918, .974] & $+.061$ [$+.018$, $+.108$]$^\dagger$ & $+.086$ [$+.045$, $+.130$] \\
\midrule
\multicolumn{7}{l}{\emph{Semi-real}} \\
B-Harv. & 20 & .846 [.790, .891] & .958 [.919, .982] & .974 [.951, .991] & $+.112$ [$+.072$, $+.157$] & $+.127$ [$+.087$, $+.182$] \\
B-RM TC & 31 & .815 [.762, .874] & .951 [.928, .972] & .963 [.943, .978] & $+.136$ [$+.080$, $+.183$] & $+.148$ [$+.092$, $+.195$] \\
B-RM Hopf & 33 & .769 [.717, .817] & .829 [.768, .885] & .904 [.866, .940] & $+.061$ [$-.008$, $+.128$]$^\dagger$ & $+.136$ [$+.071$, $+.201$] \\
B-SEIRx & 17 & .582 [.455, .687]$^\dagger$ & .822 [.713, .912] & .905 [.857, .956] & $+.240$ [$+.082$, $+.420$]$^\dagger$ & $+.324$ [$+.205$, $+.477$] \\
B-AMOC & 27 & .853 [.801, .900] & .898 [.847, .942] & .940 [.901, .971] & $+.045$ [$-.004$, $+.098$]$^\dagger$ & $+.087$ [$+.033$, $+.142$] \\
R-Bautin & 34 & \underline{.308} [.272, .341] & .543 [.485, .607]$^\dagger$ & .700 [.632, .777] & $+.235$ [$+.171$, $+.311$] & $+.392$ [$+.322$, $+.477$] \\
R-SN & 35 & .644 [.569, .717] & .802 [.749, .850] & .892 [.846, .932] & $+.158$ [$+.104$, $+.213$] & $+.248$ [$+.186$, $+.313$] \\
R-Compost & 31 & .767 [.726, .804] & .623 [.561, .689] & .665 [.615, .721] & $-.144$ [$-.214$, $-.069$] & $-.102$ [$-.159$, $-.038$] \\
R-AMOC & 28 & \underline{.257} [.206, .296] & .782 [.704, .848] & .913 [.863, .951] & $+.525$ [$+.436$, $+.614$] & $+.656$ [$+.597$, $+.719$] \\
\midrule
\multicolumn{7}{l}{\emph{Real-world \& Sim-to-Real}} \\
TAC & 12 & .521 [.445, .542]$^\dagger$ & .512 [.498, .523]$^\dagger$ & .519 [.505, .556] & $-.009$ [$-.029$, $+.061$]$^\dagger$ & $-.002$ [$-.025$, $+.104$]$^\dagger$ \\
DaphniaExt & 55 & .527 [.459, .598]$^\dagger$ & .663 [.623, .700] & .833 [.790, .868] & $+.136$ [$+.058$, $+.210$] & $+.307$ [$+.219$, $+.388$] \\
\bottomrule
\end{tabular}
}
\end{table}

\subsubsection{Score Analysis}\label{si-sec:score-head-analysis}

The CNN baselines (Bury, Huang, Zhuge) require fixed-length input windows; signals shorter than the network's expected length are extended via one of three \texttt{short\_mode} schemes (\texttt{pad}, \texttt{backfill}, \texttt{resample}).
Rather than pinning a single short-mode globally, each baseline column in the headline AUROC table independently picks its column-best \texttt{short\_mode}, and the row-wise variant (Table~\ref{tab:auroc-results-best-per-system}) re-picks per system; the picked token is reported in every baseline cell.
Across all baseline cells of the row-wise table, \texttt{resample} dominates ($\sim$71\%), with \texttt{backfill} ($\sim$19\%) and \texttt{pad} ($\sim$10\%) chosen on a minority of systems where the signal-to-padding scaling differs.
Because the picked token is already disclosed at cell granularity and no global pin is asserted, we do not include a dedicated short-mode sensitivity SI table; the row-wise picks are self-documenting.

A handful of baseline scores are anti-correlated with criticality on specific datasets, producing AUROCs significantly below chance ($<0.5$).
The clearest case is Zhuge's \texttt{signed\_tip\_margin} on SWEC-iEEG, which is constant within each class (all non-critical queries score $0$, all critical queries at a given $\Delta$ share one negative value), so its AUROC is exactly $0$ for every window and band; we therefore omit Zhuge on SWEC-iEEG.
While excluding degenerate scores, we report all baseline AUROCs as-is without auto-flipping anti-correlated scores -- flipping at evaluation time would amount to a per-cell sign-tuning.

\begin{table}[h]
\centering
\caption{Selected configuration for each column of the headline AUROC table (Table~\ref{tab:auroc-results-best-col}). Score names are defined in Table~\ref{tab:si-score-definitions}. For TipPFN and TabPFN the configuration spans the score head, context size, query-window length, and feature budget; for the baseline methods only the parameters consumed by that method are listed (e.g.\ EWS reads window and indicator; Bury, Huang and Zhuge additionally read \textit{short\_mode}). Each baseline column independently picks its best query-window length (column-best over $\{64, 96, 128\}$) and, for SWEC where the score is computed across multiple bandpower bands, its best signal band. Daphnia uses an incomparable window range $\{8, 12, 16\}$, so every column reports the per-cell best-of-$\{8,12,16\}$ for daphnia (matching the row-wise convention). The row-wise best variant (Table~\ref{tab:auroc-results-best-per-system}) re-selects window, \textit{short\_mode}, and (on SWEC) signal band separately per system.}
\label{tab:auroc-column-choices}
\resizebox{\columnwidth}{!}{%
\begin{tabular}{lll}
\toprule
Column & Score name & Configuration \\
\midrule
TipPFN / 0c & \texttt{tippfn:fc:1-median} & head=fc:1-median, ctx=0, w=128, features=16 \\
TipPFN / 1c & \texttt{tippfn:fc:1-median} & head=fc:1-median, ctx=1, w=128, features=16 \\
TipPFN / 2c & \texttt{tippfn:fc:1-median} & head=fc:1-median, ctx=2, w=128, features=16 \\
TabPFN / 1c & \texttt{tabpfn:fc:1-median} & head=fc:1-median, ctx=1, w=64, features=16 \\
TabPFN / 2c & \texttt{tabpfn:fc:1-median} & head=fc:1-median, ctx=2, w=64, features=16 \\
EWS / best & \texttt{dews:w64:var\_tau} & w64, indicator=var\_tau \\
Bury / best & \texttt{bury:resample:w128:prob\_tip} & short\_mode=resample, w128, head=prob\_tip \\
Huang / best & \texttt{huang:resample:w64:prob\_tip} & short\_mode=resample, w64, head=prob\_tip \\
Zhuge / best & \texttt{zhuge:rdtc:resample:w128:signed\_tip\_margin} & control=rdtc, short\_mode=resample, w128, head=signed\_tip\_margin \\
\bottomrule
\end{tabular}
}
\end{table}

\begin{table}[h]
\centering
\caption{\textbf{Head sensitivity of the All-datasets summary row.} Mean AUROC over $\Delta > 0$ across all 15 system/data rows, broken down by TipPFN/TabPFN context column (column groups), score scope+head (rows), and query-window length (sub-columns) at features=16. Heads: \texttt{fc}/\texttt{nc} use the forecast/nowcast query position; \texttt{1-median} and \texttt{1-mean} are one minus the median or mean of the predicted RDTC distribution, and \texttt{P(rdtc<$\tau$)} is the predicted probability that RDTC falls below $\tau$ (Table~\ref{tab:si-score-definitions}). The headline col-wise table (Table~\ref{tab:auroc-results-best-col}) pins TipPFN/TabPFN to \texttt{fc:1-median} for every column. The best cell per column group is shown in bold with an asterisk ($\boldsymbol{\cdot}^{*}$); other cells within $\sqrt{\sigma_{\mathrm{best}}^2 + \sigma_{\mathrm{cell}}^2}$ of the best (cross-system SE) are also bold (no asterisk), indicating statistically indistinguishable choices.}
\label{tab:auroc-score-head-summary}
\resizebox{\columnwidth}{!}{%
\begin{tabular}{lccccccccccccccc}
\toprule
 & \multicolumn{3}{c}{TipPFN / 0c} & \multicolumn{3}{c}{TipPFN / 1c} & \multicolumn{3}{c}{TipPFN / 2c} & \multicolumn{3}{c}{TabPFN / 1c} & \multicolumn{3}{c}{TabPFN / 2c} \\
\cmidrule(lr){2-4} \cmidrule(lr){5-7} \cmidrule(lr){8-10} \cmidrule(lr){11-13} \cmidrule(lr){14-16}
Score head & w64 & w96 & w128 & w64 & w96 & w128 & w64 & w96 & w128 & w64 & w96 & w128 & w64 & w96 & w128 \\
\midrule
fc:1-median & \textbf{.683} & \textbf{.688} & \textbf{.670} & .708 & \textbf{.759} & \textbf{.779} & .775 & \textbf{.819} & \textbf{.838} & .522 & .514 & .512 & .666 & .655 & .652 \\
fc:1-mean & \textbf{.679} & \textbf{.675} & \textbf{.681} & .726 & \textbf{.765} & \textbf{.780} & .790 & \textbf{.825} & \textbf{.842} & .518 & .511 & .507 & .647 & .637 & .635 \\
fc:P(rdtc<0.05) & \textbf{.676} & \textbf{.672} & \textbf{.673} & .726 & \textbf{.764} & \textbf{.781} & .786 & \textbf{.822} & \textbf{.842} & .498 & .494 & .487 & .605 & .597 & .590 \\
fc:P(rdtc<0.1) & \textbf{.676} & \textbf{.674} & \textbf{.674} & .728 & \textbf{.765} & \textbf{.780} & .787 & \textbf{.823} & \textbf{.843} & .513 & .507 & .501 & .617 & .609 & .604 \\
fc:P(rdtc<0.2) & \textbf{.676} & \textbf{.676} & \textbf{.670} & .729 & \textbf{.763} & \textbf{.779} & .788 & \textbf{.819} & \textbf{.840} & \textbf{.531} & .524 & .525 & .634 & .625 & .628 \\
fc:P(rdtc<0.3) & \textbf{.671} & \textbf{.679} & \textbf{.669} & .733 & \textbf{.764} & \textbf{.777} & .788 & \textbf{.817} & \textbf{.838} & \textbf{.546} & \textbf{.539} & \textbf{.541} & .658 & .649 & .650 \\
nc:1-median & \textbf{.634} & \textbf{.685} & \textbf{.663} & .662 & \textbf{.755} & \textbf{.789} & .770 & \textbf{.833} & \textbf{.856} & \textbf{.559}$^{*}$ & \textbf{.540} & \textbf{.526} & \textbf{.729}$^{*}$ & \textbf{.707} & \textbf{.692} \\
nc:1-mean & \textbf{.645} & \textbf{.689}$^{*}$ & \textbf{.674} & .681 & \textbf{.771} & \textbf{.794} & .792 & \textbf{.846} & \textbf{.864}$^{*}$ & \textbf{.544} & \textbf{.525} & .511 & \textbf{.712} & \textbf{.689} & .673 \\
nc:P(rdtc<0.05) & .501 & \textbf{.662} & \textbf{.659} & .533 & .582 & .634 & .578 & .654 & .703 & .516 & .485 & .442 & .660 & .617 & .563 \\
nc:P(rdtc<0.1) & .502 & \textbf{.671} & \textbf{.664} & .558 & .612 & .707 & .620 & .702 & .765 & .519 & .487 & .459 & .660 & .626 & .597 \\
nc:P(rdtc<0.2) & .507 & \textbf{.681} & \textbf{.659} & .606 & .694 & \textbf{.777} & .695 & .777 & \textbf{.836} & \textbf{.534} & .518 & .523 & \textbf{.679} & .660 & .655 \\
nc:P(rdtc<0.3) & .610 & \textbf{.676} & \textbf{.655} & .668 & \textbf{.769} & \textbf{.800}$^{*}$ & .759 & \textbf{.834} & \textbf{.862} & \textbf{.542} & \textbf{.539} & \textbf{.551} & \textbf{.705} & \textbf{.689} & \textbf{.684} \\
\bottomrule
\end{tabular}
}
\end{table}

\end{document}